\documentclass{article}

\usepackage[preprint]{neurips_2023}

\usepackage[utf8]{inputenc} 
\usepackage[T1]{fontenc}    
\usepackage{hyperref}       
\usepackage{url}            
\usepackage{booktabs}       
\usepackage{amsfonts}       
\usepackage{nicefrac}       
\usepackage{microtype}      
\usepackage{xcolor}         
\usepackage{comment}

\newcommand{\squeeze}{}
\usepackage{natbib}

\setcitestyle{numbers,square}
\usepackage[flushleft]{threeparttable} 

\usepackage[title]{appendix}

\usepackage{subcaption}
\usepackage{amssymb}
\usepackage{mathrsfs}
\usepackage[tbtags]{amsmath}
\usepackage{mathtools}
\usepackage{tikz-cd}
\usepackage{algorithm}
\usepackage{algpseudocode}
\usetikzlibrary {graphs} 
\usetikzlibrary{shapes.geometric, arrows}
\usepackage{enumitem}
\usetikzlibrary{arrows.meta}

\newtheorem{theorem}{Theorem}
\newtheorem{definition}{Definition}
\newtheorem{lemma}{Lemma}
\newtheorem{claim}{Claim}
\newtheorem{corollary}{Corollary}

\newtheorem{assumption}{Assumption}

\usepackage[colorinlistoftodos,bordercolor=orange,backgroundcolor=orange!20,linecolor=orange,textsize=scriptsize]{todonotes}

\usepackage{xspace}
\usepackage{tcolorbox}
\usepackage{pifont}
\usepackage{color}
\usepackage{float}
\definecolor{mydarkgreen}{RGB}{39,130,67}
\definecolor{mydarkred}{RGB}{192,47,25}
\definecolor{mydarkblue}{RGB}{39,47,180}

\newcommand{\blue}{\color{mydarkblue}}

\newcommand{\algname}[1]{{\sf \footnotesize \blue #1}\xspace}

\newcommand{\cC}{\mathcal{C}}

\newcommand{\mI}{\mathbf{I}}

\newcommand{\mbf}{\mathbf}

\newcommand{\bb}{\mathbb}

\newcommand{\bI}{\mbf I}

\newcommand{\bA}{\mbf A}

\newcommand{\rb}[1]{\left(#1\right)}

\newcommand{\norm}[1]{\left\| #1 \right\|}

\newcommand{\sqnorm}[1]{\left\| #1 \right\|^{2}}

\newcommand{\R}{\bb R}

\newcommand{\bbone}[1]{\mathbb{1}}      

\newcommand{\eqdef}{\stackrel{\text{def}}{=}}

\newcommand{\Exp}[1]{\mathbb{E}\left[#1\right]}

\tikzstyle{node} = [rectangle, rounded corners, 
minimum width=2cm, 
minimum height=1cm,
text centered, 
draw=black, 
fill=white]
\tikzstyle{arrow} = [thick,->,>=stealth]
\tikzstyle{doublearrow} = [thick,<->,>=stealth]
\setcitestyle{authoryear}

\title{A Guide Through the Zoo of Biased SGD}

\author{%
  Yury~Demidovich\\
  AI Initiative, KAUST\\
  \texttt{yury.demidovich@kaust.edu.sa} \\
   \And
   Grigory Malinovsky \\
	AI Initiative, KAUST\\
	\texttt{grigorii.malinovskii@kaust.edu.sa} \\
   \AND
   Igor Sokolov \\
   AI Initiative, KAUST\\
   \texttt{igor.sokolov.1@kaust.edu.sa} \\
   \And
   Peter Richt\'{a}rik \\
   AI Initiative, KAUST\\
   \texttt{peter.richtarik@kaust.edu.sa} \\
}

\begin{document}

	\maketitle
\begin{abstract}
	Stochastic Gradient Descent (\algname{SGD}) is arguably the most important single algorithm in modern machine learning. Although \algname{SGD} with unbiased gradient estimators has been studied extensively over at least half a century, \algname{SGD} variants relying on biased estimators are rare. Nevertheless, there has been an increased interest in this topic in recent years. However, existing literature on \algname{SGD} with biased estimators (\algname{BiasedSGD}) lacks coherence since each new paper relies on a different set of assumptions, without any clear understanding of how they are connected, which may lead to confusion. We address this gap by establishing connections among the existing assumptions, and presenting a comprehensive map of the underlying relationships. Additionally, we introduce a new set of assumptions that is provably weaker than all previous assumptions, and use it to present a thorough analysis of \algname{BiasedSGD} in both convex and non-convex settings, offering advantages over previous results. We also provide examples where biased estimators outperform their unbiased counterparts or where unbiased versions are simply not available. Finally, we demonstrate the effectiveness of our framework through experimental results that validate our theoretical findings.
\end{abstract}

\section{Introduction}	
Stochastic Gradient Descent (\algname{SGD}) \citep{RobbinsMonro:1951} is a widely used and effective algorithm for training various models in machine learning. The current state-of-the-art methods for training deep learning models are all variants of \algname{SGD} \citep{GoodBenCour, Sun}. The algorithm has been extensively studied in recent theoretical works \citep{BotCurNoce, gower2019sgd, khaled2022better}. In practice and theory, \algname{SGD} with \textit{unbiased} gradient oracles is mostly used. However, there has been a recent surge of interest in \algname{SGD} with \textit{biased} gradient oracles, which has been studied in several papers and applied in different domains.

In distributed parallel optimization where data is partitioned across multiple nodes, communication can be a bottleneck, and techniques such as structured sparsity \citep{alistarh2018sparse, WangniWangLiuZhang} or asynchronous updates \citep{NiuRechtReWright} are involved to reduce communication costs. Nonetheless, sparsified or delayed \algname{SGD}-updates are not unbiased anymore and require additional analysis \citep{stich2020error,BezHorRichSaf}.

Zeroth-order methods are often utilized when there is no access to unbiased gradients, e.g., for optimization of black-box functions \citep{NestSpok}, or for finding adversarial examples in deep learning \citep{MoosFawziFrossard,ChenZhangSharmaYiHsieh}. Many zeroth-order training methods exploit biased gradient oracles \citep{NestSpok, LiuKailChenTingChangAmini}. Various other techniques as smoothing, proximate updates and preconditioning operate with inexact gradient estimators \citep{DAspremont, SchmidtRouxBach, DevolderGlineurNesterov, TappendenRichtarikGondzio, KariStichJaggi}.

The aforementioned applications illustrate that \algname{SGD} can converge even if it performs \textit{biased} gradient updates, provided that certain ``regularity'' conditions are satisfied by the corresponding gradient estimators~\citep{BotCurNoce, AjallStich, BezHorRichSaf, CondYiRich}. Moreover, biased estimators may show better performance over their unbiased equivalents in certain settings \citep{BezHorRichSaf}.

In this work we study convergence properties and worst-case complexity bounds of stochastic gradient descent (\algname{SGD}) with a {\em biased} gradient estimator (\algname{BiasedSGD}; see Algorithm \ref{alg:SGD}) for solving general optimization problems of the form
\begin{equation*}
	\min_{x\in\mathbb{R}^d} f(x),
\end{equation*}
where the function $f:\mathbb{R}^d\to \mathbb{R}$ is possibly nonconvex, satisfies several smoothness and regularity conditions. 

\begin{assumption}\label{ass_smooth}
	Function $f$ is differentiable, $L$-smooth (i.e., $\norm{\nabla f(x)-\nabla f(y)}\leq L \norm{x-y}$ for all $x,y\in \R^d$), and bounded from below by $f^{*}\in\mathbb{R}.$
\end{assumption}
We write $g(x)$ for the gradient estimator, which is biased (i.e., $\mathbb{E}\left[g(x)\right]$ is not equal to $\nabla f(x),$ $\mathbb{E}\left[\cdot\right]$ stands for the expectation with respect to the randomness of the algorithm), in general. By a gradient estimator we mean a (possibly random) mapping $g:\mathbb{R}^d\to\mathbb{R}^d$ with some constraints. We denote by $\gamma$ an appropriately chosen learning rate, and $x^0\in\mathbb{R}^d$ is a starting point of the algorithm.
\begin{algorithm}
	\caption{Biased Stochastic Gradient Descent (\algname{BiasedSGD})}\label{alg:SGD}
	\begin{algorithmic}[1]
		\Require initial point $x^0\in\mathbb{R}^d;$ learning rate $\gamma > 0$
		\For{$t = 0, 1, 2, \ldots$}\do\\
		\State Construct a {\blue(possibly biased)} estimator $g^t\eqdef g(x^t)$ of the gradient $\nabla f(x^t)$
		\State Compute $x^{t+1} = x^t - \gamma g^t$
		\EndFor
	\end{algorithmic}
\end{algorithm}

In the strongly convex case, $f$ has a unique global minimizer which we denote by $x^{*},$ and $f(x^{*})=f^{*}.$ In the nonconvex case, $f$ can have many local minima and/or saddle points. It is theoretically intractable to solve this problem to global optimality \citep{NemirovskyYudin}. 
Depending on the assumptions on $f$, and given some error tolerance $\varepsilon>0,$ will seek to find a random vector $x\in\mathbb{R}^d$ such that one of the following inequalities holds: i) $\mathbb{E}\left[f(x) - f^{*}\right]\leq\varepsilon$ (convergence in function values); ii) $\mathbb{E}\left\|x-x^{*}\right\|^2\leq \varepsilon\left\|x^0-x^{*}\right\|^2$ (iterate convergence); iii) $\mathbb{E}\left\|\nabla f(x)\right\|^2\leq\varepsilon^2$ (gradient norm convergence).

\section{Sources of bias}\label{section_sources_of_bias_main} Practical applications of \algname{SGD} typically involve the training of supervised machine learning models via empirical risk minimization~\citep{shai_book}, which leads to optimization problems of a finite-sum structure:
\begin{equation}\label{eq_finite_sum}
	\squeeze	f(x) = \frac{1}{n}\sum \limits_{i=1}^{n}f_i(x).
\end{equation}
In the single-machine setup, $n$ is the number of data points, $f_i(x)$ represents the loss of a model $x$ on a data point $i.$ In this setting, data access is expensive, $g(x)$ is usually constructed with {\em subsampling} techniques such as minibatching and importance sampling. Generally, a subset $S\subseteq [n]$ of examples is chosen, and subsequently $g(x)$ is assembled from the information stored in the gradients of $\nabla f_i(x)$ for $i\in S$ only. This leads to estimators of the form $g(x) = \sum_{i \in S}v_i\nabla f_i(x),$ where $v_i$ are random variables typically designed to ensure the unbiasedness~\citep{gower2019sgd}. In practice, points might be sampled with unknown probabilities. In this scenario, a reasonable strategy to estimate the gradient is to take an average of all sampled $\nabla f_i.$ In general, the estimator obtained is biased, and such sources of bias can be characterized as arising from a lack of information about the subsampling strategy.

In the distributed setting, $n$ represemts the number of machines, and each $f_i$ represents the loss of model $x$ on all the training data stored on machine $i.$ Since communication is typically very expensive, modern gradient-type methods rely on various gradient compression mechanisms that are usually randomized. Given an appropriately chosen compression map $\mathcal{C}:\mathbb{R}^d\to\mathbb{R}^d,$ the local gradients $\nabla f_i(x)$ are first compressed to $\mathcal{C}_i\left(\nabla f_i(x)\right),$ where $\mathcal{C}_i$ is an independent realization of $\mathcal{C}$ sampled by machine $i$ in each iteration, and subsequently communicated to the master node, which performs aggregation (typically averaging). This gives rise to \algname{SGD} with the gradient estimator of the form
\begin{equation}\label{eq_distibuted_compression}
	\squeeze	g(x) = \frac{1}{n}\sum\limits_{i=1}^{n}\mathcal{C}_i\left(\nabla f_i(x)\right).
\end{equation}
Many important compressors performing well in practice are of biased nature (e.g., Top-$k,$ see~Def.~\ref{def_top_ell}), which, in general, makes $g(x)$ biased as well.

Biased estimators are capable of absorbing useful information in certain settings, e.g., in the heterogeneous data regime. Unbiased estimators have to be random, otherwise they are equal to the identity mapping. However, greedy deterministic gradient estimators such as  Top-$k$ often lead to better practical  performance. In \citep[Section 4]{BezHorRichSaf} the authors show an advantage of the Top-$k$ compressor over its randomized counterpart Rand-$k$ when the coordinates of the vector that we wish to compress are distributed uniformly or exponentially.  In practice, deterministic biased compressors are widely used for low precision training, and exhibit great performance \citep{alistarh2018sparse, BezHorRichSaf}.

\begin{figure}[t]
	\centering
	\footnotesize
	\begin{tikzcd}[node distance=2cm]
		\node (contr) [node] {
			\begin{tabular}{c}
				\text{\hyperlink{CON}{CON}}\\
				Asm \ref{ass_third_set}
			\end{tabular}
		};
		\node (bvd) [node, right of=contr, xshift=1.6cm, yshift=2.0cm] {
			\begin{tabular}{c}
				\text{\hyperlink{BVD}{\text{BVD}}}\\
				Asm \ref{ass_BV}
			\end{tabular}
		};
		\node (breq)[node, right of=contr, xshift=1.6cm]{
			\begin{tabular}{c}
				\text{\hyperlink{BREQ}{\text{BREQ}}}\\
				Asm \ref{ass_breq}
			\end{tabular}
		};
		\node (sg1)[node, right of=contr, xshift=5.2cm]{
			\begin{tabular}{c}
				\text{\hyperlink{SG1}{\text{SG1}}}\\
				Asm \ref{ass_first_set}
			\end{tabular}
		};
		\node (sg2)[node, right of=contr, xshift=5.2cm, yshift=-2.0cm]{
			\begin{tabular}{c}
				\text{\hyperlink{SG2}{\text{SG2}}}\\
				Asm \ref{ass_second_set}\\
			\end{tabular}
		};
		\node (bnd)[node, right of=contr, xshift=5.2cm, yshift=4cm]{
			\begin{tabular}{c}
				\text{\hyperlink{BND}{\text{BND}}}\\
				Asm \ref{ass_stich_decomposition}
			\end{tabular}
		};
		\node (fsml)[node, right of=contr, xshift=9.2cm]{
			\begin{tabular}{c}
				\text{\hyperlink{FSML}{FSML}}\\
				Asm \ref{ass_first_and_second_mmt_limits}
			\end{tabular}
		};
		\node (abc)[node, right of=contr, xshift=9.2cm, yshift=4cm]{
			\begin{tabular}{c}
				\text{\hyperlink{Biased ABC}{\text{Biased ABC}}}\\
				Asm \ref{ass_scalar_ABC}
			\end{tabular}
		};
		\node (abs) [node, right of=contr, xshift=1.6cm, yshift=6cm]{
			\begin{tabular}{c}
				\text{\hyperlink{ABS}{\text{ABS}}}\\
				Asm \ref{ass_abs_compr}
			\end{tabular}
		};
		
		\draw [-{Latex[length=3mm, width=2mm]}] (abs) -| node[anchor=north west]{\text{Thm \ref{thm_informal_abc}--\ref{item_bnd_from_ac}}}(bnd);
		\draw [-{Latex[length=3mm, width=2mm]}] (contr) |- node[anchor=north west] {\text{Thm \ref{thm_informal_abc}--\ref{item_bvd_from_contr}}} (bvd);
		\draw [-{Latex[length=3mm, width=2mm]}] (bvd) |- node[anchor=north west] {\text{Thm \ref{thm_informal_abc}--\ref{item_bnd_from_bvd}}} (bnd);
		\draw [{Latex[length=3mm, width=2mm]}-{Latex[length=3mm, width=2mm]}] (sg1) -- node[anchor=east]{\text{Thm\ref{thm_informal_abc}--\ref{item_first_two_equiv}}} (sg2);
		\draw [-{Latex[length=3mm, width=2mm]}] (fsml) -- node[anchor=east]{\text{Thm\ref{thm_informal_abc}--\ref{item_abc_from_fsml}}} (abc);
		\draw [-{Latex[length=3mm, width=2mm]}] (breq) -- node[anchor=north]{\text{Thm \ref{thm_informal_abc}--\ref{item_first_from_breq}}}(sg1);
		\draw [-{Latex[length=3mm, width=2mm]}] (bvd) -| node[anchor=north east]{\text{Thm \ref{thm_informal_abc}--\ref{item_first_from_bvd}}}(sg1);
		\draw [-{Latex[length=3mm, width=2mm]}] (sg1) -- node[anchor=north]{\text{Thm \ref{thm_informal_abc}--\ref{item_fsml_from_first}}}(fsml);
		\draw [-{Latex[length=3mm, width=2mm]}] (bnd) -- node[anchor=north]{\text{Thm \ref{thm_informal_abc}--\ref{item_abc_from_bnd}}}(abc);
		\draw [dashed,thick,blue] (contr) -- node[anchor=north]{\text{Thm \ref{thm_diagram_counterexamples}--\ref{item_no_implication_breq_contractive}}}(breq);
		\path [blue, in=-152, out=28, thick, dashed] (breq) edge node[anchor=south west]{\text{Thm \ref{thm_diagram_counterexamples}--\ref{item_no_implication_breq_stich}}}(bnd);
		\path [blue, in=-25, out=180, thick, dashed] (fsml) edge node[anchor=east]{\text{Thm \ref{thm_diagram_counterexamples}--\ref{item_no_implication_first_bnd}}}(bnd);
		\path [blue, in=152, out=-152, thick, dashed](abs) edge node[anchor=west]{\text{Thm \ref{thm_diagram_counterexamples}--\ref{item_no_implication_breq_abs}}}(breq);
		\path [blue, in=97, out=-180, thick, dashed](abs) edge node[anchor=east]{\text{Thm \ref{thm_diagram_counterexamples}--\ref{item_no_implicatiom_contractive_abs}}}(contr);
		\path [blue, in=90, out=-90, thick, dashed](abs) edge node[anchor=north]{\text{Thm \ref{thm_diagram_counterexamples}--\ref{item_no_implication_fsml_abs}}}(fsml);
	\end{tikzcd}
	\caption{Assumption hierarchy. A single arrow indicates an implication and an absence of a reverse implication. The implications are transitive. A dashed line indicates a mutual abscence of implications. Our newly proposed assumption \protect\hyperlink{Biased ABC}{Biased ABC} is the most general one.}
	\label{fig_diagram}
\end{figure}
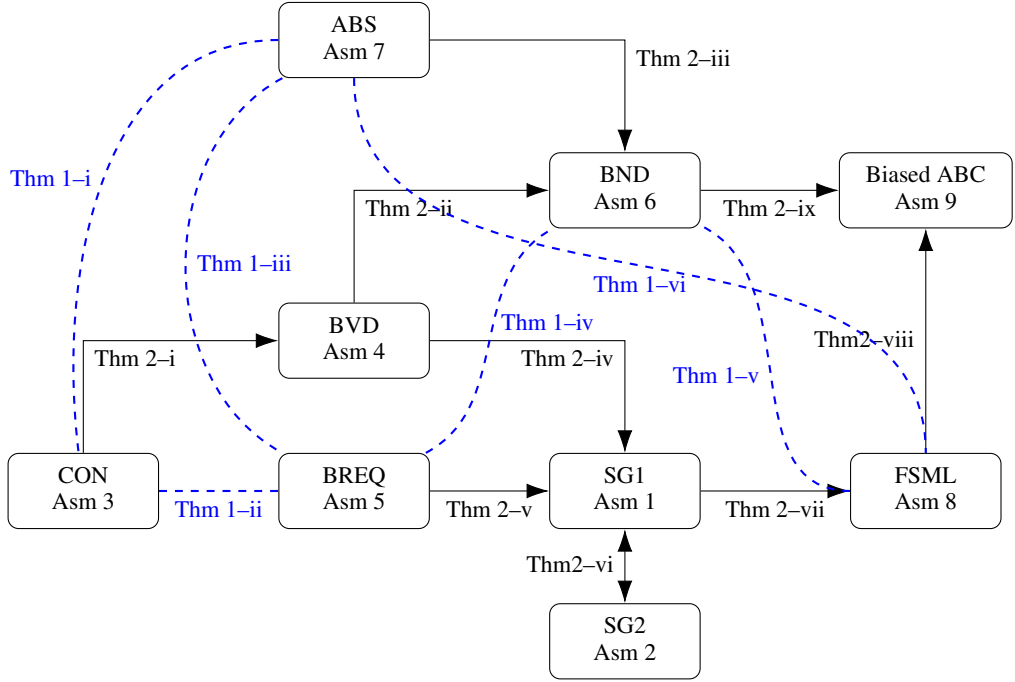

\section{Contributions}
The most commonly used assumptions for analyzing \algname{SGD} with biased estimators take the form of various structured bounds on the first and the second moments of $g(x).$ We argue that assumptions proposed in the literature are often too strong, and may be unrealistic as they do not fully capture how bias and randomness in $g(x)$ arise in practice. In order to retrieve meaningful theoretical insights into the operation of \algname{BiasedSGD}, it is important to model the bias and randomness both correctly, so that the assumptions we impart are provably satisfied, and accurately, so as to obtain as tight bounds as possible. Our work is motivated by the need of a more accurate and informative analysis of \algname{BiasedSGD} in the strongly convex and nonconvex settings, which are problems of key importance in optimization research and deep learning. Our results are generic and cover both subsampling and compression-based estimators, among others.

The key contributions of our work are:

\phantom{XX} $\bullet$ Inspired by recent developments in the analysis of \algname{SGD} in the nonconvex setting \citep{khaled2022better}, the analysis of \algname{BiasedSGD}  \citep{BotCurNoce, AjallStich}, the analysis of biased compressors \citep{BezHorRichSaf}, we propose a new assumption, which we call \hyperlink{Biased ABC}{Biased ABC}, for modeling the first and the second moments of the stochastic gradient.

\phantom{XX} $\bullet$ We show in Section \ref{section_ass_weakest} that \hyperlink{Biased ABC}{Biased ABC} is the weakest, and hence the most general, among all assumptions in the existing literature on \algname{BiasedSGD}  we are aware of (see Figure \ref{fig_diagram}), including concepts such as Contractive (\hyperlink{CON}{CON}) \citep{cordonnier2018convex, Stich-EF-NIPS2018, BezHorRichSaf}, Absolute  (\hyperlink{ABS}{ABS}) \citep{sahu2021rethinking}, Bias-Variance Decomposition~(\hyperlink{BVD}{BVD}) \citep{CondYiRich}, Bounded Relative Error Quantization (\hyperlink{BREQ}{BREQ}) \citep{khirirat2018gradient}, Bias-Noise~Decomposition~(\hyperlink{BND}{BND}) \citep{AjallStich}, Strong~Growth~1~(\hyperlink{SG1}{SG1}) and Strong~Growth~2~(\hyperlink{SG2}{SG2}) \citep{BezHorRichSaf}, and First and Second Moment Limits~(\hyperlink{FSML}{FSML}) \citep{BotCurNoce} estimators.

\phantom{XX} $\bullet$ We prove that unlike the existing assumptions, which implicitly assume that the bias comes from either perturbation or compression, \hyperlink{Biased ABC}{Biased ABC} also holds in settings such as subsampling.

\phantom{XX} $\bullet$ We recover the optimal rates for general smooth nonconvex problems and for problems under the P\L~condition in the unbiased case and prove that these rates are also optimal in the biased case.

\phantom{XX} $\bullet$ In the strongly convex case, we establish a similar convergence result in terms of iterate norms as in \citep{HuSeiLes}, however, under milder assumptions and not only for the classical version of \algname{SGD}. Our proof strategy is very different and much simpler.

\section{Existing models of biased gradient estimators}\label{sect_existing_models}

Since application of a gradient compressor to the gradient constitutes a gradient estimator, below we often reformulate known assumptions and results obtained for biased compressors in the more general form of biased gradient estimators. \citet{BezHorRichSaf} analyze \algname{SGD} under the assumption that $f$ is $\mu$-strongly convex, and propose three different assumptions for compressors.
\begin{assumption}[Strong Growth 1, \hypertarget{SG1}{SG1} -- \citet{BezHorRichSaf}]\label{ass_first_set}
	Let us say that $g(x)$ belongs to a set $\mathbb{B}^{1}(\alpha,\beta)$ of biased gradient estimators, if, for some $\alpha,\beta>0,$ for every $x\in\mathbb{R}^d,$ $g(x)$ satisfies
	\begin{equation}\label{eq_first_set}
		\alpha\left\|\nabla f(x)\right\|^2\leq\mathbb{E}\left[\left\|g(x)\right\|^2\right]\leq\beta\langle\mathbb{E}\left[g(x)\right],\nabla f(x)\rangle.
	\end{equation}
\end{assumption}
\begin{assumption}[Strong Growth 2, \hypertarget{SG2}{SG2} -- \citet{BezHorRichSaf}]\label{ass_second_set}
	Let us say that $g(x)$ belongs to a set $\mathbb{B}^{2}(\tau, \beta)$ of biased gradient estimators, if, for some $\tau,\beta>0,$ for every $x\in\mathbb{R}^d,$ $g(x)$ satisfies
	\begin{equation}\label{eq_second_set}
		\squeeze	\max\left\lbrace \tau \|\nabla f(x)\|^2,\frac{1}{\beta}\mathbb{E}\left[\left\|g(x)\right\|^2\right] \right\rbrace \leq \langle \mathbb{E}[g(x)], \nabla f(x) \rangle.
	\end{equation}
\end{assumption}
Note that each of Assumptions \ref{ass_first_set} and \ref{ass_second_set} imply
\begin{equation}\label{eq_implication}
	\mathbb{E}\left[\left\|g(x)\right\|^2\right]\leq\beta^2\left\|\nabla f(x)\right\|^2.
\end{equation}
\begin{assumption}[Contractive, \hypertarget{CON}{CON} -- \citet{BezHorRichSaf}]\label{ass_third_set}
	Let us say that $g(x)$ belongs to a set $\mathbb{B}^{3}(\delta)$ of biased gradient estimators, if, for some $\delta>0,$ for every $x\in\mathbb{R}^d,$ $g(x)$ satisfies
	\begin{equation}\label{eq_contractive_biased}
		\squeeze	\mathbb{E}\left[\left\|g(x) - \nabla f(x)\right\|^2\right]\leq\left(1 - \frac{1}{\delta}\right)\left\|\nabla f(x)\right\|^2.
	\end{equation}
\end{assumption}
The last condition is an abstraction of the contractive compression property (see Appendix~\ref{apx_contractive_relations}). \citet{CondYiRich} introduce another assumption for biased compressors, influenced by a bias-variance decomposition equation for the second moment:
\begin{equation}\label{eq_bv_decomposition}
	\mathbb{E}\left[\left\|g(x) - \nabla f(x)\right\|^2\right] = \left\|\mathbb{E}[g(x)] - \nabla f(x)\right\|^2 + \mathbb{E}\left[\left\|g(x) - \mathbb{E}[g(x)]\right\|^2\right].
\end{equation}
Let us write the assumption itself.
\begin{assumption}[Bias-Variance Decomposition, \hypertarget{BVD}{BVD} -- \citet{CondYiRich}]\label{ass_BV}
	Let 
	$0\leq\eta\leq 1,$ $\xi\geq 0,$ for all $x\in\mathbb{R}^d,$ the gradient estimator $g(x)$ satisfies
	\begin{eqnarray}
		\left\|\mathbb{E}[g(x)] - \nabla f(x)\right\|^2 & \leq & \eta\left\|\nabla f(x)\right\|^2,\label{eq_BV_bias} \\
		\mathbb{E}\left[\left\|g(x) - \mathbb{E}[g(x)]\right\|^2\right] & \leq & \xi \left\|\nabla f(x)\right\|^2.\label{eq_BV_variance}
	\end{eqnarray}
\end{assumption}

\citet{khirirat2018gradient} proposed another assumption on deterministic compressors. 
\begin{assumption}[Bounded Relative Error Quantization, \hypertarget{BREQ}{BREQ} -- \citet{khirirat2018gradient}]\label{ass_breq}
	For all $x\in\mathbb{R}^d,$ for any $\rho,\zeta\geq0,$
	\begin{eqnarray}
		\langle g(x), \nabla f(x)\rangle & \geq & \rho \left\|\nabla f(x)\right\|^2,\label{eq_breq_scalar}\\
		\left\|g(x)\right\|^2 &\leq & \zeta\left\|\nabla f(x)\right\|^2.\label{eq_breq_second_mmt}
	\end{eqnarray}
\end{assumption}

The restriction below was imposed on the gradient estimator $g(x)$ by~\citet{AjallStich}. For the purpose of clarity, we rewrote it in the notation adopted in our paper. We refer the reader to Appendix~\ref{apx_equiv_stich_reformulation} for the proof of equivalence of these two definitioins.
\begin{assumption}[Bias-Noise Decomposition, \hypertarget{BND}{BND} -- \citet{AjallStich}]\label{ass_stich_decomposition} 
	Let $M,\sigma^2,\varphi^2$ be nonnegative constants, and let $0\leq m<1.$ For all $x\in\mathbb{R}^d,$ $g(x)$ satisfies
	\begin{eqnarray}
		\mathbb{E}\left[\left\|g(x) - \mathbb{E}\left[g(x)\right]\right\|^2\right] &\leq & M\left\|\mathbb{E}\left[g(x)\right]\right\|^2+\sigma^2,\label{eq_noise_stich_simplified} \\
		\left\|\mathbb{E}\left[g(x) \right] - \nabla f(x)\right\|^2 & \leq & m\left\|\nabla f(x)\right\|^2+\varphi^2.\label{eq_bias_stich_simplified}
	\end{eqnarray}
\end{assumption}

The following assumption was introduced by \citet{sahu2021rethinking} (see also the work of \citet{danilova2022distributed}).
\begin{assumption}[Absolute Estimator, \hypertarget{ABS}{ABS} -- \citet{sahu2021rethinking}]\label{ass_abs_compr}
	For all $x\in\mathbb{R}^d,$ there exists $\Delta \geq 0$ such that
	\begin{equation}\label{eq_abs_compr}
		\mathbb{E}\left[\|g(x)-\nabla f(x)\|^2\right] \leq \Delta^2.
	\end{equation}
\end{assumption}
This condition is tightly related to the contractive compression property (see Appendix~\ref{apx_absolute_compression_relations}). Further, \citet{BotCurNoce} proposed the following restriction on a stochastic gradient estimator.
\begin{assumption}[First and Second Moment Limits, \hypertarget{FSML}{FSML} -- \citet{BotCurNoce}]\label{ass_first_and_second_mmt_limits}
	There exist constants $0<q\leq u,$ $U\geq~0,$ $Q\geq 0,$ such that, for all $x\in\mathbb{R}^d,$
	\begin{eqnarray}
		\langle\nabla f(x), \Exp{g(x)}\rangle &\geq & q \left\|\nabla f(x)\right\|^2, \label{eq_fsml_scalar}\\ 
		\left\|\Exp{g(x)}\right\| &\leq & u \left\|\nabla f(x)\right\|, \label{eq_fsml_first_mmt} \\
		\Exp{\left\|g(x) - \Exp{g(x)}\right\|^2} &\leq & U\left\|\nabla f(x)\right\|^2 + Q.\label{eq_fsml_variance}
	\end{eqnarray}
\end{assumption}

Our first theorem, described informally below and stated and proved formally in the appendix, provides required counterexamples of problems and estimators for the diagram in Figure~\ref{fig_diagram}. 
\begin{theorem}\label{thm_diagram_counterexamples}
	$\mathrm{(Informal)}$ The assumptions connected by dashed lines in Figure~\ref{fig_diagram} are mutually non-implicative.
\end{theorem}

The result says that some pairs of assumptions are in a certain sense unrelated: none implies the other, and vice versa. In the next section, we introduce a new assumption, and provide deeper connections between all assumptions.

\section{New approach: biased ABC assumption}
\subsection{Brief history}
Several existing restrictions on the first moment of the estimator were very briefly routlined in the previous section (see \eqref{eq_first_set}, \eqref{eq_BV_bias}, \eqref{eq_breq_scalar}, \eqref{eq_bias_stich_simplified}, \eqref{eq_fsml_scalar}). \citet{khaled2022better} recently introduced a very general and accurate Expected Smoothness assumption (we will call it the ABC-assumption in this paper) on the second moment of the unbiased estimator. We generalize the restrictions \eqref{eq_first_set}, \eqref{eq_breq_scalar}, \eqref{eq_fsml_scalar} on the first moment and combine them with the $\text{ABC-assumption}$ to develop our \hyperlink{Biased ABC}{Biased ABC} framework.
\begin{assumption}[$\hypertarget{Biased ABC}{\mathrm{\textbf{Biased ABC}}}$]\label{ass_scalar_ABC} There exist constants $A,B,C, b,  c\geq 0$  such that  the gradient estimator $g(x)$ for every $x\in\mathbb{R}^d$ satisfies\footnote{In \citep{khaled2022better}, the ``ABC assumption'' was introduced in the unbiased case. However, we aim to establish theory for biased estimators. If we simply remove \eqref{eq_scalar_prod}, then $g(x)=-\nabla f(x)$ satisfies \eqref{eq_ABC} with $A=0,$ $B=1,$ $C=0,$ yet \algname{BiasedSGD} clearly diverges in general.}
	\begin{eqnarray}
		\langle\nabla f(x), \mathbb{E}[g(x)]\rangle &\geq &  b \left\|\nabla f(x)\right\|^2 -   c, \label{eq_scalar_prod}\\	\
		\mathbb{E}\left[\left\|g(x)\right\|^2\right] & \leq & 2A \left(f(x) - f^{*}\right) + B\left\|\nabla f(x)\right\|^2 + C.\label{eq_ABC}
	\end{eqnarray}
\end{assumption}

\subsection{Biased ABC as the weakest assumption}\label{section_ass_weakest}
As discussed in Section \ref{sect_existing_models}, there exists a Zoo of assumptions on the stochastic gradients in  literature on \algname{BiasedSGD}. Our second theorem, described informally below and stated and proved formally in the appendix, says that our new \hyperlink{Biased ABC}{Biased ABC} assumption is the least restrictive of all the assumptions reviewed in 
Section~\ref{sect_existing_models}.

\begin{theorem}\label{thm_informal_abc}
	$\mathrm{(Informal)}$ Assumption~\ref{ass_scalar_ABC} (\hyperlink{Biased ABC}{Biased ABC}) is the weakest among {Assumptions~\ref{ass_first_set}~--~\ref{ass_scalar_ABC}}.
\end{theorem}

Inequality \eqref{eq_BV_bias} of \hyperlink{BVD}{BVD} or inequality \eqref{eq_bias_stich_simplified} of \hyperlink{BND}{BND} show that one can impose the restriction on the first moment by bounding the norm of the bias. We choose inequality \eqref{eq_scalar_prod} that restrains the scalar product between the estimator and the gradient on purpose: this approach turns out to be more general on its own. In the proof of Theorem~\ref{thm_informal_abc}-\ref{item_abc_from_bnd} (see~\eqref{eq_equiv_stich_scalar}~and~\eqref{eq_beat_stich}) we show that \eqref{eq_bias_stich_simplified} implies \eqref{eq_scalar_prod}. Below we show the existence of a counterexample that the reverse implication does not hold.
\begin{claim}\label{claim_sampling_counterex_no_stich_but_abc}
	There exists a finite-sum minimization problem for which a gradient estimator that satisfies inequality \eqref{eq_scalar_prod} of Assumption~\ref{ass_scalar_ABC} does not satisfy inequality \eqref{eq_bias_stich_simplified} of Assumption~\ref{ass_stich_decomposition}.
\end{claim}

Relationships among Assumptions~\ref{ass_first_set}--\ref{ass_scalar_ABC} are depicted in Figure~\ref{fig_diagram} based on the results of Theorem~\ref{thm_diagram_counterexamples} and Theorem~\ref{thm_informal_abc}. In Table~\ref{tab_assns_in_our_frame} we provide a representation of each of Assumptions~\ref{ass_first_set}~--~\ref{ass_first_and_second_mmt_limits} in our \hyperlink{Biased ABC}{Biased ABC} framework (based on the results of Theorem~\ref{thm_assumptions_in_our_framework}). Note that the constants in Table~\ref{tab_assns_in_our_frame} are too pessimistic: given the estimator satisfying one of these assumptions, direct computation of constants in \hyperlink{Biased ABC}{Biased ABC} scope for it might lead to much more accurate results. In Table~\ref{tab_estimators_short} we give a description of popular gradient estimators in terms of the \hyperlink{Biased ABC}{Biased ABC} framework. Finally, in Table~\ref{tab_estimators_in_assumptions_short} we list several  popular estimators and indicate which of Assumptions~\ref{ass_first_set}--\ref{ass_scalar_ABC} they satisfy.

\begin{table*}[t]
	\centering
	\footnotesize
	\begin{threeparttable}
		\begin{tabular}{|c|c c c c c|}
			\hline
			\bf Assumption & $A$ & $B$ & $C$ & $ b $&$  c$ \\ 
			\hline
			\begin{tabular}{c}
				{{\tiny Asm \ref{ass_first_set} \; (\hyperlink{SG1}{\bf SG1})}} \scriptsize{\citep{BezHorRichSaf}}
			\end{tabular}				& $0$ & $\beta^2$ & $0$ &  $\frac{\alpha}{\beta}$& $0$ \\ 					
			\hline
			\begin{tabular}{c}
				{{\tiny Asm \ref{ass_second_set}\; (\hyperlink{SG2}{\bf SG2})}} \scriptsize{\citep{BezHorRichSaf}}
			\end{tabular}				& $0$ & $\beta^2$ & $0$ & $\tau$ & $0$ \\ 					
			\hline
			\begin{tabular}{c}
				{\tiny Asm \ref{ass_third_set}\; (\hyperlink{CON}{\bf CON})} \scriptsize{\citep{BezHorRichSaf}}
			\end{tabular}				 & $0$ & $2\left(2-\frac{1}{\delta}\right)$  & $0$ & $\frac{1}{2\delta}$ & $0$\\ 					
			\hline
			\begin{tabular}{c}
				{{\tiny Asm \ref{ass_BV}\; (\hyperlink{BVD}{\bf BVD})}} \scriptsize{\citep{CondYiRich}}
			\end{tabular}				 &  $0$ &  $2(1 + \xi + \eta)$ & $0$  & $\frac{1-\eta}{2}$  & $0$  \\ 					
			\hline
			\begin{tabular}{c}
				{{\tiny Asm \ref{ass_breq}\; (\hyperlink{BREQ}{\bf BREQ})}} \scriptsize{\citep{khirirat2018gradient}}
			\end{tabular}				 & $0$ & $\zeta$ & $0$ & $\rho$ & $0$ \\ 					
			\hline
			\begin{tabular}{c}
				{{\tiny Asm \ref{ass_stich_decomposition}\; (\hyperlink{BND}{\bf BND})}} \scriptsize{\citep{AjallStich}}
			\end{tabular}				&$0$ & $2(M+1)(m+1)$  & $2(M+1)\varphi^2 +  \sigma^2$  & $\frac{1-m}{2}$  & $\frac{\varphi^2}{2}$ \\ 					
			\hline
			\begin{tabular}{c}
				{{\tiny Asm \ref{ass_abs_compr}\; (\hyperlink{ABS}{\bf ABS})}} \scriptsize{\citep{sahu2021rethinking} }
			\end{tabular}				 & $0$  & $2$  & $2\Delta^2$ & $\frac{1}{2}$ & $\frac{\Delta^2}{2}$ \\ 					
			\hline
			\begin{tabular}{c}
				{{\tiny Asm \ref{ass_first_and_second_mmt_limits}\; (\hyperlink{FSML}{\bf FSML})}} \scriptsize{\citep{BotCurNoce}}
			\end{tabular}				 & $0$ & $U+u^2$ & $Q$ & $q$ & $0$ \\ 					
			\hline
		\end{tabular}
	\end{threeparttable}
	\caption{Summary of known assumptions on biased stochastic gradients. Estimators satisfying any of them, belong to our general \protect\hyperlink{Biased ABC}{Biased ABC} framework with parameters $A,$ $B,$ $C,$ $ b$ and $  c$ provided in this table. For proofs, we refer the reader to Theorem~\ref{thm_assumptions_in_our_framework}.}
	\label{tab_assns_in_our_frame}    
\end{table*}

\begin{table*}[h]
	\centering
	\footnotesize
	\begin{threeparttable}
		\begin{tabular}{|c|c|c c c c c|}
			\hline
			\bf Estimator  & \bf Def & $A$ & $B$ & $C$ & $ b $&$  c$ \\ 
			\hline
			\begin{tabular}{c}
				{\tiny \bf Biased independent sampling}\\ \scriptsize{\tiny [This paper]}
			\end{tabular}	& Def.\ \ref{def_biased_sampling_no_replacement} & $\frac{\max_i\{L_i\}}{\min_{i}{p_i}}$ & $0$ & $2A\Delta^{*} + s^2$ & $\min \limits_i\left\lbrace p_i\right\rbrace$ & $0$\\ 					
			\hline
			\begin{tabular}{c}
				{{\tiny \bf Top-$k$}}\\ \scriptsize{\citep{aji2017sparse}}
			\end{tabular}				 & Def.\ \ref{def_top_ell} & $0$ & $1$ & $0$ & $\frac{k}{d}$ & $0$\\ 					
			\hline
			\begin{tabular}{c}
				{{\tiny \bf Rand-$k$}}\\ \scriptsize{\cite{Stich-EF-NIPS2018}}
			\end{tabular}				 & Def. \ref{def_random_l} & $0$  & $\frac{d}{k}$ & $0$ & $1$ & $0$ \\ 					
			\hline
			\begin{tabular}{c}
				{{\tiny \bf Biased Rand-$k$}}\\ \scriptsize{\citep{BezHorRichSaf}}
			\end{tabular}				 & Def. \ref{def_biased_random_l} & $0$  & $\frac{k}{d}$ & $0$ & $\frac{k}{d}$ & $0$ \\ 					
			\hline		
			\begin{tabular}{c}
				{{\tiny \bf Adaptive random sparsification}}\\ \scriptsize{\citep{BezHorRichSaf}}
			\end{tabular}				 & Def. \ref{def_adaptive_random_sparsification} & $0$  & $1$ & $0$ & $\frac{1}{d}$ & $0$ \\ 	
			\hline
			\begin{tabular}{c}
				{{\tiny \bf General unbiased rounding}}\\ \scriptsize{\citep{BezHorRichSaf}}
			\end{tabular}				 & Def. \ref{def_general_unbiased_rounding} & $0$  & \tiny $\sup\limits_{k\in\mathbb{Z}}\frac{a_k^2+a_{k+1}^2}{a_ka_{k+1}}+\frac{1}{2}$ & $0$ & $1$ & $0$ \\ 	
			\hline
			\begin{tabular}{c}
				{{\tiny \bf Natural compression}}\\ \scriptsize{\citep{CNAT}}
			\end{tabular}				 & Def. \ref{def_natural_compression} & $0$  & $\frac{9}{8}$ & $0$ & $1$ & $0$ \\ 	
			\hline
			\begin{tabular}{c}
				{{\tiny \bf Scaled integer rounding}}\\ \scriptsize{\citep{SapCanHoNelKalKinKrisMoshPorRich}}
			\end{tabular}				 & Def. \ref{def_scaled_rounding} & $0$  & $2$ & $\frac{2d}{\chi^2}$ & $\frac{1}{2}$ & $\frac{d}{2\chi^2}$ \\ 	
			\hline
		\end{tabular}
	\end{threeparttable}
	\caption{Summary of popular estimators with respective parameters $A$, $B$, $C$, $ b$ and $  c,$ satisfying our general \protect\hyperlink{Biased ABC}{Biased ABC} framework. Constants $L_i$ are from Assumption~\ref{ass_smooth_functionwise}, $\Delta^{*}$ is defined in \eqref{eq_def_delta_star}. For more estimators, see Table~\ref{tab_estimators_full}.} 
	\label{tab_estimators_short}  
\end{table*}

\begin{table*}[h]
	\centering
	\footnotesize
	\begin{threeparttable}
		\begin{tabular}{|l|c c c c c c c c c|}
			\hline
			\bf Estimator $\backslash$ \bf Assumption  & A\ref{ass_first_set} & A\ref{ass_second_set} & A\ref{ass_third_set} & A\ref{ass_BV} & A\ref{ass_breq} & A\ref{ass_stich_decomposition} & A\ref{ass_abs_compr} & A\ref{ass_first_and_second_mmt_limits}  & A\ref{ass_scalar_ABC} \\
			\hline
			\begin{tabular}{l}
				{\tiny \bf Biased independent sampling}
				\scriptsize{\tiny [This paper]}
			\end{tabular}				 & {\color{red}\ding{55}}  & {\color{red}\ding{55}}  & {\color{red}\ding{55}}  & {\color{red}\ding{55}}  & {\color{red}\ding{55}}  & {\color{red}\ding{55}}  & {\color{red}\ding{55}} & {\color{red}\ding{55}} & {\color{green}\checkmark}\\ 					
			\hline
			\begin{tabular}{l}
				{{\tiny \bf Top-$k$ sparsification}}
				\scriptsize{\citep{aji2017sparse}}
			\end{tabular}				 & {\color{green}\checkmark} & {\color{green}\checkmark} & {\color{green}\checkmark} & {\color{green}\checkmark} & {\color{green}\checkmark} & {\color{green}\checkmark} & {\color{red}\ding{55}} & {\color{green}\checkmark} & {\color{green}\checkmark} \\ 					
			\hline
			\begin{tabular}{l}
				{{\tiny \bf Rand-$k$}}
				\scriptsize{\citep{Stich-EF-NIPS2018}}
			\end{tabular}				 & {\color{green}\checkmark} & {\color{green}\checkmark}  & {\color{red}\ding{55}} & {\color{green}\checkmark} & {\color{red}\ding{55}} & {\color{green}\checkmark} & {\color{red}\ding{55}} & {\color{green}\checkmark} & {\color{green}\checkmark}\\ 					
			\hline
			\begin{tabular}{l}
				{{\tiny \bf Biased Rand-$k$}}
				\scriptsize{\citep{BezHorRichSaf}}
			\end{tabular}				 & {\color{green}\checkmark} & {\color{green}\checkmark} & {\color{green}\checkmark} & {\color{green}\checkmark} & {\color{red}\ding{55}} & {\color{green}\checkmark} & {\color{red}\ding{55}} & {\color{green}\checkmark} & {\color{green}\checkmark}\\ 					
			\hline
			\begin{tabular}{l}
				{{\tiny \bf Adaptive random sparsification}}
				\scriptsize{\citep{BezHorRichSaf}}
			\end{tabular}				 & {\color{green}\checkmark} & {\color{green}\checkmark} & {\color{green}\checkmark} & {\color{green}\checkmark} & {\color{red}\ding{55}} & {\color{green}\checkmark} & {\color{red}\ding{55}} & {\color{green}\checkmark} & {\color{green}\checkmark}\\ 					
			\hline
			\begin{tabular}{l}
				{{\tiny \bf General unbiased rounding}}
				\scriptsize{\citep{BezHorRichSaf}}
			\end{tabular}				 & {\color{green}\checkmark} & {\color{green}\checkmark} & {\color{red}\ding{55}} & {\color{green}\checkmark} & {\color{red}\ding{55}} & {\color{green}\checkmark} & {\color{red}\ding{55}}  & {\color{green}\checkmark} & {\color{green}\checkmark}\\ 					
			\hline
			\begin{tabular}{l}
				{{\tiny \bf Natural compression}}
				\scriptsize{\citep{CNAT}}
			\end{tabular}				 & {\color{green}\checkmark} & {\color{green}\checkmark} & {\color{green}\checkmark} & {\color{green}\checkmark} & {\color{red}\ding{55}}  & {\color{green}\checkmark} & {\color{red}\ding{55}}  & {\color{green}\checkmark} & {\color{green}\checkmark}\\ 					
			\hline
			\begin{tabular}{l}
				{{\tiny \bf Scaled integer rounding}}
				\scriptsize{\citep{SapCanHoNelKalKinKrisMoshPorRich}}
			\end{tabular}				 & {\color{red}\ding{55}}  & {\color{red}\ding{55}}   & {\color{red}\ding{55}}  & {\color{red}\ding{55}}   & {\color{red}\ding{55}}  & {\color{green}\checkmark} & {\color{green}\checkmark} & {\color{red}\ding{55}} & {\color{green}\checkmark}\\ 					
			\hline
		\end{tabular}
	\end{threeparttable}
	\caption{Coverage of popular estimators by known frameworks. For more estimators, see Table~\ref{tab_estimators_in_assumptions_full}.}
	\label{tab_estimators_in_assumptions_short}    
\end{table*}

\section{Convergence of biased SGD under the biased ABC assumption}
Convergence rates of theorems below are summarized in Table~\ref{tab_biased_vs_unbiased} and compared to their counterparts.
\subsection{General nonconvex case}\label{section_noncvx_main}
\begin{theorem}\label{thm_nonconvex} Let Assumptions \ref{ass_smooth} and \ref{ass_scalar_ABC} hold. Let $\delta^0\eqdef f(x^0) - f^{*}$, and choose the stepsize such that
$
	0<\gamma \leq \frac{ b}{LB}.
$
	Then the iterates $\{x^t\}_{t\geq 0}$ of \algname{BiasedSGD} (Algorithm \eqref{alg:SGD}) satisfy
	\begin{equation}\label{eq_convergence}
		\squeeze		\min \limits_{0\leq t \leq T-1}\mathbb{E}\left[\left\|\nabla f(x^t)\right\|^2\right] \leq  \frac{2\left(1+LA\gamma^2\right)^T}{ b\gamma T}\delta^0+\frac{LC\gamma}{ b} + \frac{  c}{ b}.
	\end{equation}

\end{theorem}
While one can notice the possibility of an exponential blow-up in \eqref{eq_convergence}, by carefully controlling the stepsize we still can guarantee the convergence of \algname{BiasedSGD}. In Corollaries~\ref{cor_noncvx_2}~and~\ref{cor_noncvx_3} (see the appendix) we retrieve the results of Theorem~2 and Corollary~1 from \citep{khaled2022better} for the unbiased case. In Corollary~\ref{cor_noncvx_stich} (see the appendix) we retrieve the result that is worse than that in \citep[Theorem~4]{AjallStich} by a multiplicative factor and an extra additive term, but under milder conditions (cf. \hyperlink{Biased ABC}{Biased ABC} and \hyperlink{BND}{BND} in Figure \ref{fig_diagram}; see also Claim~\ref{claim_sampling_counterex_no_stich_but_abc}). If we set $A =  c = 0,$ we recover the result of \citep[Theorem~4.8]{BotCurNoce} (see Corollary~\ref{cor_noncvx_bottou_follows} in the appendix).
\subsection{Convergence under P\L-condition}\label{section_pl_main}
One of the popular generalizations of strong convexity in the literature is the Polyak--\L ojasiewicz assumption \citep{PolyakAsn, KarNutSch, LeiHuLiTang}. First, we define this condition.
\begin{assumption}[Polyak--\L ojasiewicz]\label{ass_pl}
	There exists $\mu>0$ such that  $\left\|\nabla f(x)\right\|^2\geq 2\mu\left(f(x) - f^{*}\right),$ for all $ x\in\mathbb{R}^d$.
\end{assumption}
We now formulate a theorem that establishes the convergence of \algname{BiasedSGD} for functions satisfying this assumption and Assumption $\ref{ass_scalar_ABC}.$
\begin{theorem}\label{thm_noncvx_pl}
	Let Assumptions \ref{ass_smooth}, \ref{ass_scalar_ABC} and \ref{ass_pl} hold. Choose a stepsize such that
	\begin{equation}\label{eq_gamma_pl}
		\squeeze	0<\gamma<\min\left\lbrace\frac{\mu b}{L(A+\mu B)},\frac{1}{\mu b}\right\rbrace.
	\end{equation}
	Letting $\delta^0 \eqdef f(x^0) - f^{*}$, for every $T\geq 1,$ we have
	\begin{equation}\label{eq_convergence_pl}
		\squeeze	\mathbb{E}\left[f(x^T) - f^{*}\right] \leq \left(1 - \gamma\mu b\right)^T\delta^0+ \frac{LC\gamma}{2\mu b} + \frac{  c}{\mu  b}.
	\end{equation}
\end{theorem}
When $c=0,$ the last term in \eqref{eq_convergence_pl} disappears, and we recover the best known rates under the Polyak-- \L ojasiewicz condition \citep{KarNutSch}, but under milder conditions (see Corollary~\ref{cor_pl_2} in the appendix). Further, if we set $A=0,$ we obtain a result that is slightly weaker than the one obtained by \citet[Theorem~6]{AjallStich}, but under milder assumptions (cf. \hyperlink{Biased ABC}{Biased ABC} and \hyperlink{BND}{BND} in Figure \ref{fig_diagram}; see also Claim~\ref{claim_sampling_counterex_no_stich_but_abc}).

\subsection{Strongly convex case}\label{section_strongly_cvx_main}
\begin{assumption}\label{ass_mu_conv}
	Let $f$ be $\mu$-strongly-convex and continuously differentiable.
\end{assumption}
Since Assumption \ref{ass_pl} is more general than Assumption \ref{ass_mu_conv}, Theorem \ref{thm_noncvx_pl} can be applied to functions that satisfy Assumption \ref{ass_mu_conv}. If we set $A =   c = 0,$ we recover \citep[Theorem~4.6]{BotCurNoce} (see Corollary~\ref{cor_strongly_cvx_recover_bottou} in the appendix). If $A = C =   c = 0,$ we retrieve results comparable to those in \citep[Theorems~12--14]{BezHorRichSaf}, up to a multiplicative factor (see Corollary~\ref{cor_strongly_cvx_recover_beznosikov} in the appendix). Due to $\mu$-strong convexity, our result \eqref{eq_convergence_pl} also implies an iterate convergence, since we have $\left\|x^T-x^{*}\right\|^2\leq\frac{2}{\mu}\mathbb{E}\left[f(x^T) - f(x^{*})\right].$ However, in this case an additional factor of $\frac{2}{\mu}$ arises. Below we present a stronger result, yet, at a cost of imposing a stricter condition on the control variables from Assumption~\ref{ass_scalar_ABC}.
\begin{assumption}\label{ass_ABC_consts}
	Let $A,B,C$ and $ b$ be parameters from Assumption~\ref{ass_scalar_ABC}. Let $\mu$ be a strong convexity constant. Let $L$ be a smoothness constant. Suppose $A+ L(B+1-2 b) < \mu$ holds.
\end{assumption}
Under Assumptions \ref{ass_scalar_ABC} and \ref{ass_ABC_consts} we establish a similar result as the one obtained by \citet[Theorem~1]{HuSeiLes}. The authors impose a restriction of $\frac{1}{\kappa}$ from above on a constant with an analogous role as $B+1-2 b$ in Assumptions \ref{ass_scalar_ABC} and \ref{ass_ABC_consts} with $A=0.$ However, unlike us, the authors consider only a finite sum case which makes our result more general. Moreover, only a biased version of \algname{SGD} with a simple sampling strategy is analyzed by \citet{HuSeiLes}. Our results are applicable to a larger variety of gradient estimators and obtained under milder assumptions. Also, our proof strategy is different, and much simpler.
\begin{theorem}\label{thm_weak_bias} Let Assumptions \ref{ass_smooth}, \ref{ass_scalar_ABC}, \ref{ass_mu_conv} and \ref{ass_ABC_consts} hold. For every positive $s,$ satisfying
$
		A+ L(B+1-2 b) < s < \mu,
$
	choose a stepsize $\gamma$ such that
	\begin{equation}\label{eq_gamma_weak_bias}
		\squeeze		0<\gamma \leq \min\left\lbrace\frac{1-\frac{1}{s}\left(A+L\left(B+1-2 b\right)\right)}{A+LB}, \frac{1}{\mu-s}\right\rbrace.
	\end{equation}
	Then the iterates  of \algname{BiasedSGD} (Algorithm \ref{alg:SGD})  for every $T\geq 1$ satisfy
	\begin{equation}\label{eq_convergence_bias}
		\begin{split}
			\squeeze			\mathbb{E}\left[\left\|x^{T}-x^{*}\right\|^2\right]
			&\squeeze \leq\left(1-\gamma\left(\mu - s\right)\right)^T \left\|x^{0} - x^{*}\right\|^2 + \frac{\gamma C +\frac{C+2  c}{s}}{\mu-s}.
		\end{split}
	\end{equation}
\end{theorem}
In the standard result for (unbiased) \algname{SGD}, the convergence neighborhood term has the form of $\frac{\gamma C}{\mu},$ and it can be controlled by adjusting the stepsize. However, due to the generality of our analysis in the biased case, in \eqref{eq_convergence_bias}  we obtain an extra uncontrollable neighborhood term of the form $\frac{C+2  c}{s\left(\mu - s\right)}.$ 



When $A=C=  c=0,$ $B=1,$ $ b = 1,$ $s\to0,$ we recover exactly the classical result for \algname{GD}.
\begin{table*}[h]
	\centering
	\scriptsize
	\begin{threeparttable}
		\begin{tabular}{|c|c c c c|}
			\hline
			\bf Theorem & \bf Convergence rate &  \bf Compared to & \bf Rate we compare to & \bf Match? \\ 
			\hline
			\begin{tabular}{c}
				{{\tiny Thm \ref{thm_nonconvex}}}
			\end{tabular}				 & $\mathcal{O}\left(\frac{\delta^0 L}{\varepsilon^2}\max\left\lbrace B, \frac{12\delta^0 A}{\varepsilon^2},\frac{2C}{\varepsilon^2} \right\rbrace\right)$ & \citenum{khaled2022better}-Thm 2  & $\mathcal{O}\left(\frac{\delta^0 L}{\varepsilon^2}\max\left\lbrace B, \frac{12\delta^0 A}{\varepsilon^2},\frac{2C}{\varepsilon^2} \right\rbrace\right)$ & {\color{green}\checkmark} \\ 					
			\hline
			\begin{tabular}{c}
				{{\tiny Thm \ref{thm_nonconvex}}}
			\end{tabular}				 & \scalebox{.78}{$\mathcal{O}\left(\max\left\lbrace \frac{8(M+1)(m+1)}{(1-m)^2\varepsilon},\frac{16(M+1)\varphi^2+2\sigma^2}{(1-m)^2\varepsilon^2} \right\rbrace L\delta^0\right)$} &  \citenum{AjallStich}-Thm 4 & \scalebox{.78}{$\mathcal{O}\left(\max\left\lbrace \frac{M+1}{(1-m)\varepsilon}, \frac{2\sigma^2}{(1-m)^2\varepsilon^2}  \right\rbrace L\delta^0\right)$} & {\color{red}\ding{55}} \\ 					
			\hline
			\begin{tabular}{c}
				{{\tiny Thm \ref{thm_nonconvex}}}
			\end{tabular}				 & \scalebox{.78}{$\mathcal{O}\left(\max\left\lbrace \frac{8Q}{\varepsilon^2q^2}, \frac{4(U+u^2)}{\varepsilon q^2} \right\rbrace L\delta^0\right)$} &  \citenum{BotCurNoce}-Thm 4.8 & \scalebox{.78}{$\mathcal{O}\left(\max\left\lbrace \frac{8Q}{\varepsilon^2q^2}, \frac{4(U+u^2)}{\varepsilon q^2} \right\rbrace L\delta^0\right)$} & {\color{green}\checkmark} \\ 					
			\hline
			\begin{tabular}{c}
				{{\tiny Thm \ref{thm_noncvx_pl}}}\\
			\end{tabular}				 & \scalebox{.78}{$\widetilde{\mathcal{O}}\left(\max\left\lbrace \frac{2(M+1)(m+1)}{1-m},\frac{2(M+1)\varphi^2+\sigma^2}{\epsilon\mu(1-m) + 2\varphi^2} \right\rbrace \frac{\kappa}{1-m}\right)$} & \citenum{AjallStich}-Thm 6   & \scalebox{.78}{$\tilde{\mathcal{O}}\left(\max\left\lbrace (M+1), \frac{\sigma^2}{\varepsilon\mu(1-m)+\varphi^2}\right\rbrace\frac{\kappa}{1-m}\right)$} &  {\color{red}\ding{55}}\\ 					
			\hline
			\begin{tabular}{c}
				{\tiny Thm \ref{thm_strong_cvx}}
			\end{tabular}				 & {\tiny $\tilde{\mathcal{O}}\left(\max\left\lbrace 2, \frac{L(U+u^2)}{q^2\mu}, \frac{LQ}{\varepsilon\mu^2 q^2}\right\rbrace\right)$}  &  \citenum{BotCurNoce}-Thm 4.6 & {\tiny $\tilde{\mathcal{O}}\left(\max\left\lbrace 2, \frac{L\left(U + u^2\right)}{q^2\mu}, \frac{LQ}{\varepsilon\mu^2q^2} \right\rbrace\right)$}  & {\color{green}\checkmark}\\ 					
			\hline
			\begin{tabular}{c}
				{\tiny Thm \ref{thm_strong_cvx}}\\
			\end{tabular}				 & $\tilde{\mathcal{O}}\left(\left(\frac{\beta^2}{\alpha}\right)^2\frac{L}{\mu}\right)$  &  \citenum{BezHorRichSaf}-Thm 12 & $\tilde{\mathcal{O}}\left(\frac{\beta^2}{\alpha}\frac{L}{\mu}\right)$& {\color{red}\ding{55}}\\ 					
			\hline
			\begin{tabular}{c}
				{\tiny Thm \ref{thm_strong_cvx}}\\
			\end{tabular}				 & $\tilde{\mathcal{O}}\left(\left( \frac{\beta}{\tau}\right)^2\frac{L}{\mu}\right)$  &  \citenum{BezHorRichSaf}-Thm 13 & $\tilde{\mathcal{O}}\left(\frac{\beta}{\tau}\frac{L}{\mu}\right)$& {\color{red}\ding{55}}\\ 					
			\hline
			\begin{tabular}{c}
				{\tiny Thm \ref{thm_strong_cvx}}\\
			\end{tabular}				 & $\tilde{\mathcal{O}}\left(\delta^2\frac{L}{\mu}\right)$  &  \citenum{BezHorRichSaf}-Thm 14 & $\tilde{\mathcal{O}}\left(\delta\frac{L}{\mu}\right)$& {\color{red}\ding{55}}\\ 					
			\hline			
		\end{tabular}
	\end{threeparttable}
	\caption{Complexity comparison. We examine whether we can achieve the same convergence rate as obtained under stronger assumptions. In most cases, we ensure the same rate, albeit with inferior multiplicative factors due to the broader scope of the analysis. The notation $\tilde{\mathcal{O}}\left(\cdot\right)$ hides a logarithmic factor of $\log\frac{2\delta^0}{\varepsilon}$.}
	\label{tab_biased_vs_unbiased}    
\end{table*}
\section{Experiments}\label{sec:experiments}
To validate our theoretical findings, we conducted a series of numerical experiments on a binary classification problem. Specifically, we employed logistic regression with a non-convex regularizer: $$\squeeze \min \limits_{x \in \mathbb{R}^d}\left[f(x) \eqdef \frac{1}{n} \sum \limits \limits_{i=1}^n f_{i}(x)\right], \; \text{where} \; f_{i}(x) \eqdef \log \left(1+\exp \left(-y_{i} a_{i}^{\top} x\right)\right)+\lambda \sum \limits_{j=1}^d \frac{x_j^2}{1+x_j^2},$$
and $\left(a_{i}, y_{i}\right) \in \mathbb{R}^d \times \{-1,1\}, i=1, \ldots, n$ represent the training data samples. In all experiments, we set the regularization parameter $\lambda$ to a fixed value of $\lambda=1.$ We use datasets from the open LibSVM library \citep{chang2011libsvm}. 
We examine the performance of the proposed \algname{BiasedSGD} method with biased independent sampling without replacement (we call it \algname{BiasedSGD-ind}) in various settings (see Definition \ref{def_biased_sampling_no_replacement}). The primary goal of these numerical experiments is to demonstrate the alignment of our theoretical findings with the observed experimental results. To assess the performance of the methods throughout the optimization process, we monitor the metric $\sqnorm{\nabla f(x^t)}$, recomputed after every $10$ iterations. The algorithms are terminated after completing $5000$ iterations.
For each method, we use the largest theoretical stepsize. Specifically, for \algname{BiasedSGD-ind}, the stepsize is determined according to Corollary \ref{cor_noncvx_1} and Claim \ref{claim_sampling_no_division} with $\gamma=\min \left\{\frac{1}{\sqrt{L A K}}, \frac{ b}{L B}, \frac{  c}{L C}\right\}$, where $  c=0$,  $A = \frac{\max_i{L_i}}{\min_{i}{p_i}}$, $B = 0$, $C = 2A\Delta^{*} + s^2$, $ b = \min_{i} p_i$ and $s=0$.

More experimental details are provided in Appendix \ref{sec:exps_extra}.

\begin{figure}[h!]
	\centering
	\includegraphics[width=1.0\textwidth]{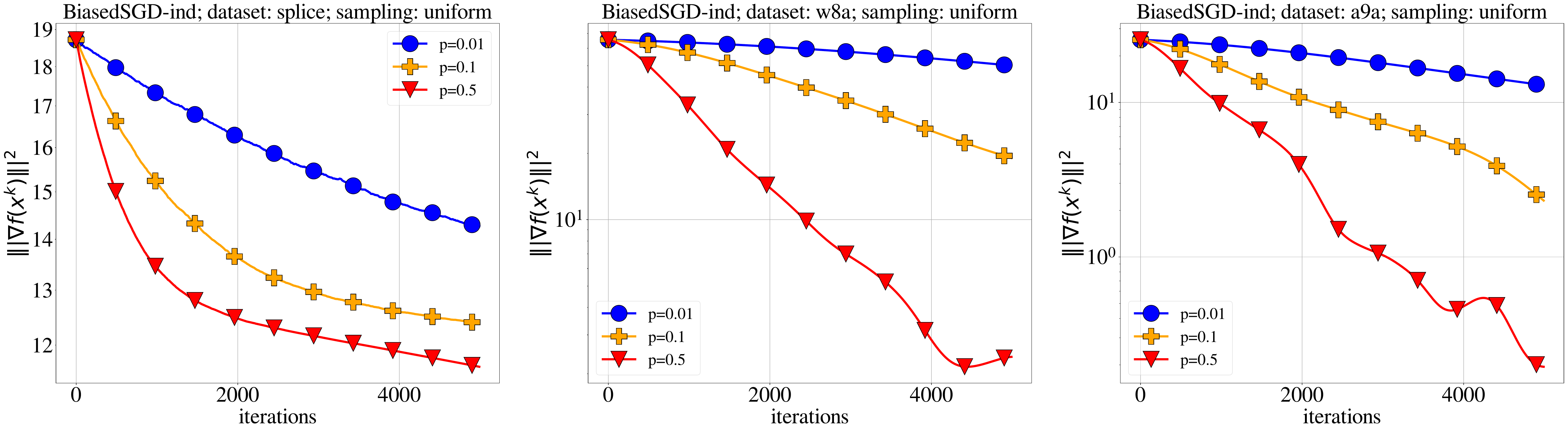}
	\caption{The performance of \algname{BiasedSGD-ind} with different choices of probabilities.}
	\label{fig:fig1}
\end{figure}

\paragraph{Experiment: The impact of the parameter $p$ on the convergence behavior.}
In the first experiment, we investigate how the convergence of \algname{BiasedSGD-ind} is affected as we increase the probabilities $p_i$, while keeping them equal for all data samples. According to the Corollary \ref{cor_noncvx_1}, larger $p_i$ values (resulting in an increase of the expected batch size) allow for a larger stepsize, which, in turn, improves the overall convergence. This behavior is evident in Figure \ref{fig:fig1}.

\clearpage

\bibliographystyle{plainnat}	
\bibliography{bibliography}

\begin{thebibliography}{53}
\providecommand{\natexlab}[1]{#1}
\providecommand{\url}[1]{\texttt{#1}}
\expandafter\ifx\csname urlstyle\endcsname\relax
  \providecommand{\doi}[1]{doi: #1}\else
  \providecommand{\doi}{doi: \begingroup \urlstyle{rm}\Url}\fi

\bibitem[Ajalloeian and Stich(2020)]{AjallStich}
Ahmad Ajalloeian and Sebastian~U Stich.
\newblock Analysis of {SGD} with biased gradient estimators.
\newblock \emph{arXiv preprint arXiv:2008.00051}, 2020.

\bibitem[Aji and Heafield(2017)]{aji2017sparse}
Alham~Fikri Aji and Kenneth Heafield.
\newblock Sparse communication for distributed gradient descent.
\newblock \emph{arXiv preprint arXiv:1704.05021}, 2017.

\bibitem[Alistarh et~al.(2018)Alistarh, Hoefler, Johansson, Konstantinov,
  Khirirat, and Renggli]{alistarh2018sparse}
Dan Alistarh, Torsten Hoefler, Mikael Johansson, Nikola Konstantinov, Sarit
  Khirirat, and Cedric Renggli.
\newblock The convergence of sparsified gradient methods.
\newblock In \emph{Advances in Neural Information Processing Systems
  (NeurIPS)}, volume~31, pages 5973--5983, 2018.

\bibitem[Beznosikov et~al.(2020)Beznosikov, Horv{\'a}th, Richt{\'a}rik, and
  Safaryan]{BezHorRichSaf}
Aleksandr Beznosikov, Samuel Horv{\'a}th, Peter Richt{\'a}rik, and Mher
  Safaryan.
\newblock On biased compression for distributed learning.
\newblock \emph{arXiv preprint arXiv:2002.12410}, 2020.

\bibitem[Bottou et~al.(2018)Bottou, Curtis, and Nocedal]{BotCurNoce}
L\'{e}on Bottou, Frank Curtis, and Jorge Nocedal.
\newblock Optimization methods for large-scale machine learning.
\newblock \emph{SIAM Review}, 60\penalty0 (2):\penalty0 223--311, 2018.

\bibitem[Chang and Lin(2011)]{chang2011libsvm}
Chih-Chung Chang and Chih-Jen Lin.
\newblock {LIBSVM}: a library for support vector machines.
\newblock \emph{{ACM} {T}ransactions on {I}ntelligent {S}ystems and
  {T}echnology (TIST)}, 2\penalty0 (3):\penalty0 1--27, 2011.

\bibitem[Chen et~al.(2021)Chen, Shen, Huang, and Liu]{chen2021quantized}
Congliang Chen, Li~Shen, Haozhi Huang, and Wei Liu.
\newblock Quantized adam with error feedback.
\newblock \emph{ACM Transactions on Intelligent Systems and Technology (TIST)},
  12\penalty0 (5):\penalty0 1--26, 2021.

\bibitem[Chen et~al.(2017)Chen, Zhang, Sharma, Yi, and
  Hsieh]{ChenZhangSharmaYiHsieh}
Pin-Yu Chen, Huan Zhang, Yash Sharma, Jinfeng Yi, and Cho-Jui Hsieh.
\newblock Zoo: Zeroth order optimization based black-box attacks to deep neural
  networks without training substitute models.
\newblock \emph{Proceedings of the 10th ACM Workshop on Artificial Intelligence
  and Security}, pages 15--26, 2017.

\bibitem[Condat et~al.(2022)Condat, Yi, and Richt{\'a}rik]{CondYiRich}
Laurent Condat, Kai Yi, and Peter Richt{\'a}rik.
\newblock Ef-bv: A unified theory of error feedback and variance reduction
  mechanisms for biased and unbiased compression in distributed optimization.
\newblock \emph{arXiv preprint arXiv:2205.04180}, 2022.

\bibitem[Cordonnier(2018)]{cordonnier2018convex}
Jean-Baptiste Cordonnier.
\newblock Convex optimization using sparsified stochastic gradient descent with
  memory.
\newblock Technical report, 2018.

\bibitem[Danilova and Gorbunov(2022)]{danilova2022distributed}
Marina Danilova and Eduard Gorbunov.
\newblock Distributed methods with absolute compression and error compensation.
\newblock In \emph{Mathematical Optimization Theory and Operations Research:
  Recent Trends: 21st International Conference, MOTOR 2022, Petrozavodsk,
  Russia, July 2--6, 2022, Revised Selected Papers}, pages 163--177. Springer,
  2022.

\bibitem[d'Aspremont(2008)]{DAspremont}
Alexandre d'Aspremont.
\newblock Smooth optimization with approximate gradient.
\newblock \emph{SIAM Journal on Optimization}, 19\penalty0 (3):\penalty0
  1171--1183, 2008.

\bibitem[Devolder et~al.(2014)Devolder, Glineur, and
  Nesterov]{DevolderGlineurNesterov}
Olivier Devolder, Fran\c{c}ois Glineur, and Yurii Nesterov.
\newblock First-order methods of smooth convex optimization with inexact
  oracle.
\newblock \emph{Math. Program.}, 146\penalty0 (1-2):\penalty0 37--75, 2014.

\bibitem[Dutta et~al.(2020)Dutta, Bergou, Abdelmoniem, Ho, Sahu, Canini, and
  Kalnis]{dutta2020discrepancy}
Aritra Dutta, El~Houcine Bergou, Ahmed~M Abdelmoniem, Chen-Yu Ho, Atal~Narayan
  Sahu, Marco Canini, and Panos Kalnis.
\newblock On the discrepancy between the theoretical analysis and practical
  implementations of compressed communication for distributed deep learning.
\newblock In \emph{Proceedings of the AAAI Conference on Artificial
  Intelligence}, volume~34, pages 3817--3824, 2020.

\bibitem[Fatkhullin et~al.(2021)Fatkhullin, Sokolov, Gorbunov, Li, and
  Richt\'{a}rik]{EF21BW}
Ilyas Fatkhullin, Igor Sokolov, Eduard Gorbunov, Zhize Li, and Peter
  Richt\'{a}rik.
\newblock {EF21} with bells \& whistles: practical algorithmic extensions of
  modern error feedback.
\newblock \emph{arXiv preprint arXiv:2110.03294}, 2021.

\bibitem[Goodfellow et~al.(2016)Goodfellow, Bengio, and Courville]{GoodBenCour}
Ian Goodfellow, Yoshua Bengio, and Aaron Courville.
\newblock \emph{Deep Learning}.
\newblock The MIT Press, 2016.

\bibitem[Gorbunov et~al.(2020)Gorbunov, Kovalev, Makarenko, and
  Richt{\'a}rik]{gorbunov2020linearly}
Eduard Gorbunov, Dmitry Kovalev, Dmitry Makarenko, and Peter Richt{\'a}rik.
\newblock Linearly converging error compensated {SGD}.
\newblock \emph{Advances in Neural Information Processing Systems},
  33:\penalty0 20889--20900, 2020.

\bibitem[Gower et~al.(2019)Gower, Loizou, Qian, Sailanbayev, Shulgin, and
  Richt{\'a}rik]{gower2019sgd}
Robert~Mansel Gower, Nicolas Loizou, Xun Qian, Alibek Sailanbayev, Egor
  Shulgin, and Peter Richt{\'a}rik.
\newblock Sgd: General analysis and improved rates.
\newblock In \emph{International conference on machine learning}, pages
  5200--5209. PMLR, 2019.

\bibitem[Gupta et~al.(2015)Gupta, Agrawal, Gopalakrishnan, and
  Narayanan]{gupta2015deep}
Suyog Gupta, Ankur Agrawal, Kailash Gopalakrishnan, and Pritish Narayanan.
\newblock Deep learning with limited numerical precision.
\newblock In \emph{International conference on machine learning}, pages
  1737--1746. PMLR, 2015.

\bibitem[Horv{\'a}th et~al.(2022)Horv{\'a}th, Ho, Horvath, Sahu, Canini, and
  Richt{\'a}rik]{CNAT}
Samuel Horv{\'a}th, Chen-Yu Ho, Ludovit Horvath, Atal~Narayan Sahu, Marco
  Canini, and Peter Richt{\'a}rik.
\newblock Natural compression for distributed deep learning.
\newblock In \emph{Mathematical and Scientific Machine Learning}, pages
  129--141. PMLR, 2022.

\bibitem[Hu et~al.(2021)Hu, Seiler, and Lessard]{HuSeiLes}
Bin Hu, Peter Seiler, and Laurent Lessard.
\newblock Analysis of biased stochastic gradient descent using sequential
  semidefinite programs.
\newblock \emph{Mathematical Programming}, 187:\penalty0 383--408, 2021.

\bibitem[Karimi et~al.(2016)Karimi, Nutini, and Schmidt]{KarNutSch}
Hamed Karimi, Julie Nutini, and Mark Schmidt.
\newblock Linear convergence of gradient and proximal-gradient methods under
  the polyak-\l ojasiewicz condition.
\newblock \emph{Machine Learning and Knowledge Discovery in Databases}, pages
  795–--811, 2016.

\bibitem[Karimireddy et~al.(2018)Karimireddy, Stich, and Jaggi]{KariStichJaggi}
Sai~Praneeth Karimireddy, Sebastian Stich, and Martin Jaggi.
\newblock Adaptive balancing of gradient and update computation times using
  global geometry and approximate subproblems.
\newblock 2018.

\bibitem[Karimireddy et~al.(2019)Karimireddy, Rebjock, Stich, and
  Jaggi]{karimireddy2019ef}
Sai~Praneeth Karimireddy, Quentin Rebjock, Sebastian~U. Stich, and Martin
  Jaggi.
\newblock Error feedback fixes {S}ign{SGD} and other gradient compression
  schemes.
\newblock In \emph{International Conference on Machine Learning (ICML)},
  volume~97, pages 3252--3261, 2019.

\bibitem[Khaled and Richt{\'a}rik(2023)]{khaled2022better}
Ahmed Khaled and Peter Richt{\'a}rik.
\newblock Better theory for {SGD} in the nonconvex world.
\newblock \emph{Transactions on Machine Learning Research}, 2023.
\newblock ISSN 2835-8856.
\newblock URL \url{https://openreview.net/forum?id=AU4qHN2VkS}.
\newblock Survey Certification.

\bibitem[Khirirat et~al.(2018{\natexlab{a}})Khirirat, Feyzmahdavian, and
  Johansson]{khirirat2018distributed}
Sarit Khirirat, Hamid~Reza Feyzmahdavian, and Mikael Johansson.
\newblock Distributed learning with compressed gradients.
\newblock \emph{arXiv preprint arXiv:1806.06573}, 2018{\natexlab{a}}.

\bibitem[Khirirat et~al.(2018{\natexlab{b}})Khirirat, Johansson, and
  Alistarh]{khirirat2018gradient}
Sarit Khirirat, Mikael Johansson, and Dan Alistarh.
\newblock Gradient compression for communication-limited convex optimization.
\newblock In \emph{2018 IEEE Conference on Decision and Control (CDC)}, pages
  166--171. IEEE, 2018{\natexlab{b}}.

\bibitem[Khirirat et~al.(2020)Khirirat, Magn{\'u}sson, and
  Johansson]{khirirat2020compressed}
Sarit Khirirat, Sindri Magn{\'u}sson, and Mikael Johansson.
\newblock Compressed gradient methods with hessian-aided error compensation.
\newblock \emph{IEEE Transactions on Signal Processing}, 69:\penalty0
  998--1011, 2020.

\bibitem[Khirirat et~al.(2022)Khirirat, Magn{\'u}sson, and
  Johansson]{khirirat2022eco}
Sarit Khirirat, Sindri Magn{\'u}sson, and Mikael Johansson.
\newblock Eco-fedsplit: Federated learning with error-compensated compression.
\newblock In \emph{ICASSP 2022-2022 IEEE International Conference on Acoustics,
  Speech and Signal Processing (ICASSP)}, pages 5952--5956. IEEE, 2022.

\bibitem[Lei et~al.(2019)Lei, Hu, Li, and Tang]{LeiHuLiTang}
Yunwei Lei, Ting Hu, Guiying Li, and Ke~Tang.
\newblock Stochastic gradient descent for nonconvex learning without bounded
  gradient assumptions.
\newblock pages 1--7, 2019.

\bibitem[Liu et~al.(2018)Liu, Kailkhura, Chen, Ting, Chang, and
  Amini]{LiuKailChenTingChangAmini}
Sijia Liu, Bhavya Kailkhura, Pin-Yu Chen, Paishun Ting, Shiyu Chang, and Lisa
  Amini.
\newblock Zeroth-order stochastic variance re- duction for nonconvex
  optimization.
\newblock \emph{Advances in neural information processing systems (NeurIPS)},
  31:\penalty0 3727--3737, 2018.

\bibitem[Mishchenko et~al.(2021)Mishchenko, Wang, Kovalev, and
  Richt{\'a}rik]{mishchenko2021intsgd}
Konstantin Mishchenko, Bokun Wang, Dmitry Kovalev, and Peter Richt{\'a}rik.
\newblock Intsgd: Adaptive floatless compression of stochastic gradients.
\newblock \emph{arXiv preprint arXiv:2102.08374}, 2021.

\bibitem[Moosavi-Dezfooli et~al.(2016)Moosavi-Dezfooli, Fawzi, Fawzi, and
  Frossard]{MoosFawziFrossard}
Seyed-Mohsen Moosavi-Dezfooli, Alhussein Fawzi, Omar Fawzi, and Pascal
  Frossard.
\newblock Universal adversarial perturbations.
\newblock \emph{arXiv preprints arXiv:1610.08401}, 2016.

\bibitem[Nemirovsky and Yudin(1983)]{NemirovskyYudin}
Arkadi Nemirovsky and David Yudin.
\newblock \emph{Problwm Complexity and Method Efficiency in Optimization}.
\newblock Wiley, New York, 1983.

\bibitem[Nesterov and Spokoiny(2017)]{NestSpok}
Yurii Nesterov and Vladimir Spokoiny.
\newblock Random gradient-free minimization of convex functions.
\newblock \emph{Found. Comput. Math.}, 17\penalty0 (2):\penalty0 527--566,
  2017.

\bibitem[Niu et~al.(2011)Niu, Recht, Re, and Wright]{NiuRechtReWright}
Feng Niu, Benjamin Recht, Christopher Re, and Stephen Wright.
\newblock Hogwild: A lock-free approach to parallelizing stochastic gradient
  descent.
\newblock In \emph{Advances in Neural Information Processing Systems
  (NeurIPS)}, volume~24, pages 693--701, 2011.

\bibitem[Polyak(1963)]{PolyakAsn}
Boris Polyak.
\newblock Gradient methods for minimizing functionals.
\newblock \emph{U.S.S.R. Comput. Math. Math. Phys.}, 3\penalty0 (4):\penalty0
  864--878, 1963.

\bibitem[Polyak(1987)]{Polyak}
Boris Polyak.
\newblock \emph{Introduction to Optimization}.
\newblock OptimizationSoftware, Inc., 1987.

\bibitem[Richt\'{a}rik et~al.(2021)Richt\'{a}rik, Sokolov, and
  Fatkhullin]{EF21}
Peter Richt\'{a}rik, Igor Sokolov, and Ilyas Fatkhullin.
\newblock {EF21}: A new, simpler, theoretically better, and practically faster
  error feedback.
\newblock In \emph{Advances in Neural Information Processing Systems}, 2021.

\bibitem[Richt{\'a}rik et~al.(2022)Richt{\'a}rik, Sokolov, Gasanov, Fatkhullin,
  Li, and Gorbunov]{richtarik20223pc}
Peter Richt{\'a}rik, Igor Sokolov, Elnur Gasanov, Ilyas Fatkhullin, Zhize Li,
  and Eduard Gorbunov.
\newblock 3pc: Three point compressors for communication-efficient distributed
  training and a better theory for lazy aggregation.
\newblock In \emph{International Conference on Machine Learning}, pages
  18596--18648. PMLR, 2022.

\bibitem[Robbins and Monro(1951)]{RobbinsMonro:1951}
Herbert Robbins and Sutton Monro.
\newblock A stochastic approximation method.
\newblock \emph{Annals of Mathematical Statistics}, 22:\penalty0 400--407,
  1951.

\bibitem[Sahu et~al.(2021)Sahu, Dutta, M~Abdelmoniem, Banerjee, Canini, and
  Kalnis]{sahu2021rethinking}
Atal Sahu, Aritra Dutta, Ahmed M~Abdelmoniem, Trambak Banerjee, Marco Canini,
  and Panos Kalnis.
\newblock Rethinking gradient sparsification as total error minimization.
\newblock \emph{Advances in Neural Information Processing Systems},
  34:\penalty0 8133--8146, 2021.

\bibitem[Sapio et~al.(2019)Sapio, Canini, Ho, Nelson, Kalnis, Kim,
  Krishnamurthy, Moshref, Ports, and Richt{\'a}rik]{sapio2019scaling}
Amedeo Sapio, Marco Canini, Chen-Yu Ho, Jacob Nelson, Panos Kalnis, Changhoon
  Kim, Arvind Krishnamurthy, Masoud Moshref, Dan~RK Ports, and Peter
  Richt{\'a}rik.
\newblock Scaling distributed machine learning with in-network aggregation.
\newblock \emph{arXiv preprint arXiv:1903.06701}, 2019.

\bibitem[Sapio et~al.(2021)Sapio, Canini, Ho, Nelson, Kalnis, Kim,
  Krishnamurthy, Moshref, Ports, and
  Richt\'{a}rik]{SapCanHoNelKalKinKrisMoshPorRich}
Amedeo Sapio, Marco Canini, Chen-Yu Ho, Jacob Nelson, Panos Kalnis, Changhoon
  Kim, Arvind Krishnamurthy, Masoud Moshref, Dan Ports, and Peter
  Richt\'{a}rik.
\newblock Scaling distributed machine learning with in-network aggregation.
\newblock In \emph{In 18th USENIX Symposium on Networked Systems Design and
  Implementation (NSDI 21)}, pages 785--808, 2021.

\bibitem[Schmidt et~al.(2011)Schmidt, Roux, and Bach]{SchmidtRouxBach}
Mark Schmidt, Nicolas Roux, and Francis Bach.
\newblock Convergence rates of inexact proximal-gradient methods for convex
  optimization.
\newblock \emph{Advances in neural information processing systems (NeurIPS)},
  24:\penalty0 1458--1466, 2011.

\bibitem[Shalev-Shwartz and Ben-David(2014)]{shai_book}
Shai Shalev-Shwartz and Shai Ben-David.
\newblock \emph{Understanding machine learning: from theory to algorithms}.
\newblock Cambridge University Press, 2014.

\bibitem[Stich and Karimireddy(2020)]{stich2020error}
Sebastian~U Stich and Sai~Praneeth Karimireddy.
\newblock The error-feedback framework: Better rates for sgd with delayed
  gradients and compressed updates.
\newblock \emph{The Journal of Machine Learning Research}, 21\penalty0
  (1):\penalty0 9613--9648, 2020.

\bibitem[Stich et~al.(2018)Stich, Cordonnier, and Jaggi]{Stich-EF-NIPS2018}
Sebastian~U. Stich, J.-B. Cordonnier, and Martin Jaggi.
\newblock Sparsified {SGD} with memory.
\newblock In \emph{Advances in Neural Information Processing Systems
  (NeurIPS)}, 2018.

\bibitem[Str{\"o}m(2015)]{strom2015scalable}
Nikko Str{\"o}m.
\newblock Scalable distributed dnn training using commodity gpu cloud
  computing.
\newblock 2015.

\bibitem[Sun(2020)]{Sun}
Ruo-Yu Sun.
\newblock Optimization for deep learning: An overview.
\newblock \emph{Journal of the Operations Research Society of China},
  8\penalty0 (2):\penalty0 249--294, 2020.

\bibitem[Tang et~al.(2020)Tang, Lian, Yu, Zhang, and Liu]{DoubleSqueeze}
Hanlin Tang, Xiangru Lian, Chen Yu, Tong Zhang, and Ji~Liu.
\newblock {D}ouble{S}queeze: {P}arallel stochastic gradient descent with
  double-pass error-compensated compression.
\newblock In \emph{Proceedings of the 36th International Conference on Machine
  Learning (ICML)}, 2020.

\bibitem[Tappenden et~al.(2016)Tappenden, Richt{\'a}rik, and
  Gondzio]{TappendenRichtarikGondzio}
Rachael Tappenden, Peter Richt{\'a}rik, and Jacek Gondzio.
\newblock Inexact coordinate descent: Complexity and preconditioning.
\newblock \emph{Journal of Optimization Theory and Applications}, 170:\penalty0
  144--176, 2016.

\bibitem[Wangni et~al.(2018)Wangni, Wang, Liu, and Zhang]{WangniWangLiuZhang}
Jianqiao Wangni, Jialei Wang, Ji~Liu, and Tong Zhang.
\newblock Gradient sparsification for communication-efficient distributed
  optimization.
\newblock In \emph{Advances in Neural Information Processing Systems
  (NeurIPS)}, volume~31, pages 1306--1316, 2018.

\end{thebibliography}

\newpage
\tableofcontents
\newpage
\appendix
	\section{Experiments: missing details }\label{sec:exps_extra}
	This section completes the experimental details mentioned in Section~\ref{sec:experiments}. The corresponding code can be found in the provided repository: \url{https://github.com/IgorSokoloff/guide-biased-sgd-experiments}.
	
	\paragraph{Datasets, Hardware, and Code Implementation.}
	The experiments utilized publicly available LibSVM datasets \citet{chang2011libsvm}, specifically the \texttt{splice}, \texttt{a9a}, and \texttt{w8a}. These algorithms were developed using Python 3.8 and executed on a machine equipped with 48 cores of Intel(R) Xeon(R) Gold 6246 CPU @ 3.30GHz. A summarized description of the datasets is available in Table~\ref{tbl:datasets_summary1}.
	
\begin{table}[h!]
	\centering
	\caption{Summary of the datasets}
	\label{tbl:datasets_summary1}
	\begin{tabular}{l l l l}
		\toprule
		Dataset & $n$ (dataset size) & $d$ (\# of features) \\
		\midrule			
		\texttt{splice} & $1000$ & $60$\\ 
		\texttt{a9a} & $32560$ & $123$ \\ 
		\texttt{w8a} & $49749$ & $300$ \\ 
		\bottomrule
	\end{tabular}
\end{table}

\paragraph{Hyperparameters.}
For the selected logistic regression problem, the smoothness constants $L$ and $L_{i}$ of the functions $f$ and $f_i$ were explicitly calculated as shown below:

\begin{eqnarray}
	L &=& \lambda_{max}\rb{\frac{1}{4m}\bA^{\top}\bA + 2\lambda \bI}\notag \\
	L_{i} &=& \lambda_{\max}\rb{ \frac{1}{4} a_{i}a_{i}^{\top} + 2\lambda \mI}.\notag
\end{eqnarray}

In the above equations, $\bA$ represents the dataset (data matrix), and $a_{i}$ signifies its $i$-th row.
Smoothness constants for the logistic regression objective on the selected datasets are presented in Table \ref{tbl:hyperparameters_per_dataset_summary2}.
	\begin{table}[h!]
	\caption{Smoothness Constants for Logistic Regression with $\lambda = 1$}
	\label{tbl:hyperparameters_per_dataset_summary2}
	\centering
	\begin{tabular}{l l l}
		\toprule
		Dataset &  $L$ & $L_{\max}$ \\
		\midrule			
		\texttt{w8a} &  $1.66$ & $29.5$ \\
		\texttt{a9a} &  $2.57$ & $4.5$ \\
		\texttt{splice} &  $97.83$ & $163.25$ \\
		\bottomrule
	\end{tabular}
\end{table}

Each method utilized the largest possible theoretical stepsize. 
	For the \algname{BiasedSGD-ind} method, the stepsize is determined based on Corollary \ref{cor_noncvx_1} and Claim \ref{claim_sampling_no_division} with $\gamma=\min \left\{\frac{1}{\sqrt{L A K}}, \frac{ b}{L B}, \frac{ c}{L C}\right\}$, where $ c=0$, $A = \frac{\max_i{L_i}}{\min_{i}{p_i}}$, $B = 0$, $C = 2A\Delta^{*} + s^2$, $ b = \min_{i} p_i$ and $s=0$.
	
	\paragraph{Experiment: The impact of the parameter $p$ on the convergence behavior (extra details).}
	The experiment visualized in Figure \ref{fig:fig1} involves varying the probability parameter $p$ within the set $\left\{0.01, 0.1, 0.5\right\}$. This manipulation directly influences the value of $A$, consequently affecting the theoretical stepsize $\gamma$.  In the context of \algname{BiasedSGD-ind}, the stepsize $\gamma$ is defined as $\min \frac{1}{\sqrt{L A K}}$. A comprehensive compilation of these parameters is represented in Table \ref{tbl:hyperparameters_per_dataset_summary3}.

	\begin{table}[h!]
	\caption{Parameters $A$ and theoretical stepsizes, determined by the choice of parameter $p$ and dataset}
	\label{tbl:hyperparameters_per_dataset_summary3}
	\centering
	\begin{tabular}{l l l l}
		\toprule
		Dataset & $p$ & $A$ & \begin{tabular}{c}Theoretical stepsize for \algname{BiasedSGD-ind}\\
			$\gamma=\min \frac{1}{\sqrt{L A K}}$
		\end{tabular} \\
		\midrule			
		\texttt{splice} & $0.01$ & $16325.0$ & $3.54 \cdot 10^{-4}$ \\
		& $0.1$ & $1632.5$ & $1.12 \cdot 10^{-3}$ \\
		& $0.5$ & $326.5$ & $2.50 \cdot 10^{-3}$ \\
		\midrule
		\texttt{a9a} & $0.01$ & $550.0$ & $1.01 \cdot 10^{-2}$ \\
		& $0.1$ & $55.0$ & $3.19 \cdot 10^{-2}$ \\
		& $0.5$ & $11.0$ & $7.13 \cdot 10^{-2}$ \\
		\midrule
		\texttt{w8a} & $0.01$ & $3050.0$ & $4.96 \cdot 10^{-3}$ \\
		& $0.1$ & $305.0$ & $1.57 \cdot 10^{-2}$ \\
		& $0.5$ & $61.0$ & $3.51 \cdot 10^{-2}$ \\
		\bottomrule
	\end{tabular}
\end{table}
	
	\section{Sources of bias: further discussion and new estimators}\label{section_sources_of_bias_appendix}
	In Section~\ref{section_sources_of_bias_main} of the main part of the paper we describe different sources of bias and provide general forms of estimators that arise in each scenario. However, we do not present any concrete practical examples of stochastic gradients. In this section we define several important realistic estimators and characterize them in terms of \hyperlink{Biased ABC}{Biased ABC} framework. For proofs of results in this section, see Section~\ref{section_sources_proofs}.
	
	For a finite-sum problem \ref{eq_finite_sum}, consider a setting when the bias is induced by a subsampling strategy of which we lack the information. Let us introduce (without aiming to be exhaustive) a specific (and practical) sampling distribution and an estimator, which satisfies Assumption~\ref{ass_scalar_ABC}.
	
	\begin{definition}[Biased independent sampling without replacement]\label{def_biased_sampling_no_replacement}
		Let $p_1,p_2,\ldots,p_n$ be probabilities, $0<p_i\leq 1$ for all $i\in[n],$ $\sum_{i=1}^{n}p_i\in(0,n].$ For every $i\in[n],$ define a random set as follows:
		$$
		S_i = 
		\begin{cases}
			\{i\} & \text{with probability } p_i,\\
			\varnothing & \text{with probability } 1-p_i.
		\end{cases}
		$$
		Define a random subset $S\subseteq [n]$ by taking the union of these random sets: $S\eqdef\bigcup_{i=1}^nS_i.$
		Put 
		\begin{equation}\label{eq_indicators}
			\mathbb{I}_{i\in S} = 
			\begin{cases}
				1, & i\in S,\\
				0, & \text{otherwise}.
			\end{cases}
		\end{equation}
		For every $i\in[n],$ define $v_i=\frac{\mathbb{I}_{i\in S}}{|S|}.$ Let $g(x) = \tilde{g}(x)+\textbf{X},$ where 
		\begin{equation*}
			\tilde{g}(x) = \frac{1}{|S|}\sum_{i=1}^n\mathbb{I}_i\nabla f_i(x),
		\end{equation*}
		and $\textbf{X}$ is a random variable independent of $S,$ such that $\mathbb{E}[\textbf{X}]= 0,$ $\mathbb{V}[\textbf{X}] = s^2.$
	\end{definition}
	The practical setting where this stochastic gradient might be useful can have the following structure. There is an oracle that, for every $i\in[n],$ decides with an unknown probability $p_i$ whether to provide the information of $\nabla f_i$ at the iteration $k$ or not. Since the probabilities $p_i$ are unknown, they may be substituted for their estimators $\mathbb{I}_i.$ The stochastic gradient is then calculated as a simple average of all gradients with these estimators as weights. Note that a setting with $\sum_{i=1}^{n}p_i=1$ corresponds to the single-machine setup.
	
	The subsampling strategy from Definition~\ref{def_biased_sampling_no_replacement} can be used in another practical scenario. Consider a situation where access to the entire dataset is not available. In such cases, a \textit{fixed batch strategy} can be employed. This strategy involves sampling a single batch $S$ at step $0$ and subsequently using it throughout the entire optimization process.
	
	In the proof of Theorem~\ref{thm_informal_abc} (parts~\ref{item_abc_from_fsml}~and~\ref{item_abc_from_bnd}), we demonstrate that in a very simple setting the stochastic gradient from Definition~\ref{def_biased_sampling_no_replacement} does not satisfy Assumptions \ref{ass_stich_decomposition} and \ref{ass_first_and_second_mmt_limits} (and, therefore, to any other assumption from Section~\ref{sect_existing_models}). We want to show that under very mild restrictions on functions $f_i,$ $g(x)$ satisfies \hyperlink{Biased ABC}{Biased ABC} assumption.
	\begin{assumption}\label{ass_smooth_functionwise}
		Each $f_i$ is bounded from below by $f_i^{*}$ and $L_i$-smooth. That is, for all $x,y\in\mathbb{R}^d,$ we have
		\begin{equation*}
			f_i(y) \leq f_i(x) + \langle \nabla f_i(x), y - x\rangle +\frac{L_i}{2}\left\|y-x\right\|^2.
		\end{equation*}
	\end{assumption}
	Here and many times below in the paper we rely on the following important lemma.
	\begin{lemma}\label{lemma_smooth_bregman}
		Let $f$ be a function for which Assumption \ref{ass_smooth} is satisfied. Then, for all $x\in\mathbb{R}^d,$ we have
		\begin{equation*}
			\left\|\nabla f(x)\right\|^2\leq 2LD_f(x, x^{*}).
		\end{equation*}
	\end{lemma}
	In the nonconvex case the expression takes the following form:
	\begin{equation*}
		\left\| \nabla f(x) \right\|^2\leq 2L\left(f(x) - f^{*}\right),\quad \forall x\in\mathbb{R}^d.
	\end{equation*}
	This lemma appears in \citep{khaled2022better} and in several recent works on the convergence of \algname{SGD}. We give its proof in Sectioin~\ref{section_lemma_smooth_bregman_proof}. Equipped with Lemma \ref{lemma_smooth_bregman}, we can prove the following claim that motivates the inclusion of a Bregman Divergence term in \eqref{eq_ABC}. The reason why biased sampling gradient estimator does not satisfy Assumptions~\ref{ass_first_set},~\ref{ass_stich_decomposition}~and~\ref{ass_first_and_second_mmt_limits} is because its variance contains a sum of squared client gradient norms, which, in general, can not be bounded in terms of the squared norm of the full gradient. In fact, for a variety of biased sampling estimators this obstacle may occur, and this additionally motivates establishing new theory under the general assumption proposed in the present paper.
	\begin{claim}\label{claim_sampling_no_division}
		Suppose Assumptions \ref{ass_smooth} and \ref{ass_smooth_functionwise} hold.  Let 
		\begin{equation}\label{eq_def_delta_star}
			\Delta^{*} \eqdef \frac{1}{n}\sum_{i=1}^{n}\left(f^{*} - f_i^{*}\right).
		\end{equation}
		Then, gradient estimator from Definitioin~\ref{def_biased_sampling_no_replacement} satisfies Assumption \ref{ass_scalar_ABC} with $ b = \min_i\left\lbrace p_i\right\rbrace,$ $  c=0,$
		\begin{equation*}
			A = \frac{\max_i\{L_i\}}{\min_{i}{p_i}},\quad B = 0, \quad C = 2A\Delta^{*} + s^2.
		\end{equation*}
	\end{claim}
	
	In \citep{khaled2022better}, for a finite-sum problem \eqref{eq_finite_sum}, in the unbiased case the following general stochastic gradient is considered. Given a sampling vector $v\in\mathbb{R}^d$ drawn from some distribution $\mathcal{D}$ (where a sampling vector is one such that $\mathbb{E}_{\mathcal{D}}\left[v_i\right]=c_i,$ $c_i\geq 0,$ for all $i\in[n]$), for $x\in\mathbb{R}^d,$ define the stochastic gradient $g(x)\eqdef\frac{1}{n}\sum_{i=1}^{n}v_i\nabla f_i(x).$ We do not require $v_i$ to cause unbiasedness. Under mild assumptions on functions $f_i$ and the sampling vectors $v_i,$ we prove that $g(x)$ satisfies \hyperlink{Biased ABC}{Biased ABC} assumption, for all non-degenerate distributions $\mathcal{D}.$
	\begin{claim}\label{claim_distributions}
		Suppose Assumption~\ref{ass_smooth_functionwise} holds and, for all $i\in[n],$ we have $\mathbb{E}\left[v_i^2\right]<\infty.$ Then Assumption~\ref{ass_scalar_ABC} holds for $g(x)$ with $A=\max_i\left\lbrace L_i\mathbb{E}\left[v_i^2\right] \right\rbrace ,$ $B=0,$ $C=2A\Delta^{*},$ $ b=\min_i\left\lbrace c_i \right\rbrace,$ $  c=0.$
	\end{claim}
	Note, that in \citep[Proposition~2]{khaled2022better} it is proven that $\Delta^{*}\geq 0.$ The requirement of $\mathbb{E}\left[v_i^2\right]<\infty$ is very weak and satisfied for  almost all practical subsampling schemes in the literature. However, the generality of Claim~\ref{claim_distributions} comes at a cost since it leads to very pessimistic choices of constants in Assumption~\ref{ass_scalar_ABC}. 
	
	Our framework is general enough to establish the convergence of biased stochastic gradient quantization or compression schemes. Consider the finite-sum problem \eqref{eq_finite_sum} and let us propose the following new practical biased gradient estimator.
	\begin{definition}[Distributed general biased rounding]\label{def_distributed_biased_rounding}
		Let $\{a_k\}_{k\in\mathbb{Z}}$ be an arbitrary increasing sequence of positive numbers such that $\inf_k\{a_k\}=0,$ and $\sup_k\{a_k\} = \infty.$ Then, for all $j\in[n],$ $i\in[d],$ define
		\begin{equation*}
			\tilde{g}_j(x)_i \eqdef \mathrm{sign}\left(\nabla f(x)_i\right)\;\mathrm{arg}\min_{y\in\{a_k\}}|y-|\nabla f(x)_i||,\quad i\in[d].
		\end{equation*}
		For every $j\in[n],$ define mutually independent random variables
		\begin{equation*}
			\mathbb{I}_j = 
			\begin{cases}
				1, & \text{with probability }0 < p_j < 1,\\
				0, & \text{with probability }1 - p_j.
			\end{cases}
		\end{equation*}
		For every $x\in\mathbb{R}^d,$ define a gradient estimator
		\begin{equation*}
			g(x) = \frac{1}{n}\sum_{j=1}^{n}\left(\mathbb{I}_j\tilde{g}_j(x)+ \left(1 - \mathbb{I}_j\right)\nabla f_j(x)\right).
		\end{equation*}
	\end{definition}
	The practical setting where $g(x)$ might be used is a distributed problem where client node $j\in[n]$ decides with probability $p_j$ whether to send the compressed gradient or not. Master nodes which does not know $p_j$ simply averages the received stochastic gradients. In this case we preserve more information in comparison to the setting when we use compression at every step. On the other hand, gradients are compressed with positive probability, and we diminish the communication complexity versus the setting without any compression. That is, we have a flexible setting which is useful in practice.
	
	As before, we prove that $g(x)$ satisfies \hyperlink{Biased ABC}{Biased ABC} assumptioin under very mild conditions.
	\begin{claim}\label{claim_distributed_biased_rounding}
		Suppose Assumption \ref{ass_smooth_functionwise} holds and, for all $i\in[n],$ we have $\mathbb{E}\left[v_i^2\right]<\infty.$ Then the distributed general biased rounding estimator $g(x)$ satisfies Assumption \ref{ass_scalar_ABC} with
		\begin{equation}\label{eq_A_distributed_rounding}
			A = A_r \eqdef \frac{2}{n}\max_j\{L_j\}\max_j\{p_j(1-p_j)\}\left(\left(\sup_{k\in\mathbb{Z}}\frac{2a_{k+1}}{a_k+a_{k+1}}\right)^2 + 1\right),
		\end{equation}
		\begin{equation}\label{eq_B_distributed_rounding}
			B = B_r \eqdef 2\max_j\{p_j^2\}\left(\left(\sup_{k\in\mathbb{Z}}\frac{2a_{k+1}}{a_k+a_{k+1}}\right)^2 + 1\right),
		\end{equation}
		\begin{equation}\label{eq_C_distributed_rounding}
			C = C_r \eqdef \frac{4}{n}\max_j\{L_j\}\max_j\{p_j(1-p_j)\}\left(\left(\sup_{k\in\mathbb{Z}}\frac{2a_{k+1}}{a_k+a_{k+1}}\right)^2 + 1\right)\Delta^{*},
		\end{equation}
		\begin{equation}\label{eq_psi_distributed_rounding}
			b =  b_r \eqdef \max_j\{p_j\} \cdot \inf_{k\in\mathbb{Z}}\frac{2a_k}{a_k+a_{k+1}}  + \max_j\{1 - p_j\}
		\end{equation}
		\begin{equation}\label{eq_chi_distributed_rounding}
			c =   c_r = 0.
		\end{equation}
	\end{claim}
	From Claims~\ref{claim_sampling_no_division},~\ref{claim_distributions} and \ref{claim_distributed_biased_rounding} we see that, in fact, \hyperlink{Biased ABC}{Biased ABC} is not an additional assumption, but an inequality that is automatically satisfied under such settings.
	
	One of the simplest models of bias is the case of additive noise, that is
	\begin{equation*}
		g(x) = \nabla f(x) + \mathcal{Z},
	\end{equation*}
	where $\mathcal{Z}$ is a random variable satisfying $\mathbb{E}\left[\mathcal{Z}\right]= a,$ $a\in\mathbb{R}^d,$ $\mathbb{E}\left[\left\|\mathcal{Z}\right\|^2\right]=\sigma^2,$ $\sigma\in\mathbb{R}.$ It may happen in practise that, e.g., during the communication process in the distributed setting of the finite-sum problem \eqref{eq_finite_sum} transmitted gradients become noisy, and this simple model captures such a scenario. Models of this type were previously analyzed in \citep{AjallStich}. Clearly, \hyperlink{BND}{BND} assumption is satisfied. It means (see Figure~\ref{fig_diagram}), that they are covered by \hyperlink{Biased ABC}{Biased ABC} framework as well. However, models of this type impose rather strong restrictions on the stochastic gradient: they fail to capture a multiplicative biased noise that arises in the case of gradient compression operators and are not suitable for simulating subsampling schemes.
	
	\section{Known gradient estimators in biased ABC framework}\label{section_grad_estimators}
	In this section we define several known biased gradient estimators and for each of them, we present values of control variables $A,B,C,b,c$ within our \hyperlink{Biased ABC}{Biased ABC} framework. Also, these values are shown in Table~\ref{tab_estimators_full} for convenience of the reader. Formal proofs can be found in Section~\ref{section_grad_estimators_proofs}. In Table~\ref{tab_estimators_in_assumptions_full} we demonstrate a summary on inclusioin of each estimator from this section into every framework from Section~\ref{sect_existing_models}.
	\begin{definition}[Top-$k$ sparsifier -- \cite{aji2017sparse, alistarh2018sparse}]\label{def_top_ell}
		Let gradient estimator $g(x)$ be defined as
		\begin{equation*}
			g(x) \eqdef \sum_{i=d-k+1}^{d}\left(\nabla f(x)\right)_{(i)}e_{(i)}, \quad \forall x\in\mathbb{R}^d,
		\end{equation*}
		where coordinates are ordered with respect to their absolute values: 
		$$
		|\left(\nabla f(x)\right)_{(1)}|\leq|\left(\nabla f(x)\right)_{(2)}|\leq\ldots\leq|\left(\nabla f(x)\right)_{(d)}|.
		$$
	\end{definition}
	\begin{claim}\label{claim_top_l}
		Top-$k$ sparsifier $g(x)$ satisfies Assumption \ref{ass_scalar_ABC} with $ b = \frac{k}{d},$ $  c=0,$ $A=0,$ $B=1,$ $C=0.$
	\end{claim}
	\begin{definition}[Rand-$k$ -- \cite{Stich-EF-NIPS2018}]\label{def_random_l}
		For every $x\in\mathbb{R}^d,$ let
		\begin{equation*}
			g(x)\eqdef\frac{d}{k}\sum_{i\in S}\left(\nabla f(x)\right)_ie_i,
		\end{equation*}
		where $S$ is a random subset of $[d]$ chosen uniformly.
	\end{definition}
	\begin{claim}\label{claim_random_l}
		Rand-$k$ estimator $g(x)$ satisfies Assumption~\ref{ass_scalar_ABC} with $A=0,$ $B=\frac{d}{k},$ $C=0,$ $ b=1,$ $  c=0.$
	\end{claim}
	
	\begin{definition}[Biased Rand-$k$ sparsifier -- \cite{BezHorRichSaf}]\label{def_biased_random_l}
		For every $x\in\mathbb{R}^d,$ let
		\begin{equation*}
			g(x)\eqdef\sum_{i\in S}\left(\nabla f(x)\right)_ie_i,
		\end{equation*}
		where $S$ is a random subset of $[d]$ chosen uniformly.
	\end{definition}
	
	\begin{claim}\label{claim_biased_random_l}
		Biased Rand-$k$ sparsifier $g(x)$ satisfies Assumption \ref{ass_scalar_ABC} with $ b = \frac{k^2}{d^2},$ $  c=0,$ $A=C=0,$ $B=\frac{k}{d}.$
	\end{claim}
	
	\begin{definition}[Adaptive random sparsification -- \cite{BezHorRichSaf}]\label{def_adaptive_random_sparsification}
		Adaptive random sparsification estimator is defined via
		$$
		g(x)\eqdef\left(\nabla f(x)\right)_i e_i \quad \text { with probability } \quad \frac{\left|\left(\nabla f(x)\right)_i\right|}{\left\|\nabla f(x)\right\|_1}
		$$
	\end{definition}
	\begin{claim}\label{claim_adaptive_random_l_sparsifier}
		Adaptive random sparsifier $g(x)$ satisfies Assumption~\ref{ass_scalar_ABC} with $A=C=  c=0,$ $B=1,$ $ b=\frac{1}{d}.$
	\end{claim}
	\begin{definition}[General unbiased rounding estimator -- \cite{BezHorRichSaf}]\label{def_general_unbiased_rounding}
		Let $\{a_k\}_{k\in\mathbb{Z}}$ be an arbitrary increasing sequence of positive numbers such that $\inf_k a_k = 0,$ $\sup_k a_k = \infty.$ Define the rounding estimator $g(x)$ in the following way: if $a_k\leq \left| \nabla f(x)_i \right|\leq a_{k+1},$ for a coordinate $i\in[d],$ then
		\begin{equation*}
			g(x)_i = 
			\begin{cases}
				\mathrm{sign}(\nabla f(x)_i)a_k, & \text{with probability } \frac{a_{k+1} - |\nabla f(x)_i|}{a_{k+1} - a_k},\\
				\mathrm{sign}(\nabla f(x)_i)a_{k+1}, & \text{with probability } \frac{|\nabla f(x)_i| - a_k}{a_{k+1} - a_k}.\\
			\end{cases}
		\end{equation*}
	\end{definition}
	Put
	\begin{equation}\label{eq_rounding_constant}
		Z \eqdef \sup_{k\in\mathbb{Z}}\left(\frac{a_k}{a_{k+1}}+\frac{a_{k+1}}{a_k}+2\right).
	\end{equation}
	\begin{claim}\label{claim_general_adaptive_rounding}
		General unbiased rounding estimator $g(x)$ satisfies Assumption~\ref{ass_scalar_ABC} with $A=C=  c=0,$ $B=\frac{Z}{4},$ $ b=1.$
	\end{claim}
	
	\begin{definition}[General biased rounding -- \cite{BezHorRichSaf}]\label{def_general_biased_rounding}
		Let $\left(a_k\right)_{k \in \mathbb{Z}}$ be an arbitrary increasing sequence of positive numbers such that $\inf a_k=0$ and $\sup a_k=\infty$. Then general biased rounding is defined via
		$$
		g(x)_i\eqdef\operatorname{sign}\left(\left(\nabla f(x)\right)_i\right) \arg \min _{t \in\left(a_k\right)}|t-| \left(\nabla f(x)\right)_i||, \quad i \in[d] .
		$$
	\end{definition}
	Put
	\begin{equation}\label{eq_biased_rounding_consts}
		F = \sup _{k \in \mathbb{Z}} \frac{2 a_{k+1}}{a_k+a_{k+1}},\; G = \inf _{k \in \mathbb{Z}} \frac{2 a_{k}}{a_k+a_{k+1}}.
	\end{equation}
	\begin{claim}\label{claim_biased_rounding}
		Adaptive random sparsifier $g(x)$ satisfies Assumption~\ref{ass_scalar_ABC} with $A=C=  c=0,$ $B=F^2,$ $ b = \frac{G^2}{F}.$
	\end{claim}
	\begin{definition}[Natural compression -- \cite{CNAT}]\label{def_natural_compression}
		Natural compression estimator $g_{nat}(x)$ is the special case of general unbiased rounding operator (see Definition~\ref{def_general_unbiased_rounding}) when $a_k=2^k,$ $k\in\mathbb{N}.$
	\end{definition}
	\begin{claim}\label{claim_natural_compression}
		Natural compression estimator $g(x)$ satisfies Assumption~\ref{ass_scalar_ABC} with $A=C=  c=0,$ $B=\frac{9}{8},$ $ b=1.$
	\end{claim}
	\begin{definition}[General exponential dithering  -- \cite{BezHorRichSaf}]\label{def_general_exponential_dithering}
		For $a>1$, define general exponential dithering estimator with respect to $k_p$-norm and with $s$ exponential levels $0<a^{1-s}<a^{2-s}<\cdots<a^{-1}<1$ via
		$$
		\left(g(x)\right)_i\eqdef\left\|\nabla f(x)\right\|_p \times \operatorname{sign}\left(\left(\nabla f(x)\right)_i\right) \times \xi\left(\frac{\left|\left(\nabla f(x)\right)_i\right|}{\|\nabla f(x)\|_p}\right),
		$$
		where the random variable $\xi(t)$ for $t \in\left[a^{-u-1}, a^{-u}\right]$ is set to either $a^{-u-1}$ or $a^{-u}$ with probabilities proportional to $a^{-u}-t$ and $t-a^{-u-1}$, respectively.
	\end{definition}
	Put $r=\min (p, 2)$ and
	\begin{equation}\label{eq_dithering_constant}
		H_a=\frac{1}{4}\left(a+\frac{1}{a}+2\right)+d^{\frac{1}{r}} a^{1-s} \min \left(1, d^{\frac{1}{r}} a^{1-s}\right)
	\end{equation}
	\begin{claim}\label{claim_general_exponential_dithering}
		General exponential dithering estimator $g(x)$ satisfies Assumption~\ref{ass_scalar_ABC} with $A~=~C~=~  c=~0,$ $B=H_a,$ $ b=1,$ where $H_a$ is defined in \eqref{eq_dithering_constant}.
	\end{claim}
	\begin{definition}[Natural dithering -- \cite{CNAT}]\label{def_natural_dithering}
		Natural dithering without norm compression is the special case of general exponential dithering when $a=2$ (see Definition~\ref{def_general_exponential_dithering}).
	\end{definition}
	\begin{claim}\label{claim_natural_dithering}
		Natural dithering estimator satisfies Assumption~\ref{ass_scalar_ABC} with $A=C=  c=0,$ $B=H_2,$ $ b=1.$
	\end{claim}
	\begin{definition}[Composition of Top-$k$ with exponential dithering -- \cite{BezHorRichSaf}]\label{def_topk_exp_dithering}
		Let $g_{\text{top}}(x)$ be the Top-$k$ sparsification operator (see Definition~\ref{def_top_ell}) and $g_{dith}(x)$ be general exponential dithering operator with some base $a>1$ and parameter $H_a$ from \eqref{eq_dithering_constant}. Define a new compression operator as the composition of these two:
		$$
		g(x)\eqdef g_{\text{dith}}\left(g_{\text{top}}(x)\right).
		$$
	\end{definition}
	In this definition we imply that the dithering operator is applied to the vector yielded after Top-$k$ sparsification, not to the gradient as it was defined.
	\begin{claim}\label{claim_composition_top_l_exp_dithering}
		Composition of Top-$k$ with exponential dithering estimator $g(x)$ satisfies Assumption~\ref{ass_scalar_ABC} with $A=C=  c=0,$ $B=H_a^2,$ $ b=\frac{k}{dH_a}.$
	\end{claim}
	\begin{definition}[Gaussian smoothing -- \cite{Polyak}]\label{def_gaussian_smoothing}
		The following zero-order stochastic gradient, which we call Gaussian smoothing as in \citep{AjallStich}, is defined as
		\begin{equation*}
			g_{GS}(x) = \frac{f(x+\tau z) - f(x)}{\tau}\cdot z,
		\end{equation*}
		where $\tau>0$ is a smoothing parameter, and $z\sim \mathcal{N}\left(0, I\right)$ is a random Gaussian vector. 
	\end{definition}
	\begin{claim}\label{claim_gaussian_smoothing}
		Gaussian smoothing estimator $g(x)$ satisfies Assumption~\ref{ass_scalar_ABC} with
		\begin{equation*}
			A=A_{GS}\eqdef0,\;B=B_{GS}\eqdef2(d+4),\;C=C_{GS}\eqdef\frac{\tau^2}{2}L^2(d+6)^3,
		\end{equation*}
		\begin{equation}\label{eq_gaussian_smoothing_consts}
			b= b_{GS}=\frac{1}{2},\;   c=  c_{GS}\eqdef\frac{\tau^2}{8}L^2(d+3)^3.
		\end{equation}
	\end{claim}
	\begin{definition}[Hard-threshold sparsifier -- \cite{sahu2021rethinking} ]\label{def_hard_threshold_sparsifier}
		For some $w\geq 0,$ define the estimator $g^w_{HT}(x)$ as
		\begin{equation*}
			\left(g^{w}_{HT}(x)\right)_i =
			\begin{cases}
				\left(\nabla f(x)\right)_i, & \left|\left(\nabla f(x)\right)_i\right|\geq w,\\
				0, & \text{otherwise},
			\end{cases}
		\end{equation*}
		for every $i\in[d].$
	\end{definition}
	\begin{claim}\label{claim_hard_threshold_sparsifier}
		Hard-threshold estimator sastisfies Assumption~\ref{ass_scalar_ABC} with $A=C=0,$  $B=1,$ $ b=1,$ $  c = w^2d.$
	\end{claim}
	\begin{definition}[Scaled integer rounding -- \cite{SapCanHoNelKalKinKrisMoshPorRich}]\label{def_scaled_rounding}
		In a distributed setting \eqref{eq_distibuted_compression}, for every $i\in[n],$ let $\mathcal{C}_i:\nabla f_i(x)\to\frac{1}{\chi}R\left(\chi\nabla f_i(x) \right),$ where $\chi>0$ is a scaling factor, $R$ is a rounding to the nearest integer operator. That is, a scaling integer rounding estimator is defined as
		$$
		g(x) = \frac{1}{n}\sum_{i=1}^{n}\frac{1}{\chi}R\left( \chi\nabla f_i(x) \right).
		$$
	\end{definition}
	\begin{claim}\label{claim_scaled_rounding}
		Scaling integer estimator satisfies Assumption~\ref{ass_scalar_ABC} with $A=0,$ $B=2,$ $C=\frac{2d}{\chi^2},$ $b=\frac{1}{2},$ $c=\frac{d}{2\chi^2}.$
	\end{claim}

	\begin{definition}[Biased dithering -- \cite{khirirat2018gradient}]\label{def_biased_dithering}
		Biased dithering estimator $g(x)$ is defined as
		$$
		\rb{g(x)}_i= \norm{\nabla f(x)} \operatorname{sign}\rb{\left(\nabla f(x)\right)_i}, \quad i\in[d], \quad \forall x\in\mathbb{R}^d.
		$$
	\end{definition}
	\begin{claim}\label{claim_biased_dithering}
		Biased dithering operator satisfies Assumption~\ref{ass_scalar_ABC} with $A=0,$ $B= d,$ $C=0,$ $b=1,$ $c=0.$
	\end{claim}
	\begin{definition}[Sign compression -- \citep{karimireddy2019ef}]\label{def_sign_compression}
	Sign compression operator is defined as
		\begin{equation*}
				g(x)\eqdef\frac{\|\nabla f(x)\|_1}{d} \operatorname{sign}\left(\nabla f(x)\right), \quad \forall x\in\mathbb{R}^d.
		\end{equation*}
	\end{definition}
	\begin{claim}\label{claim_sign_compression}
		Sign compression operator satisfies Assumption~\ref{ass_scalar_ABC} with $A=C=c=0,$ $B=2\left(2-\frac{1}{d}\right),$ $b=\frac{1}{2d}.$
	\end{claim}
	
	In Table~\ref{tab_estimators_full} we gather the results from the current section. In Table~\ref{tab_estimators_in_assumptions_full} we show whether the estimators in this section fit or not to mentioned in the present work frameworks.
	\begin{table*}[h]
		\centering
		\scriptsize
		
		\begin{threeparttable}
			\begin{tabular}{|c|c|c c c c c|}
				\hline
				Name of an estimator  & Definition & $A$ & $B$ & $C$ & $ b $&$  c$ \\ 
				\hline
				\begin{tabular}{c}
					{\tiny \bf Biased independent sampling}\\ \scriptsize{\tiny This paper}
				\end{tabular}	& Def.\ \ref{def_biased_sampling_no_replacement} & $\frac{\max_i\{L_i\}}{\min_{i}{p_i}}$ & $0$ & $2A\Delta^{*} + s^2$ & $\min_i\left\lbrace p_i\right\rbrace$ & $0$\\ 					
				\hline
				\begin{tabular}{c}
					{\tiny \bf Distributed general biased rounding}\\ \scriptsize{\tiny This paper}
				\end{tabular}				 &  Def. \ref{def_distributed_biased_rounding} & $ A_{r}$ & $B_r$ & $C_r$ & $ b_{r}$ & $  c_r$\\ 					
				\hline
				\begin{tabular}{c}
					{{\tiny \bf Top-$k$}}\\ \scriptsize{\citep{aji2017sparse, alistarh2018sparse}}
				\end{tabular}				 & Def.\ \ref{def_top_ell} & $0$ & $1$ & $0$ & $\frac{k}{d}$ & $0$\\ 					
				\hline
				\begin{tabular}{c}
					{{\tiny \bf Rand-$k$}}\\ \scriptsize{\citep{Stich-EF-NIPS2018}}
				\end{tabular}				 & Def. \ref{def_random_l} & $0$  & $\frac{d}{k}$ & $0$ & $1$ & $0$ \\ 					
				\hline
				\begin{tabular}{c}
					{{\tiny \bf Biased Rand-$k$}}\\ \scriptsize{\citep{BezHorRichSaf}}
				\end{tabular}				 & Def. \ref{def_biased_random_l} & $0$  & $\frac{k}{d}$ & $0$ & $\frac{k}{d}$ & $0$ \\ 					
				\hline		
				\begin{tabular}{c}
					{{\tiny \bf Adaptive random sparsification}}\\ \scriptsize{\citep{BezHorRichSaf}}
				\end{tabular}				 & Def. \ref{def_adaptive_random_sparsification} & $0$  & $1$ & $0$ & $\frac{1}{d}$ & $0$ \\ 	
				\hline
				\begin{tabular}{c}
					{{\tiny \bf General unbiased rounding}}\\ \scriptsize{\citep{BezHorRichSaf}}
				\end{tabular}				 & Def. \ref{def_general_unbiased_rounding} & $0$  & $\frac{Z}{4}$ & $0$ & $1$ & $0$ \\ 	
				\hline
				\begin{tabular}{c}
					{{\tiny \bf General biased rounding}}\\ \scriptsize{\citep{BezHorRichSaf}}
				\end{tabular}				 & Def. \ref{def_general_biased_rounding} & $0$  & $F^2$ & $0$ & $\frac{G^2}{F}$ & $0$ \\ 	
				\hline
				\begin{tabular}{c}
					{{\tiny \bf Natural compression}}\\ \scriptsize{\citep{CNAT}}
				\end{tabular}				 & Def. \ref{def_natural_compression} & $0$  & $\frac{9}{8}$ & $0$ & $1$ & $0$ \\ 	
				\hline
				\begin{tabular}{c}
					{{\tiny \bf General exponential dithering}}\\ \scriptsize{\citep{BezHorRichSaf}}
				\end{tabular}				 & Def. \ref{def_general_exponential_dithering} & $0$  & $H_a$ & $0$ & $1$ & $0$ \\ 	
				\hline
				\begin{tabular}{c}
					{{\tiny \bf Natural dithering}}\\ \scriptsize{\citep{CNAT}}
				\end{tabular}				 & Def. \ref{def_natural_dithering} & $0$  & $H_2$ & $0$ & $1$ & $0$ \\ 	
				\hline
				\begin{tabular}{c}
					{{\tiny \bf Composition of Top-$k$  and exp dithering}}\\ \scriptsize{\citep{BezHorRichSaf}}
				\end{tabular}				 & Def. \ref{def_topk_exp_dithering} & $0$  & $H_a^2$ & $0$ & $\frac{k}{d H_a}$ & $0$ \\ 	
				\hline
				\begin{tabular}{c}
					{{\tiny \bf Gaussian smoothing}}\\ \scriptsize{\citep{Polyak}}
				\end{tabular}				 & Def. \ref{def_gaussian_smoothing} & $A_{GS}$  & $B_{GS}$ & $C_{GS}$ & $ b_{GS}$ & $  c_{GS}$ \\ 	
				\hline
				\begin{tabular}{c}
					{{\tiny \bf Hard-threshold sparsifier}}\\ \scriptsize{\citep{sahu2021rethinking}}
				\end{tabular}				 & Def. \ref{def_hard_threshold_sparsifier} & $0$  & $1$ & $0$ & $1$ & $w^2d$ \\ 	
				\hline
				\begin{tabular}{c}
					{{\tiny \bf Scaled integer rounding}}\\ \scriptsize{\citep{SapCanHoNelKalKinKrisMoshPorRich}}
				\end{tabular}				 & Def. \ref{def_scaled_rounding} & $0$  & $2$ & $\frac{2d}{\chi^2}$ & $\frac{1}{2}$ & $\frac{d}{2\chi^2}$ \\ 	
				\hline
				\begin{tabular}{c}
					{{\tiny \bf Biased dithering}}\\ \scriptsize{\citep{khirirat2018distributed}}
				\end{tabular}				 & Def. \ref{def_biased_dithering} & $0$  & $d$ & $0$ & $1$ & $0$ \\ 	
				\hline
				\begin{tabular}{c}
					{{\tiny \bf Sign compression}}\\ \scriptsize{\citep{karimireddy2019ef}}
				\end{tabular}				 & Def. \ref{def_sign_compression} & $0$  & $4-\frac{2}{d}$ & $0$ & $\frac{1}{2d}$ & $0$ \\ 	
				\hline
			\end{tabular}
		\end{threeparttable}
		\caption{Summary of the estimators with respective parameters $A$, $B$, $C$, $ b$ and $  c,$ satisfying our general {Biased ABC} framework. Constants $L_i$ are from Assumption~\ref{ass_smooth_functionwise}, $\Delta^{*}$ is defined in \eqref{eq_def_delta_star}, $A_r,B_r,C_r, b_{r},  c_r$ are defined in \eqref{eq_A_distributed_rounding}--\eqref{eq_chi_distributed_rounding}, $Z$ is defined in \eqref{eq_rounding_constant}, $F$ and $G$ are defined in \eqref{eq_biased_rounding_consts}, $H_a$ is defined in \eqref{eq_dithering_constant}, $A_{GS},B_{GS},C_{GS}, b_{GS},  c_{GS}$ are defined in \eqref{eq_gaussian_smoothing_consts}.} 
		\label{tab_estimators_full}  
	\end{table*}
	
	\begin{table*}[h]
		\centering
		\scriptsize
		\begin{threeparttable}
			\begin{tabular}{|l|c c c c c c c c c|}
				\hline
				Name of an estimator $\backslash$ Assumption  & A\ref{ass_first_set} & A\ref{ass_second_set} & A\ref{ass_third_set} & A\ref{ass_BV} & A\ref{ass_breq} & A\ref{ass_stich_decomposition} & A\ref{ass_abs_compr} & A\ref{ass_first_and_second_mmt_limits}  & A\ref{ass_scalar_ABC} \\
				\hline
				\hline
				\begin{tabular}{l}
					{\tiny \bf Biased independent sampling}
					\scriptsize{\tiny [This paper]}
				\end{tabular}				 & {\color{red}\ding{55}}  & {\color{red}\ding{55}}  & {\color{red}\ding{55}}  & {\color{red}\ding{55}}  & {\color{red}\ding{55}}  & {\color{red}\ding{55}}  & {\color{red}\ding{55}} & {\color{red}\ding{55}} & {\color{green}\checkmark}\\ 					
				\hline
				\begin{tabular}{l}
					{\tiny \bf Distributed general biased rounding} \scriptsize{\tiny [This paper]}
				\end{tabular}				 & {\color{red}\ding{55}} & {\color{red}\ding{55}}  & {\color{red}\ding{55}}  & {\color{red}\ding{55}}  & {\color{red}\ding{55}}  & {\color{red}\ding{55}}  & {\color{red}\ding{55}} & {\color{red}\ding{55}} & {\color{green}\checkmark}\ \\ 					
				\hline
				\begin{tabular}{l}
					{{\tiny\bf Top-$k$ sparsification}}
					\scriptsize{\citep{aji2017sparse, alistarh2018sparse}}
				\end{tabular}				 & {\color{green}\checkmark} & {\color{green}\checkmark} & {\color{green}\checkmark} & {\color{green}\checkmark} & {\color{green}\checkmark} & {\color{green}\checkmark} & {\color{red}\ding{55}} & {\color{green}\checkmark} & {\color{green}\checkmark} \\ 					
				\hline
				\begin{tabular}{l}
					{{\tiny\bf Rand-$k$}}
					\scriptsize{\citep{Stich-EF-NIPS2018}}
				\end{tabular}				 & {\color{green}\checkmark} & {\color{green}\checkmark}  & {\color{red}\ding{55}} & {\color{green}\checkmark} & {\color{red}\ding{55}} & {\color{green}\checkmark} & {\color{red}\ding{55}} & {\color{green}\checkmark} & {\color{green}\checkmark}\\ 					
				\hline
				\begin{tabular}{l}
					{{\tiny\bf Biased Random-$k$}}
					\scriptsize{\citep{BezHorRichSaf}}
				\end{tabular}				 & {\color{green}\checkmark} & {\color{green}\checkmark} & {\color{green}\checkmark} & {\color{green}\checkmark} & {\color{red}\ding{55}} & {\color{green}\checkmark} & {\color{red}\ding{55}} & {\color{green}\checkmark} & {\color{green}\checkmark}\\ 					
				\hline
				\begin{tabular}{l}
					{{\tiny\bf Adaptive random sparsification}}
					\scriptsize{\citep{BezHorRichSaf}}
				\end{tabular}				 & {\color{green}\checkmark} & {\color{green}\checkmark} & {\color{green}\checkmark} & {\color{green}\checkmark} & {\color{red}\ding{55}} & {\color{green}\checkmark} & {\color{red}\ding{55}} & {\color{green}\checkmark} & {\color{green}\checkmark}\\ 					
				\hline
				\begin{tabular}{l}
					{{\tiny\bf General unbiased rounding}}
					\scriptsize{\citep{BezHorRichSaf}}
				\end{tabular}				 & {\color{green}\checkmark} & {\color{green}\checkmark} & {\color{red}\ding{55}} & {\color{green}\checkmark} & {\color{red}\ding{55}} & {\color{green}\checkmark} & {\color{red}\ding{55}}  & {\color{green}\checkmark} & {\color{green}\checkmark}\\ 					
				\hline
				\begin{tabular}{l}
					{{\tiny\bf General biased rounding}}
					\scriptsize{\citep{BezHorRichSaf}}
				\end{tabular}				 & {\color{green}\checkmark} & {\color{green}\checkmark} & {\color{green}\checkmark} & {\color{green}\checkmark} & {\color{green}\checkmark} & {\color{green}\checkmark} & {\color{red}\ding{55}} & {\color{green}\checkmark} & {\color{green}\checkmark}\\ 					
				\hline
				\begin{tabular}{l}
					{{\tiny\bf Natural compression}}
					\scriptsize{\citep{CNAT}}
				\end{tabular}				 & {\color{green}\checkmark} & {\color{green}\checkmark} & {\color{green}\checkmark} & {\color{green}\checkmark} & {\color{red}\ding{55}}  & {\color{green}\checkmark} & {\color{red}\ding{55}}  & {\color{green}\checkmark} & {\color{green}\checkmark}\\ 					
				\hline
				\begin{tabular}{l}
					{{\tiny\bf General exponential dithering}}
					\scriptsize{\citep{BezHorRichSaf}}
				\end{tabular}				 & {\color{green}\checkmark} & {\color{green}\checkmark} & {\color{green}\checkmark} & {\color{green}\checkmark} & {\color{red}\ding{55}} & {\color{green}\checkmark} & {\color{red}\ding{55}} & {\color{green}\checkmark} & {\color{green}\checkmark}\\ 					
				\hline
				\begin{tabular}{l}
					{{\tiny\bf Natural dithering}}
					\scriptsize{\citep{CNAT}}
				\end{tabular}				 & {\color{green}\checkmark} & {\color{green}\checkmark} & {\color{green}\checkmark} & {\color{green}\checkmark} & {\color{red}\ding{55}} & {\color{green}\checkmark} & {\color{red}\ding{55}} & {\color{green}\checkmark} & {\color{green}\checkmark}\\ 				
				\hline
				\begin{tabular}{l}
					{{\tiny\bf Composition of Top-$k$  and exp dithering}}
					\scriptsize{\citep{BezHorRichSaf}}
				\end{tabular}				 & {\color{green}\checkmark} & {\color{green}\checkmark} & {\color{green}\checkmark} & {\color{green}\checkmark} & {\color{red}\ding{55}} & {\color{green}\checkmark} & {\color{red}\ding{55}} & {\color{green}\checkmark} & {\color{green}\checkmark}\\ 					
				\hline
				\begin{tabular}{l}
					{{\tiny\bf Gaussian smoothing}}
					\scriptsize{\citep{Polyak}}
				\end{tabular}				 & {\color{red}\ding{55}} & {\color{red}\ding{55}} & {\color{red}\ding{55}} & {\color{red}\ding{55}} & {\color{red}\ding{55}} & {\color{green}\checkmark} & {\color{red}\ding{55}} & {\color{red}\ding{55}} & {\color{green}\checkmark}\\ 					
				\hline
				\begin{tabular}{l}
					{{\tiny\bf Hard-threshold sparsifier}}
					\scriptsize{\citep{sahu2021rethinking}}
				\end{tabular}				 & {\color{green}\checkmark}  & {\color{green}\checkmark}  & {\color{green}\checkmark}  & {\color{green}\checkmark}  & {\color{green}\checkmark} & {\color{green}\checkmark} & {\color{green}\checkmark} & {\color{green}\checkmark} & {\color{green}\checkmark}\\ 					
				\hline
				\begin{tabular}{l}
					{{\tiny\bf Scaled integer rounding}}
					\scriptsize{\citep{SapCanHoNelKalKinKrisMoshPorRich}}
				\end{tabular}				 & {\color{green}\checkmark}  & {\color{green}\checkmark}  & {\color{red}\ding{55}}  & {\color{green}\checkmark}   & {\color{green}\checkmark}  & {\color{green}\checkmark} & {\color{green}\checkmark} & {\color{green}\checkmark} & {\color{green}\checkmark}\\ 					
				\hline
				\begin{tabular}{l}
					{{\tiny\bf Biased dithering}}
					\scriptsize{\citep{khirirat2018distributed}}
				\end{tabular}				 & {\color{green}\checkmark} & {\color{green}\checkmark} & {\color{red}\ding{55}}  & {\color{red}\ding{55}}  & {\color{green}\checkmark} & {\color{red}\ding{55}} & {\color{red}\ding{55}}  & {\color{green}\checkmark} & {\color{green}\checkmark}\\ 					
				\hline
				\begin{tabular}{l}
					{{\tiny \bf Sign compression}}
					\scriptsize{\citep{karimireddy2019ef}}
				\end{tabular}				 & {\color{green}\checkmark} & {\color{green}\checkmark} & {\color{green}\checkmark} & {\color{green}\checkmark} & {\color{green}\checkmark} & {\color{green}\checkmark} & {\color{red}\ding{55}} & {\color{green}\checkmark} & {\color{green}\checkmark}\\ 					
				\hline
			\end{tabular}
		\end{threeparttable}
		\caption{Summary on an inclusion of popular estimators into every known framework.}
		\label{tab_estimators_in_assumptions_full}    
	\end{table*}
	
	\section{Relations between assumptions \ref{ass_first_set}--\ref{ass_scalar_ABC}}
	\subsection{Counterexamples to Figure~\ref{fig_diagram}}\label{section_counterexamples_in_diagram}
	In Section~\ref{sect_existing_models} of the main part of the paper we outlined Theorem~\ref{thm_diagram_counterexamples} in an informal way. Below we state it rigorously.
	
	\noindent\textbf{Theorem 1} \;(Formal) {\it The following relations hold:
		\begin{enumerate}[label=\roman*, wide, labelwidth=!, labelindent=5pt]
			\item\label{item_no_implicatiom_contractive_abs} There is a minimization problem for which Assumption~\ref{ass_third_set} is satisfied, but Assumption~\ref{ass_abs_compr} is not. That is, (\hyperlink{CON}{CON}) does not imply (\hyperlink{ABS}{ABS}). The reverse implication also does not hold true.
			
			\item\label{item_no_implication_breq_contractive} There is a minimization problem for which Assumption~\ref{ass_third_set} is satisfied, but Assumption~\ref{ass_breq} is not. That is, (\hyperlink{CON}{CON}) does not imply (\hyperlink{BREQ}{BREQ}). The reverse implication also does not hold true.
			
			\item\label{item_no_implication_breq_abs} There is a minimization problem for which Assumption~\ref{ass_breq} is satisfied, but Assumption~\ref{ass_abs_compr} is not. That is, (\hyperlink{BREQ}{BREQ}) does not imply (\hyperlink{ABS}{ABS}). The reverse implication also does not hold true.

			\item\label{item_no_implication_breq_stich} There is a minimization problem for which Assumption~\ref{ass_breq} is satisfied, but Assumption~\ref{ass_stich_decomposition} is not. That is, (\hyperlink{BREQ}{BREQ}) does not imply (\hyperlink{BND}{BND}). The reverse implication also does not hold true.
			
			\item\label{item_no_implication_first_bnd} There is a minimization problem for which Assumption~\ref{ass_first_set} is satisfied, but Assumption~\ref{ass_stich_decomposition} is not. That is, (\hyperlink{SG1}{SG1}) does not imply (\hyperlink{BND}{BND}). The reverse implication also does not hold true.
			
			\item\label{item_no_implication_fsml_abs} There is a minimization problem for which Assumption \ref{ass_abs_compr} is satisfied, but Assumption \ref{ass_first_and_second_mmt_limits} is not. That is, (\hyperlink{ABS}{ABS}) does not imply (\hyperlink{FLSML}{FSML}). The reverse implication also does not hold true.
		\end{enumerate}
	}
	
	Clearly, this theorem implies that there is a mutual abscence of implications between Assumption~\ref{ass_abs_compr}~(\hyperlink{ABS}{ABS}) and Assumption~\ref{ass_BV} (\hyperlink{BVD}{BVD}), Assumption~\ref{ass_abs_compr}~(\hyperlink{ABS}{ABS}) and Assumption~\ref{ass_first_set}~(\hyperlink{SG1}{SG1}), Assumption~\ref{ass_abs_compr}~(\hyperlink{ABS}{ABS}) and Assumption~\ref{ass_second_set}~(\hyperlink{SG2}{SG2}), Assumption~\ref{ass_BV}~(\hyperlink{BVD}{BVD}) and Assumption~\ref{ass_breq}~(\hyperlink{BREQ}{BREQ}).
	
	\noindent\textbf{Proof of Theorem \ref{thm_diagram_counterexamples}}
	Let us prove all of the assertions stated above in Theorem \ref{thm_diagram_counterexamples} one by one.
	
	\noindent\ref{item_no_implicatiom_contractive_abs} Consider $f(x) = x^2,$ $g(x) = \frac{3}{2}\nabla f(x) = 3x.$ We have
	\begin{equation}\label{eq_bv_no_ac}
		\begin{split}
			\mathbb{E}\left[\left\|g(x) - \nabla f(x)\right\|^2\right] &= \left\|\frac{1}{2}\nabla f(x)\right\|^2\\
			&= x^2,
		\end{split}
	\end{equation}
	which implies due to \eqref{eq_bv_decomposition} that
	\begin{equation}\label{eq_noise_no_ac}
		\left\|\mathbb{E}\left[g(x)\right] - \nabla f(x)\right\|^2 \leq x^2,
	\end{equation}
	\begin{equation}\label{eq_bias_no_ac}
		\mathbb{E}\left[\left\|g(x) - \mathbb{E}\left[g(x)\right]\right\|^2\right] \leq x^2.
	\end{equation}
	Clearly, the estimator satisfies Assumption \ref{ass_third_set} with $\delta=\frac{4}{3}.$
	
	Clearly, the right-hand side of \eqref{eq_bv_no_ac} can not be bounded by any constant $\Delta^2,$ for all $x\in\mathbb{R}.$ Therefore, $g(x)$ does not satisfy Assumption \ref{ass_abs_compr}.
	
	Let us show that the reverse implication does not hold as well.
	
	Let $f(x)=x^2,$ $x\in\mathbb{R}.$ Let $g(x) = 2x + 1.$ Then $g(x)$ satisfies Assumptions~\ref{ass_abs_compr}. Indeed, 
	\begin{equation}\label{eq_counterex_parabola_1}
		\Exp{\left\|g(x)-\Exp{g(x)}\right\|^2} = 0,
	\end{equation}
	\begin{equation}\label{eq_counterex_parabola_2}
		\left\|\Exp{g(x)} - \nabla f(x)\right\|^2 = 1,
	\end{equation}
	which means that, due to \eqref{eq_bv_decomposition}, we have $\Exp{\left\|g(x)-\Exp{g(x)}\right\|^2}=1$, and we can choose $\Delta^2=1.$ 
	
	However, there is no $\delta > 0,$ such that $\Exp{\left\|g(x)-\Exp{g(x)}\right\|^2}=1$ can be bounded from above by $\left(1-\frac{1}{\delta}\right)\left\|\nabla f(x)\right\|^2=4\left(1-\frac{1}{\delta}\right)x^2,$ for all $x\in\mathbb{R}.$ Therefore, $g(x)$ does not satisfy Assumption~\ref{ass_third_set}.\\
	
	\noindent\ref{item_no_implication_breq_contractive} The implication does not hold trivially, since Assumption \ref{ass_breq} is formulated for deterministic estimators only.
	
	Let us show that the reverse implication does not hold as well.
	
	Suppose $g(x)=3\nabla f(x)$ is a deterministic gradient estimator of $f(x)$ with $\left\|\nabla f(x)\right\|^2$ unbounded from above by a constant. Then $g(x)$ satisfies Assumption \ref{ass_breq}. Indeed, we have
	\begin{equation*}
		\langle g(x), \nabla f(x) \rangle = 3\left\|\nabla f(x)\right\|^2,
	\end{equation*}
	\begin{equation*}
		\left\|g(x)\right\|^2 = 9 \left\|\nabla f(x)\right\|^2.
	\end{equation*}
	It means that we can choose $\rho=3,$ $\zeta=9.$ However, since we have
	\begin{equation*}
		\left\|\mathbb{E}\left[g(x)\right] - \nabla f(x)\right\|^2 = 4\left\|\nabla f(x)\right\|^2,
	\end{equation*}
	and the variance is $0$ ($g(x)$ is deterministice), there is no $\delta > 0,$ such that 
	$$
	\Exp{\left\|g(x)-\Exp{g(x)}\right\|^2}=4\left\|\nabla f(x)\right\|^2
	$$
	can be bounded from above by $\left(1-\frac{1}{\delta}\right)\left\|\nabla f(x)\right\|^2,$ for all $x\in\mathbb{R}.$ Therefore, $g(x)$ does not satisfy Assumption~\ref{ass_third_set}.\\
	
	\noindent\ref{item_no_implication_breq_abs} Consider the example of the problem and the estimator from the proof of Theorem~\ref{thm_diagram_counterexamples}--\ref{item_no_implicatiom_contractive_abs}. Let $f(x) = x^2,$ $g(x) = \frac{3}{2}\nabla f(x) = 3x.$ We have
	\begin{equation*}
		\langle g(x), \nabla f(x) \rangle = 6x^2,\quad \left\|g(x)\right\|^2 = 9x^2,
	\end{equation*}
	which means that this estimator satisfies Assumption \ref{ass_breq} with $\rho = \frac{3}{2},$ $\zeta=\frac{9}{4}.$
	
	Clearly, the right-hand side of \eqref{eq_bv_no_ac} can not be bounded by any constant $\Delta^2,$ for all $x\in\mathbb{R}.$ Therefore, $g(x)$ does not satisfy Assumption \ref{ass_abs_compr}.
	
	The reverse implication does not hold trivially, since Assumption \ref{ass_breq} is formulated for deterministic estimators only.\\
	
	\noindent\ref{item_no_implication_breq_stich}
	Suppose $g(x)=3\nabla f(x)$ is a deterministic gradient estimator of $f(x)$ with $\left\|\nabla f(x)\right\|^2$ unbounded from above by a constant. In the proof of Theorem~\ref{thm_diagram_counterexamples}--\ref{item_no_implication_breq_contractive} we showed that $g(x)$ satisfies Assumption~\ref{ass_breq} with $\rho=3,$ $\zeta=9.$ However, since we have
	\begin{equation*}
		\left\|\mathbb{E}\left[g(x)\right] - \nabla f(x)\right\|^2 = 4\left\|\nabla f(x)\right\|^2,
	\end{equation*}
	we are not able to find $0\leq m\leq 1$ and $\varphi^2\geq 0,$ such that
	\begin{equation*}
		\left\|\mathbb{E}\left[g(x)\right] - \nabla f(x)\right\|^2 \leq \eta\left\|\nabla f(x)\right\|^2 + \varphi^2,
	\end{equation*}
	for all $x\in\mathbb{R}^d.$
	Therefore, $g(x)$ does not satisfy Assumption~\ref{ass_BV}.
	
	The reverse implication does not hold trivially, since Assumption \ref{ass_breq} is formulated for deterministic estimators only.\\
	
	\noindent\ref{item_no_implication_first_bnd} Recall the stochastic estimator from Definition \ref{def_general_unbiased_rounding}.
	
	Suppose $g(x)$ is a general unbiased rounding estimator multiplied by a factor of $3.$ Suppose that $\left\|\nabla f(x)\right\|^2$ is not bounded from above. The estimator $g(x)$ is biased:
	\begin{equation*}
		\mathbb{E}\left[g(x)\right] = 3\nabla f(x).
	\end{equation*}
	Therefore,
	\begin{equation}\label{eq_scaled_rounding_bias}
		\left\|\mathbb{E}\left[g(x)\right] - \nabla f(x)\right\|^2 = 4\left\|\nabla f(x)\right\|^2.
	\end{equation}
	This biased estimator does not satisfy Assumption~\ref{ass_stich_decomposition} since there is no $0\leq m < 1,$ such that $\left\|\mathbb{E}\left[g(x)\right] - \nabla f(x)\right\|^2\leq m\left\|\nabla f(x)\right\|^2 + \varphi^2.$
	
	Without loss of generality we assume that $x\geq 0.$
	\begin{equation}\label{eq_scaled_rounding_variance}
		\begin{split}
			\mathbb{E}\left[\left\|g(x) - \mathbb{E}\left[g(x)\right]\right\|^2\right] & = \mathbb{E}\left[\left\|g(x)\right\|^2\right] - 9\left\|\nabla f(x)\right\|^2\\
			& = \left(\frac{9}{4}\sup_{k\in\mathbb{N}}\left(\frac{a_k}{a_{k+1}}+\frac{a_{k+1}}{a_k}+2\right)-9\right)\left\|\nabla f(x)\right\|^2\\
			&\geq 0.
		\end{split}
	\end{equation}
	
	Observe that $\langle \mathbb{E}\left[g(x)\right], \nabla f(x)\rangle = 3\left\|\nabla f(x)\right\|^2.$ It means that the gradient estimator satisfies Assumption~\ref{ass_first_set} with $\alpha = \frac{9Z}{4},$ $\beta = \frac{3Z}{4},$ where $Z$ is defined in \eqref{eq_rounding_constant}.
	
	Let us show that the reverse implication does not hold as well.
	
	As in the proof of Theorem~\ref{thm_diagram_counterexamples}--\ref{item_no_implicatiom_contractive_abs}, let $f(x)=x^2,$ $x\in\mathbb{R},$ $g(x) = 2x + 1.$ From \eqref{eq_counterex_parabola_1} and \eqref{eq_counterex_parabola_2}, we conclude that $g(x)$ satisfies Assumptions~\ref{ass_stich_decomposition} with $M=\sigma^2=m=0,$ $\varphi^2 = 1.$
	
	However, there is no constant $\frac{\alpha}{\beta}\geq 0,$ such that a function
	\begin{equation*}
		\langle \mathbb{E}\left[g(x)\right], \nabla f(x)\rangle = 2x(2x+1)
	\end{equation*}
	can be bounded from below by
	\begin{equation*}
		\frac{\alpha}{\beta}\left\|\nabla f(x)\right\|^2=\frac{\alpha}{\beta}4x^2,
	\end{equation*}
	for all $x.$ Therefore, $g(x)$ does not satisfy Assumption~\ref{ass_first_set}.\\
	
	\ref{item_no_implication_fsml_abs} Let $f(x)=x^2,$ $x\in\mathbb{R},$ $g(x) = 2x + 1.$ In the proof of Theorem \ref{thm_diagram_counterexamples}--\ref{item_no_implicatiom_contractive_abs} we showed that $g(x)$ satisfies Assumption \ref{ass_abs_compr}. However, $g(x)$ does not satisfy Assumption \ref{ass_first_and_second_mmt_limits}. There is no constant $q\geq 0,$ such that a function
	$$
	\langle\Exp{g(x)}, \nabla f(x)\rangle = 2x(2x+1)
	$$
	can be bounded from below by
	$$
	q\left\|\nabla f(x)\right\|^2 = 4qx^2,
	$$
	for all $x.$ Therefore, $g(x)$ does not satisfy Assumption \ref{ass_first_and_second_mmt_limits}.
	
	Let us show that the reverse implication does not hold as well.
	
	Suppose $g(x)$ is a general unbiased rounding estimator (see Definition \ref{def_general_unbiased_rounding}) multiplied by a factor of $3.$ Suppose that $\left\|\nabla f(x)\right\|^2$ is not bounded from above. This estimator satisfies Assumption \ref{ass_first_and_second_mmt_limits}. Indeed, observe that $\langle \mathbb{E}\left[g(x)\right], \nabla f(x)\rangle = 3\left\|\nabla f(x)\right\|^2.$ Also, $\left\|\Exp{g(x)}\right\|^2 = 9 \left\|\nabla f(x)\right\|^2.$ Therefore, we can choose $q=u=3,$ $U = Z - 9,$ $Q=0.$ 
	
	Due to \eqref{eq_bv_decomposition}, \eqref{eq_scaled_rounding_bias} and \eqref{eq_scaled_rounding_variance}, we have
	\begin{equation*}
		\begin{split}
			\Exp{\left\|g(x) - \nabla f(x)\right\|^2} &= 4\left\|\nabla f(x)\right\|^2 + \left(\frac{9}{4}\sup_{k\in\mathbb{N}}\left(\frac{a_k}{a_{k+1}}+\frac{a_{k+1}}{a_k}+2\right)-9\right)\left\|\nabla f(x)\right\|^2\\
			& \geq 4\left\|\nabla f(x)\right\|^2.
		\end{split}
	\end{equation*}
	Then $g(x)$ does not satisfy Assumption~\ref{ass_abs_compr} since there is no $\Delta\geq 0,$ such that $4\left\|\nabla f(x)\right\|^2\leq \Delta^2$ holds, for all $x\in\mathbb{R}^d.$
	\begin{flushright}
		$\blacksquare$
	\end{flushright}
	\subsection{Implications in Figure~\ref{fig_diagram}}\label{section_thm_abc_proof}
	In Section~\ref{section_ass_weakest} of the main part of the paper we outlined Theorem~\ref{thm_informal_abc} in an informal way. Below we state it rigorously.
	
	\noindent\textbf{Theorem 2} \;(Formal) {\it
		Let Assumption \ref{ass_smooth} hold for the function $f.$ Then the following relations hold:
		\begin{enumerate}[label=\roman*, wide, labelwidth=!, labelindent=5pt]
			\item\label{item_bvd_from_contr}  Suppose a gradient estimator $g(x)$ satisfies Assumption~\ref{ass_third_set}. Then $g(x)$ satisfies Assumption~\ref{ass_BV} with $\eta=1-\frac{1}{\delta},$ $\xi=1-\frac{1}{\delta}.$ That is, $\left(\hyperlink{CON}{CON}\right)$ implies $\left(\hyperlink{BVD}{BVD}\right).$ The reverse implication does not hold.
			\item\label{item_bnd_from_bvd} Suppose a gradient estimator $g(x)$ satisfies Assumption~\ref{ass_BV}. Then $g(x)$ satisfies Assumption~\ref{ass_stich_decomposition} with $m=\eta,$ $\varphi^2=0,$ $M = \frac{2\xi(1+\eta)}{(1-\eta)^2},$ $\sigma^2=0.$ That is, $\left(\hyperlink{BVD}{BVD}\right)$ implies $\left(\hyperlink{BND}{BND}\right).$ The reverse implication does not hold.
			\item\label{item_bnd_from_ac} Suppose a gradient estimator $g(x)$ satisfies Assumption~\ref{ass_abs_compr}. Then $g(x)$ satisfies Assumption \ref{ass_stich_decomposition} with $M=m=0,$ $\sigma^2=\varphi^2=\Delta^2.$ That is, $\left(\hyperlink{ABS}{ABS}\right)$ implies $\left(\hyperlink{BND}{BND}\right).$ The reverse implication does not hold.
			\item\label{item_first_from_bvd} Suppose a gradient estimator $g(x)$ satisfies Assumption~\ref{ass_BV}. Then $g(x)$ satisfies Assumption~\ref{ass_first_set} with $\alpha = \frac{\left(1-\eta\right)^2}{2(1+\eta)},$ $\beta = \frac{2}{1-\eta}\max\{\xi, 2\xi+\eta - 1\}.$ That is, $\left(\hyperlink{BVD}{BVD}\right)$ implies $\left(\hyperlink{SG1}{\text{SG1}}\right).$  The reverse implication does not hold.
			\item\label{item_first_from_breq} Suppose a gradient estimator $g(x)$ satisfies Assumption~\ref{ass_breq}. Then $g(x)$ satisfies Assumption~\ref{ass_first_set}. That is, $\left(\hyperlink{BREQ}{BREQ}\right)$ implies $\left(\hyperlink{SG1}{\text{SG1}}\right).$  The reverse implication does not hold.
			\item\label{item_first_two_equiv} Assumption \ref{ass_first_set} $\left(\hyperlink{SG1}{\text{SG1}}\right)$ is equivalent to Assumption~\ref{ass_second_set} $\left(\hyperlink{SG2}{\text{SG2}}\right).$
			\item\label{item_fsml_from_first} Suppose a gradient estimator $g(x)$ satisfies Assumption~\ref{ass_first_set}. Then $g(x)$ satisfies Assumption~\ref{ass_first_and_second_mmt_limits} with $u=U=\beta^2,$ $Q=0,$ $q=\frac{\alpha}{\beta}.$ That is, (\hyperlink{SG1}{SG1}) implies (\hyperlink{FSML}{FSML}). The reverse implication does not hold.
			\item\label{item_abc_from_fsml} Suppose a gradient estimator $g(x)$ satisfies Assumption~\ref{ass_first_and_second_mmt_limits}. Then $g(x)$ satisfies Assumption~\ref{ass_scalar_ABC} with $A=0,$ $B=U+u^2,$ $C=Q,$ $ b=q,$ $  c=0.$ That is, $\left(\hyperlink{FSML}{\text{FSML}}\right)$ implies $\left(\hyperlink{Biased ABC}{Biased \; ABC}\right).$ The reverse implication does not hold.
			\item\label{item_abc_from_bnd} Suppose a gradient estimator $g(x)$ satisfies Assumption~\ref{ass_stich_decomposition}. Then $g(x)$ satisfies Assumption~\ref{ass_scalar_ABC} with $A=0,$ $B = 2(M+1)(m+1),$ $C = 2(M+1)\varphi^2 +  \sigma^2,$ $ b=\frac{1-m}{2},$ $  c=\frac{\varphi^2}{2}.$ That is, $\left(\hyperlink{BND}{BND}\right)$ implies $\left(\hyperlink{Biased ABC}{Biased \; ABC}\right).$  The reverse implication does not hold.
		\end{enumerate}
	}
	
	\noindent\textbf{Proof of Theorem \ref{thm_informal_abc}} Let us prove all of the assertions stated above in Theorem \ref{thm_informal_abc} one by one.\\
	\noindent\ref{item_bvd_from_contr}. From \eqref{eq_contractive_biased} and from \eqref{eq_bv_decomposition}, we easily derive the following inequalities:
	\begin{equation*}
		\left\|\mathbb{E}\left[g(x)\right] - \nabla f(x)\right\|^2\leq\left(1-\frac{1}{\delta}\right)\left\|\nabla f(x)\right\|^2,
	\end{equation*}
	and
	\begin{equation*}
		\begin{split}
			\mathbb{E}\left[\left\|g(x) - \mathbb{E}\left[g(x)\right]\right\|^2\right] &\leq \left(1-\frac{1}{\delta}\right)\left\|\nabla f(x)\right\|^2.\\
		\end{split}
	\end{equation*}
	Therefore, we can choose $\eta=1-\frac{1}{\delta},$ $\xi=1-\frac{1}{\delta}.$\\
	
	Next, let us show that the reverse implication does not hold. Suppose $g(x)$ is a gradient estimator of the following form:
	\begin{equation*}
		g(x) = \nabla f(x) + X, \text{ where }
		X = \begin{cases}
			4\;\nabla f(x), & \text{with probability }\frac{1}{4}\\
			0, & \text{with probability }\frac{3}{4}.
		\end{cases}
	\end{equation*}
	For the estimator $g(x)$ we have
	\begin{equation*}
		\left\|\mathbb{E}\left[g(x)\right]-\nabla f(x)\right\|^2 = \left\|\nabla f(x)\right\|^2,
	\end{equation*}
	and
	\begin{equation*}
		\mathbb{E}\left[\left\|g(x) - \mathbb{E}\left[g(x)\right]\right\|^2\right] = \mathbb{E}\left[\left\|X\right\|^2\right] - \left\|\mathbb{E}\left[X\right]\right\|^2= 3\left\|\nabla f(x)\right\|^2.
	\end{equation*}
	We can choose $\eta = 1,$ $\xi = 3,$ so $g(x)$ satisfies Assumption \ref{ass_BV}. But there is no $\delta\geq 1,$ such that, for all $x\in\mathbb{R}^d,$ 
	\begin{equation*}
		\mathbb{E}\left[\left\|g(x) - \nabla f(x)\right\|^2\right] \stackrel{\eqref{eq_bv_decomposition}}{=} \mathbb{E}\left[\left\|g(x) - \mathbb{E}\left[g(x)\right]\right\|^2\right] + \left\|\mathbb{E}\left[g(x)\right]-\nabla f(x)\right\|^2 = 4\left\|\nabla f(x) \right\|^2
	\end{equation*}
	does not exceed $\left(1 - \frac{1}{\delta}\right)\left\|\nabla f(x)\right\|^2.$ Then $g(x)$ does not satisfy Assumption \ref{ass_third_set}.\\
	
	\noindent\ref{item_bnd_from_bvd}. Since we know that
	\begin{equation}\label{eq_claim_stich_from_bv}
		\left\|\mathbb{E}\left[g(x)\right] - \nabla f(x)\right\|^2\leq \eta\left\|\nabla f(x)\right\|^2,
	\end{equation}
	we can choose $m=\eta$ and $\varphi^2=0.$ By Young's Inequality (Lemma \ref{lemma_young}, \eqref{eq_scalar_prod_young}), from \eqref{eq_claim_stich_from_bv} we derive that
	\begin{equation*}
		\begin{split}
			(1-\eta)\left\|\nabla f(x)\right\|^2 &\leq 2\langle \mathbb{E}[g(x)],\nabla f(x)\rangle - \left\|\mathbb{E}[g(x)]\right\|^2\\
			& \leq \frac{(1-\eta)\left\|\nabla f(x)\right\|^2}{2} + \frac{2\left\|\mathbb{E}[g(x)]\right\|^2}{(1-\eta)} - \left\|\mathbb{E}[g(x)]\right\|^2.
		\end{split}
	\end{equation*}
	Hence,
	\begin{equation*}
		\left\|\nabla f(x)\right\|^2  \leq \frac{2(1+\eta)}{\left(1-\eta\right)^2} \left\| \mathbb{E}\left[g(x)\right]\right\|^2.
	\end{equation*}
	Also, we know that
	\begin{equation*}
		\mathbb{E}\left[\left\|g(x) - \mathbb{E}\left[g(x)\right]\right\|^2\right] \leq \xi\left\|\nabla f(x)\right\|^2.
	\end{equation*}
	Therefore, we arrive at
	\begin{equation*}
		\begin{split}
			\mathbb{E}\left[\left\|g(x) - \mathbb{E}\left[g(x)\right]\right\|^2\right] & \leq \frac{2\xi(1+\eta)}{\left(1-\eta\right)^2} \left\| \mathbb{E}\left[g(x)\right]\right\|^2.
		\end{split}
	\end{equation*}
	We can choose $M = \frac{2\xi(1+\eta)}{\left(1-\eta\right)^2},$ $\sigma^2=0.$\\
	
	Next, let us show that the reverse implication does not hold. As in the proof of Theorem \ref{thm_diagram_counterexamples}--\ref{item_no_implicatiom_contractive_abs}, let $f(x)=x^2,$ $x\in\mathbb{R}.$ Let $g(x) = 2x + 1.$ From \eqref{eq_counterex_parabola_1} and \eqref{eq_counterex_parabola_2}, we conclude that $g(x)$ satisfies Assumption~\ref{ass_stich_decomposition} with $M=\sigma^2=m=0,$ $\varphi^2 = 1.$
	
	However, there is no $0\leq\eta\leq 1,$ such that $\left\|\mathbb{E}\left[g(x)\right] - \nabla f(x)\right\|^2 = 1$ is bounded from above by $\xi\left\|\nabla f(x)\right\|^2=4\eta x^2,$ for all $x\in\mathbb{R}.$ It means that Assumption \ref{ass_BV} does not hold.\\
	
	\ref{item_bnd_from_ac} 	Indeed, \eqref{eq_bv_decomposition} and \eqref{eq_abs_compr} imply $\mathbb{E}\left[\|g(x)-\mathbb{E}\left[g(x)\right]\|^2\right] \leq \Delta^2$ and $\|\mathbb{E}\left[g(x)\right]-\nabla f(x)\|^2 \leq \Delta^2.$ Therefore, Assumption~\ref{ass_stich_decomposition} is satisfied with $M=m=0,$ $\sigma^2=\varphi^2=\Delta^2.$\\
	
	Next, let us prove that the reverse implication does not hold. Consider the example of the problem and the estimator from the proof of Theorem \ref{thm_diagram_counterexamples}--\ref{item_no_implication_breq_abs}. From \eqref{eq_noise_no_ac} and \eqref{eq_bias_no_ac} we conclude that the estimator satisfies Assumption~\ref{ass_stich_decomposition} with $M=\frac{1}{9},$ $m=\frac{1}{4},$ but Assumption \ref{ass_abs_compr} is not satisfied.
	
	\ref{item_first_from_bvd}. Since we know that
	\begin{equation}\label{eq_claim_first_set_from_bv} 
		\left\|\mathbb{E}\left[g(x)\right] - \nabla f(x)\right\|^2\leq \eta\left\|\nabla f(x)\right\|^2,
	\end{equation}
	we obtain
	\begin{equation}
		\left(1-\eta\right)\left\|\nabla f(x)\right\|^2\leq 2\langle \mathbb{E}\left[g(x)\right], \nabla f(x)\rangle - \left\|\mathbb{E}\left[g(x)\right]\right\|^2.
	\end{equation}
	Then
	\begin{equation*}
		\begin{split}
			\mathbb{E}\left[\left\|g(x) - \mathbb{E}\left[g(x)\right]\right\|^2\right] &\leq \xi\left\|\nabla f(x)\right\|^2\\
			& \leq\frac{2\xi}{1-\eta}\langle \mathbb{E}\left[g(x)\right], \nabla f(x) \rangle - \frac{\xi}{1-\eta}\left\|\mathbb{E}\left[g(x)\right]\right\|^2.
		\end{split}
	\end{equation*}
	If $\xi+\eta\leq 1,$  we obtain that
	\begin{equation*}
		\mathbb{E}\left[\left\|g(x)\right\|^2\right] \leq \frac{2\xi}{1-\eta}\langle \mathbb{E}\left[g(x)\right], \nabla f(x) \rangle.
	\end{equation*}
	Otherwise,
	\begin{equation*}
		\begin{split}
			\mathbb{E}\left[\left\|g(x)\right\|^2\right] &\leq \frac{2\xi}{1-\eta}\langle \mathbb{E}\left[g(x)\right], \nabla f(x) \rangle + \left(\frac{\xi}{1-\eta} - 1\right)\left\|\mathbb{E}\left[g(x)\right]\right\|^2\\
			& \leq \frac{2(2\xi+\eta-1)}{1-\eta}\langle \mathbb{E}\left[g(x)\right], \nabla f(x) \rangle .
		\end{split}
	\end{equation*}
	Hence, we can choose $\beta = \frac{2}{1-\eta}\max\{\xi, 2\xi+\eta - 1\}.$ Further, by Young's Inequality (Lemma~\ref{lemma_young},~\eqref{eq_scalar_prod_young}), from \eqref{eq_claim_stich_from_bv} we derive that
	\begin{equation*}
		\begin{split}
			(1-\eta)\left\|\nabla f(x)\right\|^2 &\leq 2\langle \mathbb{E}[g(x)],\nabla f(x)\rangle - \left\|\mathbb{E}[g(x)]\right\|^2\\
			& \leq \frac{(1-\eta)\left\|\nabla f(x)\right\|^2}{2} + \frac{2\left\|\mathbb{E}[g(x)]\right\|^2}{(1-\eta)} - \left\|\mathbb{E}[g(x)]\right\|^2.
		\end{split}
	\end{equation*}
	Then we have
	\begin{equation*}
		\left\|\mathbb{E}\left[g(x)\right]\right\|^2\geq \frac{\left(1-\eta\right)^2}{2(1+\eta)} \left\|\nabla f(x)\right\|^2.
	\end{equation*}
	Therefore, we can choose $\alpha = \frac{\left(1-\eta\right)^2}{2(1+\eta)}.$\\
	
	Let us show that the inverse implication does not hold. 
	
	Consider the problem and the estimator from the proof of Theorem \ref{thm_diagram_counterexamples}--\ref{item_no_implication_first_bnd}.	Since $\left\|\nabla f(x)\right\|^2$ is not bounded from above, this estimator does not satisfy Assumption \ref{ass_BV}: there is no $0\leq\eta\leq 1$ such that 
	$$
	\left\|\mathbb{E}\left[g(x)\right] - \nabla f(x)\right\|^2\leq\eta\left\|\nabla f(x)\right\|^2.
	$$ 
	
	However, recall that Assumption \ref{ass_first_set} is satisfied with $\alpha = Z,$ $\beta = \frac{Z}{3},$ where $Z$ is defined in \eqref{eq_rounding_constant}.\\
	
	\noindent\ref{item_first_from_breq}. Observe that
	\begin{equation*}
		\left\|\nabla f(x)\right\|^2 \leq \frac{1}{\rho}\langle g(x), \nabla f(x) \rangle.
	\end{equation*}
	Therefore,
	\begin{equation*}
		\left\|g(x)\right\|^2 \leq \zeta\left\|\nabla f(x)\right\|^2 \leq \frac{\zeta}{\rho}\langle g(x), \nabla f(x) \rangle,
	\end{equation*}
	and we can choose $\beta = \frac{\zeta}{\rho}$ in Assumption \ref{ass_first_set}. By Young's Inequality (Lemma~\ref{lemma_young},~\eqref{eq_scalar_prod_young}), we have
	\begin{equation*}
		\begin{split}
			\rho\left\|\nabla f(x)\right\|^2 & \leq \langle g(x), \nabla f(x) \rangle\\
			&  \leq \frac{\left\|g(x)\right\|^2}{2\rho} + \frac{\rho\left\|\nabla f(x)\right\|^2}{2}.\\
		\end{split}
	\end{equation*}
	This implies that $\left\|g(x)\right\|^2 \geq \rho^2\left\|\nabla f(x)\right\|^2,$ and we can choose $\alpha=\rho^2$ in Assumption \ref{ass_first_set}.\\
	
	The reverse implication does not hold. Since Assumption \ref{ass_breq} is formulated for deterministic estimators only, any stochastic estimator that satisfies Assumption \ref{ass_first_set} does not satisfy Assumption \ref{ass_breq}.\\
	
	\noindent\ref{item_first_two_equiv}. It follows from assertions 1 and 2 of Theorem \ref{thm_gradients_equivalence}.\\
	
	\ref{item_fsml_from_first} Recall that Assumption \ref{ass_first_set} implies \eqref{eq_implication}. Since $\left\|\Exp{g(x)}\right\|^2\leq \Exp{\left\|g(x)\right\|^2},$ we can choose $u = \beta.$ From $\langle \Exp{g(x)}, \nabla f(x)\rangle \geq \alpha\left\|\nabla f(x)\right\|,$ we conclude that $q$ can be set to $\frac{\alpha}{\beta}.$ Furthermore, $\Exp{\left\|g(x) - \Exp{g(x)}\right\|^2}\leq\Exp{\left\|g(x)\right\|^2}$ and \eqref{eq_implication} imply that we can put $U$ equal to $\beta^2,$ $Q=0.$ Note, that Theorem \ref{thm_gradients_equivalence} states that $\beta^2\geq\alpha.$ Therefore, the requirement $q\leq u$ from Assumption \ref{ass_first_and_second_mmt_limits} is also satisfied.
	
	Let us prove that the reverse implication does not hold. For every $x\in\mathbb{R},$ consider $f(x) = x^3,$ $g(x) = Y\nabla f(x) + Z,$ where $Y$ is a random variable with $\mathrm{Bern}\left(\frac{1}{2}\right)$ distribution, independent of a random variable $Z$ that attains values $\pm 1$ with equal probability. First, we establish relations \eqref{eq_fsml_scalar}, \eqref{eq_fsml_first_mmt} and \eqref{eq_fsml_variance} in this setting:
	\begin{equation*}
		\langle \nabla f(x), \Exp{g(x)} \rangle = \frac{1}{2}\left\|\nabla f(x)\right\|^2 = \frac{9}{2}x^4,
	\end{equation*}
	\begin{equation*}
		\left\|\Exp{g(x)} \right\|^2 = \frac{1}{4}\left\|\nabla f(x)\right\|^2 = \frac{9}{4}x^4,
	\end{equation*}
	\begin{equation*}
		\begin{split}
			\Exp{\left\|g(x)\right\|^2} - \left\|\Exp{g(x)}\right\|^2 & = \Exp{Y^2\left\|\nabla f(x)\right\|^2 + 2YZ\nabla f(x) + Z^2} - \frac{1}{4}\left\|\nabla f(x)\right\|^2\\
			& = \frac{1}{2}\left\|\nabla f(x)\right\|^2 + 1 - \frac{1}{4}\left\|\nabla f(x)\right\|^2\\
			& = \frac{1}{4}\left\|\nabla f(x)\right\|^2 + 1\\
			& = \frac{9}{4}x^4 + 1.
		\end{split}
	\end{equation*}
	This implies, that $g(x)$ satisfies Assumption \ref{ass_first_and_second_mmt_limits} with $q=u=\frac{1}{2},$ $U = \frac{1}{4}$ and $Q=1.$
	
	Consider the implication \eqref{eq_implication} from Assumption \ref{ass_first_set}. Notice, that 
	$$
	\Exp{\left\|g(x)\right\|^2} = \frac{1}{2}\left\|\nabla f(x)\right\|^2 + 1 = \frac{9}{2}x^4 + 1,
	$$
	and it can not be bounded from above by $\beta^2\left\|\nabla f(x)\right\|^2=9\beta^2x^4,$ for all $x\in\mathbb{R}.$ Therefore,~\eqref{eq_implication} does not hold, which means that Assumption \ref{ass_first_set} also does not hold.\\
	
	\noindent\ref{item_abc_from_fsml}. Suppose $g(x)$ satisfies Assumption \ref{ass_first_and_second_mmt_limits}.
	
	From \eqref{eq_fsml_scalar}, we conclude that $ b$ can be chosen as $q,$ $  c$ can be chosen as $0.$ Further, \eqref{eq_fsml_variance} implies that
	$$
	\Exp{\left\|g(x)\right\|^2} \leq U\left\|\nabla f(x)\right\|^2 + \left\|\Exp{g(x)}\right\|^2 + Q.
	$$
	From \eqref{eq_fsml_first_mmt}, we obtain that
	$$
	\Exp{\left\|g(x)\right\|^2} \leq \left(U + u^2\right)\left\|\nabla f(x)\right\|^2 + Q.
	$$
	Therefore, we can choose $A=0,$ $B = U + u^2,$ $C = Q.$\\
	
	Next, let us prove that the reverse implication does not hold. Recall that any gradient estimator that satisfies Assumptions \ref{ass_first_set} should also satisfy \eqref{eq_implication}. Let $f(x) = f_1(x) + f_2(x)=0,$ $f_1(x) = \frac{x_1^2}{2} - \frac{x_2^2}{2},$ $f_2(x) = -\frac{x_1^2}{2}+ \frac{x_2^2}{2}.$ Consider $g(x) = \tilde{g}(x) + \textbf{X}$ from Definition~\ref{def_biased_sampling_no_replacement} with $p_1=p_2=\frac{1}{3}.$ Due to Claim~\ref{claim_sampling_no_division}, it satisfies Assumption~\ref{ass_scalar_ABC} (functions $f_1$ and $f_2$ are $1$-smooth). First, we determine distributions of random variables $\frac{\mathbb{I}_1}{|S|},$ $\frac{\mathbb{I}_1^2}{|S|^2},$ $\frac{\mathbb{I}_1\mathbb{I}_2}{|S|^2}$ (see Table~\ref{table_distribution_of_sampling}).
	\begin{table}
		\begin{center}
			\begin{tabular}{|c|c|c|c|c|}
				\hline
				$\mathbb{I}_1$ & $1$ & $0$ & $0$ & $1$ \\
				\hline 
				$\mathbb{I}_2$ & $0$ & $1$ & $0$ & $1$ \\
				\hline
				$\frac{\mathbb{I}_1}{|S|}$ & $1$ & $0$ & $0$ & $\frac{1}{2}$\\
				\hline
				$\frac{\mathbb{I}_1^2}{|S|^2}$ & $1$ & $0$ & $0$ & $\frac{1}{4}$\\
				\hline
				$\frac{\mathbb{I}_1\mathbb{I}_2}{|S|^2}$ & $0$ & $0$ & $0$ & $\frac{1}{4}$\\
				\hline
				\text{probability} & $\frac{2}{9}$ & $\frac{2}{9}$ & $\frac{4}{9}$ & $\frac{1}{9}$\\
				\hline
			\end{tabular}
			\caption{Distributions of random variables $\frac{\mathbb{I}_1}{|S|},$ $\frac{\mathbb{I}_1^2}{|S|^2},$ $\frac{\mathbb{I}_1\mathbb{I}_2}{|S|^2},$ counterexample for Theorem~\ref{thm_informal_abc}~-~\ref{item_abc_from_fsml}}
			\label{table_distribution_of_sampling}
		\end{center}
	\end{table}
	
	Let us calculate the second moment of this stochastic estimator:
	\begin{equation}\label{eq_secondmmt_counterex_sampling}
		\begin{split}
			\mathbb{E}\left[\left\|g(x)\right\|^2\right] &= \mathbb{E}\left[\left\|\tilde{g}(x)\right\|^2\right] + \mathbb{E}\left[\left\|\textbf{X}\right\|^2\right]\\
			&=\mathbb{E}\left[\frac{\mathbb{I}_1^2}{|S|^2}\right]\nabla f_1(x) + \mathbb{E}\left[\frac{\mathbb{I}_2^2}{|S|^2}\right]\nabla f_2(x)\\
			& + 2\mathbb{E}\left[\frac{\mathbb{I}_1\mathbb{I}_2}{|S|^2}\right]\langle\nabla f_1(x), \nabla f_2(x)\rangle + s^2\\
			& = \frac{1}{4}\left(x_1^2+x_2^2\right) + \frac{1}{4}\left(x_1^2 + x_2^2\right)\\
			& -\frac{2}{36}(x_1^2+x_2^2) + s^2 \\
			&= \frac{4}{9}\left(x_1^2+x_2^2\right) + s^2.\\
		\end{split}
	\end{equation}
	Further, since $\mathbb{E}\left[\frac{\mathbb{I}_1}{|S|}\right] = \mathbb{E}\left[\frac{\mathbb{I}_2}{|S|}\right] = \frac{5}{18}$ (see Table \ref{table_distribution_of_sampling}), we have
	\begin{equation}\label{eq_firstmmt_counterex_sampling}
		\left\|\Exp{g(x)}\right\|^2 = \left\|\frac{5}{18}\left(\nabla f_1(x) + \nabla f_2(x)\right)\right\|^2=0.
	\end{equation}
	Hence, the variance of $g(x)$ coincides with its second moment. Clearly, the second moment~\eqref{eq_secondmmt_counterex_sampling} can not be bounded from above by $U\left\|\nabla f(x)\right\|^2 + Q = Q,$ for all $x.$ Therefore, this gradient estimator does not satisfy Assumption~\ref{ass_first_and_second_mmt_limits}.	
	
	\noindent\ref{item_abc_from_bnd}. First, we bound the second moment of $g(x):$
	\begin{equation*}
		\begin{split}
			\mathbb{E}\left[\left\|g(x)\right\|^2\right] &= \mathbb{E}\left[\left\|g(x) - \mathbb{E}\left[g(x)\right]\right\|^2\right] + \left\|\mathbb{E}\left[g(x)\right]\right\|^2\\
			& = \mathbb{E}\left\|\mathcal{N}(x, Y)\right\|^2 + \left\|\nabla f(x) + b(x)\right\|^2\\
			& \leq \left(M+1\right)\left\|\nabla f(x) + b(x)\right\|^2 + \sigma^2\\
			& \leq 2\left(M+1\right)\left\|\nabla f(x)\right\|^2 + 2\left(M+1\right)\left\|b(x)\right\|^2+\sigma^2\\
			&\leq 2\left(M+1\right)\left(m+1\right)\left\|\nabla f(x)\right\|^2 + 2\left(M+1\right)\varphi^2 + \sigma^2.\\
		\end{split}
	\end{equation*}
	
	We can choose $A=0,$ $B = 2(M+1)(m+1),$ $C = 2(M+1)\varphi^2 + \sigma^2$ in Assumption~\ref{ass_scalar_ABC}. Further, note that \eqref{eq_bias_stich_simplified} can be rewritten in an equivalent way in terms of the lower bound on the scalar product:
	\begin{equation}\label{eq_equiv_stich_scalar}
		\begin{split}
			\langle\nabla f(x), \mathbb{E}\left[g(x)\right]\rangle& = \frac{\left\|\nabla f(x)\right\|^2}{2}+\frac{\left\|\mathbb{E}\left[g(x)\right]\right\|^2}{2} - \frac{\left\|\mathbb{E}\left[g(x)\right]-\nabla f(x) \right\|^2}{2}\\
			& \geq \frac{1-m}{2}\left\|\nabla f(x)\right\|^2 + \frac{\left\|\mathbb{E}\left[g(x)\right]\right\|^2}{2} - \frac{\varphi^2}{2}.
		\end{split}
	\end{equation}
	Therefore,
	\begin{equation}\label{eq_beat_stich}
		\langle\nabla f(x), \mathbb{E}\left[g(x)\right]\rangle \geq \frac{1-m}{2}\left\|\nabla f(x)\right\|^2 - \frac{\varphi^2}{2}.
	\end{equation}
	Observe that in \eqref{eq_beat_stich} we used only a trivial lower bound of $0$ on $\mathbb{E}\left[g(x)\right],$ which signifies that our assumption on scalar product \eqref{eq_scalar_prod} is less restrictive than the Assumption~\ref{eq_bias_stich_simplified} on the bias term.\\
	
	Let us prove that the reverse implication does not hold. Consider the problem and the estimator from the proof of Theorem \ref{thm_diagram_counterexamples}--\ref{item_abc_from_fsml}. From \eqref{eq_secondmmt_counterex_sampling} and \eqref{eq_firstmmt_counterex_sampling}, we obtain that 
	$$
	\mathbb{E}\left[\left\|g(x) - \mathbb{E}[g(x)]\right\|^2\right] = \mathbb{E}\left[\left\|g(x)\right\|^2\right]=\frac{4}{9}\left(x_1^2+x_2^2\right) + s^2.
	$$
	Observe that it can not be bounded from above by $M\left\|\mathbb{E}[g(x)]\right\|^2+\sigma^2=\sigma^2,$ for all $x.$ Hence, it does not satisfy Assumption~\ref{ass_stich_decomposition}.
	\begin{flushright}
		$\blacksquare$
	\end{flushright}
	
	\textbf{Proof of Claim \ref{claim_sampling_counterex_no_stich_but_abc}}
	Let $p_1=p_2=\frac{1}{3}$ be probabilities. For every $i\in\{1,2\},$ define a random set as follows:
	$$
	S_i = 
	\begin{cases}
		\{i\} & \text{with probability } p_i,\\
		\varnothing & \text{with probability } 1-p_i.
	\end{cases}
	$$
	Define a random subset $S\subseteq \{1, 2\}$ by taking the union of these random sets: 
	$$
	S\eqdef S_1\cup S_2.
	$$
	For every $i\in\{1,2\},$ define $v_i=\frac{\mathbb{I}_{i\in S}}{p_i^2}.$ Let 
	\begin{equation*}
		g(x) = \frac{1}{2}\sum_{i=1}^nv_i\nabla f_i(x).
	\end{equation*}
	
	Consider $f(x) = \frac{1}{2}\left(f_1(x)+f_2(x)\right),$ where $f_1(x)=x_1^2,$  $f_2(x)=x_2^2.$ For the introduced stochastic gradient, we have
	\begin{equation}\label{eq_scalar_counterex}
		\begin{split}
			\langle\mathbb{E}\left[g(x)\right],\nabla f(x) \rangle &= 3\left(x_1^2+x_2^2\right).
		\end{split}
	\end{equation}
	Therefore, $g(x)$ satisfies \eqref{eq_scalar_prod} of Assumption \ref{ass_scalar_ABC} with $ b=3,$ $  c=0.$ Observe that
	\begin{equation}\label{eq_sampling_second_mmt}
		\begin{split}
			\mathbb{E}\left[\left\|g(x)\right\|^2\right] = 27\left(x_1^2+x_2^2\right).
		\end{split}
	\end{equation}
	Therefore, $g(x)$ also satisfies \eqref{eq_ABC} with $A = 0, B = 27, C=0.$
	
	Recall that inequality \eqref{eq_bias_stich_simplified} of Assumption \ref{ass_stich_decomposition} is equivalent to \eqref{eq_equiv_stich_scalar}.
	
	Since $\left\|\mathbb{E}\left[g(x)\right]\right\|^2=9\left(x_1^2+x_2^2\right),$ the right-hand side of~\eqref{eq_equiv_stich_scalar} is equal to
	\begin{equation*}
		\frac{10-m}{2}\left(x_1^2+x_2^2\right) - \frac{\varphi^2}{2},
	\end{equation*}
	$0\leq m < 1,$ $\varphi^2\geq 0.$ This expression can not bound \eqref{eq_scalar_counterex} from below, for all $x=(x_1, x_2)\in\mathbb{R}^2.$ Hence, this gradient estimator does not satisfy \eqref{eq_bias_stich_simplified} of Assumption \ref{ass_stich_decomposition}.
	\begin{flushright}
		$\blacksquare$
	\end{flushright}
	
	\section{General nonconvex case: history and corollaries from Theorem~\ref{thm_nonconvex}}
	In Section~\ref{section_noncvx_main} we have formulated Theorem~\ref{thm_nonconvex} on convergence of \algname{BiasedSGD} under \hyperlink{Biased ABC}{Biased ABC} assumption and compared the rate obtained to the known convergence results in nonconvex case. Below we present recent results, derive several corollaries from Theorem~\ref{thm_nonconvex} and make a formal comparison of our results to the known results.
	
	\subsection{Known results} Convergence of \algname{BiasedSGD} in general smooth case has been studied in several papers. The next two results are Lemma~3 and Theorem~4 from \citep{AjallStich}. We formulate them as a theorem and its corollary respectively.
	\begin{theorem}
		Under Assumptions~\ref{ass_smooth}~and~\ref{ass_stich_decomposition}, and for any stepsize $\gamma\leq\frac{1}{(M+1)L},$ it holds after $T$ steps of \algname{BiasedSGD} that
		\begin{equation*}
			\frac{1}{T}\sum_{t=0}^{T-1}\Exp{\left\|\nabla f(x^t)\right\|^2}\leq \frac{2\delta^0}{T\gamma (1-m)} + \frac{\gamma L\sigma^2}{1-m} + \frac{\varphi^2}{1-m}.
		\end{equation*}
	\end{theorem}
	\begin{corollary}\label{cor_stich_noncvx}
		Under Assumptions~\ref{ass_smooth}~and~\ref{ass_stich_decomposition}, and by choosing the stepsize $\gamma=\min\left\lbrace \frac{1}{(M+1)L}, \frac{\varepsilon(1-m)}{2L\sigma^2} \right\rbrace,$ for $\varepsilon>0,$ we have that
		\begin{equation*}
			T = \mathcal{O}\left(\max\left\lbrace \frac{4(M+1)}{\varepsilon (1-m)}, \frac{8\sigma^2}{\varepsilon^2(1-m)^2}\right\rbrace L\delta^0\right)
		\end{equation*}
		iterations suffice to obtain $$\frac{1}{T}\sum_{t=0}^{T-1}\Exp{\left\|\nabla f(x^t)\right\|^2}=\mathcal{O}\left(\varepsilon + \frac{\varphi^2}{1-m}\right).$$
	\end{corollary}
	The convergence result that we get in Theorem~\ref{thm_nonconvex} is formulated in terms of minimum of expected squared gradient norms. However, in Corollary~\ref{cor_stich_noncvx} the convergence established not for the minimum, but for the mean of expected squared gradient norms. Since minimum is smaller than the mean, we can immediately restate Corollary~\ref{cor_stich_noncvx} in a slightly weaker form:
	\begin{corollary}\label{cor_stich_noncvx_minimum}
		Under Assumptions~\ref{ass_smooth}~and~\ref{ass_stich_decomposition}, and by choosing the stepsize $\gamma=\min\left\lbrace \frac{1}{(M+1)L}, \frac{\varepsilon(1-m)}{2L\sigma^2} \right\rbrace,$  for $\varepsilon>0,$ we have that
		\begin{equation*}
			T = \mathcal{O}\left(\max\left\lbrace \frac{4(M+1)}{\varepsilon (1-m)}, \frac{8\sigma^2}{\varepsilon^2(1-m)^2}\right\rbrace L\delta^0\right)
		\end{equation*}
		iterations suffice to obtain $$\min_{0\leq t \leq T-1}\Exp{\left\|\nabla f(x^t)\right\|^2}=\mathcal{O}\left(\varepsilon + \frac{\varphi^2}{1-m}\right).$$
	\end{corollary}
	The result below is Theorem~4.8 from \citep{BotCurNoce}.
	\begin{theorem}\label{thm_noncvx_bottou}
		Under Assumptions~\ref{ass_smooth}~and~\ref{ass_first_and_second_mmt_limits}, and for any stepsize $0<\gamma\leq\frac{q}{L(U+u^2)},$ for all $T\in\mathbb{N},$ the following inequality holds:
		\begin{equation*}
			\frac{1}{T}\sum_{t=0}^{T-1}\Exp{\left\|\nabla f(x^t)\right\|^2}\leq\frac{\gamma LQ}{q} + \frac{2\delta^0}{Tq\gamma}.
		\end{equation*}
	\end{theorem}
	To be able to make a further comparison of convergence rates, we need to establish the rate the above theorem yields. Once again, the convergence result that we get in Theorem~\ref{thm_nonconvex} is formulated in terms of minimum of expected squared gradient norms. However, in Corollary~\ref{thm_noncvx_bottou} the convergence established not for the minimum, but for the mean of expected squared gradient norms. Since minimum is smaller than the mean, we can immediately write the corollary in a slightly weaker form:
	\begin{corollary}\label{cor_noncvx_bottou_rate}
		For $\varepsilon>0,$ choose stepsize $\gamma>0$ as $\gamma=\min\left\lbrace \frac{\varepsilon q}{2LQ}, \frac{q}{L(U+u^2)} \right\rbrace.$ Then, if
		\begin{equation*}
			T \geq \max\left\lbrace \frac{8Q}{\varepsilon^2q^2}, \frac{4(U+u^2)}{\varepsilon q^2} \right\rbrace L\delta^0,
		\end{equation*}
		we have that $$\min_{0\leq t \leq T-1} \Exp{\left\|\nabla f(x^t)\right\|^2}\leq \varepsilon.$$
	\end{corollary}
	\subsection{Corollaries from Theorem~\ref{thm_nonconvex}}
	In general, Theorem~\ref{thm_nonconvex} guarantees the convergence towards some neghborhood of the $\varepsilon$-stationary point, that can not be made less than $\frac{  c}{ b}.$ Therefore, we have the following corollary.
	\begin{corollary}\label{cor_noncvx_1}
		Choose the stepsize $\gamma>0$ as $\gamma = \min\left\lbrace \frac{1}{\sqrt{LAT}} , \frac{b}{LB}, \frac{c}{LC}\right\rbrace.$ Then if
		\begin{equation*}
			T\geq\frac{6\delta^0L}{c}\max\left\lbrace \frac{B}{ b}, \frac{6\delta^0 A}{  c},\frac{C}{c} \right\rbrace,
		\end{equation*}
		we have $$\min_{0\leq t \leq T-1}\mathbb{E}\left[\left\|\nabla f(x^t)\right\|^2\right]\leq \frac{3c}{b}.$$
	\end{corollary}
	Next two corollaries are Theorem 2 and Corollary 1 from \citep{khaled2022better}. However, in that work the authors obtain these results in the unbiased case, i.e. when $\mathbb{E}\left[g(x)\right] = \nabla f(x)$ holds, for all $x\in\mathbb{R}^d.$ In our case we only require $\langle \mathbb{E}\left[g(x)\right], \nabla f(x)\rangle \geq \left\|\nabla f(x)\right\|^2$ to hold, for all $x\in\mathbb{R}^d.$
	\begin{corollary}\label{cor_noncvx_2}
		Suppose $  c=0,$ $ b=1.$ Choose the stepsize such that $0<\gamma\leq \frac{1}{LB}.$ Then the iterates $\{x^t\}_{t\geq 0}$ of \algname{BiasedSGD} (Algorithm \eqref{alg:SGD}) satisfy
		\begin{equation}\label{eq_convergence_simplified}
			\min_{0\leq t \leq T-1}\mathbb{E}\left[\left\|\nabla f(x^t)\right\|^2\right] \leq  \frac{2\left(1+LA\gamma^2\right)^T}{\gamma T}\delta^0+LC\gamma.
		\end{equation}
	\end{corollary}
	\begin{corollary}\label{cor_noncvx_3}
		Suppose $c=0$ and $b=1.$ Fix $\varepsilon>0.$ Choose the stepsize $\gamma>0$ as $\gamma = \min\left\lbrace \frac{1}{\sqrt{LAT}},\frac{1}{LB},\frac{\varepsilon}{2LC} \right\rbrace.$ Then, if
		\begin{equation*}
			T\geq \frac{12\delta^0 L}{\varepsilon^2}\max\left\lbrace B, \frac{12\delta^0 A}{\varepsilon^2},\frac{2C}{\varepsilon^2} \right\rbrace,
		\end{equation*}
		we have $$\min_{0\leq t \leq T-1}\mathbb{E}\left[\left\|\nabla f(x^t)\right\|\right] \leq \varepsilon.$$
	\end{corollary}
	The next corollary contains the result similar to the one obtained in Theorem~4 from \citep{AjallStich}. However, we impose weaker assumptions (compare \hyperlink{Biased ABC}{Biased ABC} and \hyperlink{BND}{BND} in Figure \ref{fig_diagram}; see also Claim~\ref{claim_sampling_counterex_no_stich_but_abc}).
	\begin{corollary}\label{cor_noncvx_stich}
		Suppose $A=0,$ $b\leq 1.$ Choose stepsize $\gamma>0$ as $\gamma=\min\left\lbrace \frac{b}{LB}, \frac{\varepsilon b}{2LC} \right\rbrace.$ Then, for $\varepsilon>0,$ we have that
		\begin{equation*}
			\mathcal{T} = \mathcal{O}\left(\max\left\lbrace \frac{8C}{b^2\varepsilon^2}, \frac{4B}{b^2\varepsilon} \right\rbrace L\delta^0\right)
		\end{equation*}
		iterations suffice for $$\min_{0\leq t \leq T-1}\Exp{\left\|\nabla f(x^t)\right\|^2} = \mathcal{O}\left(\varepsilon + \frac{c}{b}\right).$$
	\end{corollary}
	If we substitute $B$ for $2(M+1)(m+1),$ $C$ for $2(M+1)\varphi^2 +  \sigma^2,$ $ b$ for $\frac{1-m}{2},$ $  c$ for $\frac{\varphi^2}{2}$ in accordance with Theorem~\ref{thm_assumptions_in_our_framework} (see also Table~\ref{tab_assns_in_our_frame}), Corollary~\ref{cor_noncvx_stich} yields the rate of $\mathcal{O}\left(\max\left\lbrace \frac{8(M+1)(m+1)}{(1-m)^2\varepsilon},\frac{16(M+1)\varphi^2+2\sigma^2}{(1-m)^2\varepsilon^2} \right\rbrace L\delta^0\right)$ while Corollary~\ref{cor_stich_noncvx_minimum} (see Theorem~4 from \citep{AjallStich}) grants the rate of $\mathcal{T} = \mathcal{O}\left(\max\left\lbrace \frac{2\sigma^2}{(1-m)^2\varepsilon^2}, \frac{M+1}{(1-m)\varepsilon} \right\rbrace L\delta^0\right).$ Our result is worse by a factor of $\frac{1}{1-m}$ and by an additive term of $\mathcal{O}\left(\frac{(M+1)\varphi^2}{(1-m)^2\varepsilon^2}L\delta^0\right).$
	\begin{corollary}\label{cor_noncvx_bottou_follows}
		Suppose $A=c=0.$ For $\varepsilon>0,$ choose stepsize $\gamma = \min\left\lbrace \frac{b}{LB}, \frac{b\varepsilon}{LC} \right\rbrace.$ Then, if
		\begin{equation*}
			T\geq \max\left\lbrace \frac{8C}{\varepsilon^2b^2}, \frac{4B}{\varepsilon b^2} \right\rbrace L\delta^0, 
		\end{equation*}
		we have that $$\min_{0\leq t \leq T-1}\Exp{\left\|\nabla f(x^t)\right\|^2}\leq\varepsilon.$$
	\end{corollary}
	To recover the result from Corollary~\ref{cor_noncvx_bottou_rate}, one needs to substitute $B$ for $U+u^2,$ $C$ for $Q,$ $ b$ for $q$ in accordance with the representation of Assumption~\ref{ass_first_and_second_mmt_limits} in \hyperlink{Biased ABC}{Biased ABC} framework (see Theorem~\ref{thm_assumptions_in_our_framework} and Table~\ref{tab_assns_in_our_frame}).
	\subsection{Proof of Corollary~\ref{cor_noncvx_bottou_rate}}
	If $\gamma = \frac{\varepsilon q}{2LQ},$ and $T \geq \frac{8LQ\delta^0}{\varepsilon^2q^2},$ then we have that
	\begin{equation*}
		\frac{\gamma LQ}{q}\leq\frac{\varepsilon}{2},\quad \frac{2\delta^0}{Tq\gamma}=\frac{4LQ\delta^0}{T\varepsilon q^2} \leq\frac{\varepsilon}{2}.
	\end{equation*}
	If $\gamma = \frac{q}{L(U+u^2)}$ and $T\geq \frac{4L(U+u^2)\delta^0}{\varepsilon q^2},$ then we obtain that
	\begin{equation*}
		\frac{\gamma LQ}{q} \leq \frac{\varepsilon q}{2LQ}\cdot\frac{LQ}{q} \leq\frac{\varepsilon}{2},\qquad\frac{2\delta^0}{Tq\gamma}=\frac{2\delta^0L(U+u^2)}{Tq^2}\leq\frac{\varepsilon}{2}.
	\end{equation*}
	Therefore, we get that
	\begin{equation*}
		\min_{0\leq t \leq T-1}\Exp{\left\|\nabla f(x^t)\right\|^2}\leq\varepsilon.
	\end{equation*}
	\begin{flushright}
		$\blacksquare$
	\end{flushright}
	\subsection{Key lemma}
	Our main convergence result in the nonconvex scenario relies on the following key lemma.
	\begin{lemma}\label{lemma_weights}
		Let Assumptions \ref{ass_smooth} and \ref{ass_scalar_ABC}. Choose stepsize $\gamma$ satisfying
		\begin{equation}\label{eq_gamma_noncvx}
			0<\gamma\leq\frac{ b}{LB}.
		\end{equation}
		Then, for any $T\geq 1,$ the iterates $\{x^t\}$ of Algorithm \ref{alg:SGD} satisfy
		\begin{equation*}
			\frac{b}{2}\sum_{t=0}^{T-1}w_tr^t\leq \frac{w_{-1}}{\gamma}\delta^0 - \frac{w_{T-1}}{\gamma}\delta^{T} + \frac{LC\gamma+  c}{2}\sum_{t=0}^{T-1}w_t.
		\end{equation*}
	\end{lemma}
	
	\noindent\textbf{Proof of Lemma \ref{lemma_weights}} From Assumption \ref{ass_smooth} we have
	\begin{equation}\label{eq_l_smooth}
		\begin{split}
			f(x^{t+1})  & \leq f(x^t) + \langle\nabla f(x^t), x^{t+1} - x^t\rangle + \frac{L}{2}\left\|x^{t+1} - x^t\right\|^2\\
			& = f(x^t) - \gamma\langle \nabla f(x^t), g^t \rangle + \frac{L\gamma^2}{2}\left\|g^t\right\|^2.\\
		\end{split}
	\end{equation}
	
	Let us take expectation of both sides of \eqref{eq_l_smooth} conditioned on $x^t$ and apply Assumption \ref{ass_scalar_ABC}:
	\begin{equation}\label{eq_supplementary}
		\begin{split}
			\mathbb{E}\left[f(x^{t+1}) |x^t\right] & \leq f(x^t) - \gamma  b\left\|f(x^t)\right\|^2 +   c\gamma\\
			& + \frac{L\gamma^2}{2}\left(2 A (f(x^t) - f^{*}) + B\left\|\nabla f(x^t)\right\|^2 + C\right)\\
			& = f(x^t) - \gamma \left(b - \frac{LB\gamma}{2}\right)\left\|\nabla f(x^t)\right\|^2\\
			& + LA\gamma^2\left(f(x^t) - f^{*}\right) + \frac{LC\gamma^2}{2} +   c\gamma.
		\end{split}
	\end{equation}
	Subtract $f^{*}$ from both sides. Take  expectation on both sides and use the tower property. For every $t\geq 0,$ put $\delta^t\eqdef\mathbb{E}\left[f(x^t)-f^{*}\right]$ and $r^t\eqdef\mathbb{E}\left[\left\|\nabla f(x^t)\right\|^2\right].$ We obtain that
	\begin{equation*}
		\begin{split}
			\gamma\left( b - \frac{LB\gamma}{2}\right)r^t\leq \left(1+LA\gamma^2\right)\delta^t- \delta^{t+1} +\frac{LC\gamma^2}{2} +   c\gamma.
		\end{split}
	\end{equation*}
	Due to our choice of stepsize \eqref{eq_gamma_noncvx}, we obtain that
	\begin{equation}\label{eq_to_weigh}
		\begin{split}
			\frac{\gamma b}{2}r^t\leq \left(1+LA\gamma^2\right)\delta^t - \delta^{t+1} +\frac{LC\gamma^2}{2} +   c\gamma.
		\end{split}
	\end{equation}
	
	Fix $w_{-1}>0$ and, for all $t\geq 0,$ define $w_t=\frac{w_{t-1}}{1+LA\gamma^2}.$ Multiplying both sides of \eqref{eq_to_weigh} by $\frac{w_t}{\gamma},$ we obtain
	\begin{equation*}
		\frac{ b w_t r^t}{2}\leq\frac{w_{t-1}}{\gamma}\delta^t - \frac{w_t}{\gamma}\delta^{t+1} + \frac{LC\gamma w_t}{2}+\frac{c w_t}{2}.
	\end{equation*}
	For every $0\leq t\leq T-1,$ sum these inequalities. We arrive at
	\begin{equation}\label{eq_weighted}
		\frac{ b}{2}\sum_{t=0}^{T-1}w_tr^t\leq \frac{w_{-1}}{\gamma}\delta^0 - \frac{w_{T-1}}{\gamma}\delta^{T} + \frac{LC\gamma+  c}{2}\sum_{t=0}^{T-1}w_t.
	\end{equation}
	\begin{flushright}
		$\blacksquare$
	\end{flushright}
	\subsection{Proof of Theorem \ref{thm_nonconvex}}
	From \eqref{eq_weighted} we derive that
	\begin{equation}\label{eq_weighted_simplified}
		\frac{ b}{2}\sum_{t=0}^{T-1}w_tr^t\leq \frac{w_{-1}}{\gamma}\delta^0 + \frac{LC\gamma+  c}{2}\sum_{t=0}^{T-1}w_t.
	\end{equation}
	Observe that we can obtain the following lower bound on a sum of weights:
	\begin{equation*}
		\sum_{t=0}^{T-1}w_t\geq Tw_{T-1} = \frac{Tw_{-1}}{\left(1+LA\gamma^2\right)^T}.
	\end{equation*}
	Dividing both parts of \eqref{eq_weighted_simplified} by $\sum_{t=0}^{T-1}w_t$ and using the lower bound on it, we get the statement of Theorem \ref{thm_nonconvex}:
	\begin{equation*}
		\min_{0\leq t \leq T-1}r^t \leq   \frac{2\left(1+LA\gamma^2\right)^T}{ b\gamma T}\delta^0+\frac{LC\gamma}{b}+\frac{c}{b}.
	\end{equation*}
	\begin{flushright}
		$\blacksquare$
	\end{flushright}
	\subsection{Proof of Corollary \ref{cor_noncvx_1}}
	We bound each term in the right-hand side of \eqref{eq_convergence}	by $\frac{c}{ b}.$
	
	If $\gamma=\frac{1}{\sqrt{LAT}},$ and if $T\geq \frac{36\left(\delta^0\right)^2LA}{c^2},$ then we have
	\begin{equation*}
		\frac{2\left(1+LA\gamma^2\right)^T}{ b\gamma T}\delta^0 \leq \frac{6\delta^0\sqrt{LA}}{ b\sqrt{T}}\leq\frac{c}{ b}.
	\end{equation*}
	If $\gamma=\frac{ b}{LB},$ and if $T\geq \frac{6LB\delta^0}{bc},$ then we obtain
	\begin{equation*}
		\frac{2\left(1+LA\gamma^2\right)^T}{ b\gamma T}\delta^0 
		\leq \frac{6LB\delta^0}{ b T} \leq \frac{  c}{ b}.
	\end{equation*}
	If $\gamma=\frac{  c}{LC},$ and if $T\geq \frac{6LC\delta^0}{  c^2},$ then we obtain
	\begin{equation*}
		\frac{2\left(1+LA\gamma^2\right)^T}{\gamma T}\delta^0 
		\leq \frac{6LC\delta^0}{ b  c T} \leq \frac{c}{b}.
	\end{equation*}
	Due to the choice of $\gamma,$ we have $\frac{LC\gamma}{ b}\leq\frac{  c}{ b}.$ The last term is $\frac{  c}{ b}$ itself.
	
	Therefore, we obtain
	\begin{equation*}
		\min_{0\leq t \leq T-1}\mathbb{E}\left[\left\|\nabla f(x^t)\right\|^2\right] \leq \frac{3c}{ b}.
	\end{equation*}
	\begin{flushright}
		$\blacksquare$
	\end{flushright}
	\subsection{Proof of Corollary \ref{cor_noncvx_2}} The proof is easy: one needs to substitute $ b$ for $1$ and $  c$ for $0$ in \eqref{eq_convergence}.
	\begin{flushright}
		$\blacksquare$
	\end{flushright}
	\subsection{Proof of Corollary \ref{cor_noncvx_3}} 
	We bound each term in the right-hand side of \eqref{eq_convergence_simplified} by $\frac{\varepsilon^2}{2}.$
	
	If $\gamma=\frac{1}{\sqrt{LAT}},$ and if $T\geq \frac{144\left(\delta^0\right)^2LA}{\varepsilon^4},$ then we have
	\begin{equation*}
		\frac{2\left(1+LA\gamma^2\right)^T}{\gamma T}\delta^0 \leq \frac{6\delta^0\sqrt{LA}}{\sqrt{T}}\leq\frac{\varepsilon^2}{2}.
	\end{equation*}
	If $\gamma=\frac{1}{LB},$ and if $T\geq \frac{12LB\delta^0}{\varepsilon^2},$ then we obtain
	\begin{equation*}
		\frac{2\left(1+LA\gamma^2\right)^T}{\gamma T}\delta^0 
		\leq \frac{6LB\delta^0}{T} \leq \frac{\varepsilon^2}{2}.
	\end{equation*}
	If $\gamma=\frac{\varepsilon}{2LC},$ and if $T\geq \frac{24LC\delta^0}{\varepsilon^4},$ then we obtain
	\begin{equation*}
		\frac{2\left(1+LA\gamma^2\right)^T}{\gamma T}\delta^0 
		\leq \frac{12LC\delta^0}{\varepsilon^2 T} \leq \frac{\varepsilon^2}{2}.
	\end{equation*}
	Due to the choice of $\gamma,$ we have $LC\gamma\leq\frac{\varepsilon^2}{2}.$
	
	Therefore, we obtain
	\begin{equation*}
		\min_{0\leq t \leq T-1}\mathbb{E}\left[\left\|\nabla f(x^t)\right\|\right] \leq \varepsilon.
	\end{equation*}
	\begin{flushright}
		$\blacksquare$
	\end{flushright}
	\subsection{Proof of Corollary~\ref{cor_noncvx_stich}}
	When $A=0,$ from \eqref{eq_convergence} we have that
	$$
	\min_{0\leq t \leq T-1}r^t\leq \frac{2}{b\gamma T}\delta^0 + \frac{LC\gamma}{b} + \frac{c}{b}.
	$$
	If $\gamma = \frac{\varepsilon b}{2LC}$ and $T\geq\frac{8\delta^0 LC}{b^2\varepsilon^2},$ then we get that
	\begin{equation*}
		\frac{2}{b\gamma T}\delta^0 = \frac{4LC\delta^0}{b^2T\varepsilon}\leq\frac{\varepsilon}{2},\quad \frac{LC\gamma}{b}\leq\frac{\varepsilon}{2}.
	\end{equation*} 
	If $\gamma = \frac{b}{LB}$ and $T\geq\frac{4\delta^0LB}{b^2T},$ then we obtain that
	\begin{equation*}
		\frac{2}{b\gamma T}\delta^0 = \frac{2LB\delta^0}{b^2T}\leq\frac{\varepsilon}{2}, \qquad \frac{LC\gamma}{b} = \frac{LC}{b}\cdot\frac{b}{LB} \leq\frac{LC}{b}\cdot\frac{\varepsilon b}{2LC} = \frac{\varepsilon}{2}.
	\end{equation*}
	It follows that $\min_{0\leq t \leq T-1}r^t = \mathcal{O}\left(\varepsilon+\frac{c}{b}\right).$
	\begin{flushright}
		$\blacksquare$
	\end{flushright}
	\subsection{Proof of Corollary~\ref{cor_noncvx_bottou_follows}}
	It follows from \eqref{eq_convergence}, that when $A=c=0,$ holds
	\begin{equation*}
		\min\limits_{0\leq t \leq T-1}\mathbb{E}\left[\left\|\nabla f(x^t)\right\|^2\right] \leq  \frac{2\delta^0}{b\gamma T}	+\frac{LC\gamma}{b}.
	\end{equation*}
	If $\gamma=\frac{b\varepsilon}{2LC},$ and $T\geq \frac{8L\delta^0C}{b^2\varepsilon^2},$ then we have that
	\begin{equation*}
		\frac{LC\gamma}{b}\leq\frac{\varepsilon}{2}, \quad \frac{2\delta^0}{b\gamma T} = \frac{4\delta^0LC}{b^2\varepsilon T}\leq\frac{\varepsilon}{2}.
	\end{equation*}
	if $\gamma = \frac{b}{LB},$ and $T \geq \frac{4\delta^0LB}{b^2\varepsilon},$ then we obtain that
	\begin{equation*}
		\frac{LC\gamma}{b}\leq \frac{b\varepsilon}{2LC}\cdot\frac{LC}{b}=\frac{\varepsilon}{2},\qquad \frac{2\delta^0}{b\gamma T} = \frac{2\delta^0LB}{b^2T}\leq\frac{\varepsilon}{2}.
	\end{equation*} 
	It follows that $\min_{0\leq t \leq T-1}\Exp{\left\|\nabla f(x^t)\right\|^2} \leq \varepsilon.$
	\begin{flushright}
		$\blacksquare$
	\end{flushright}
	\section{Convergence under P\L-condition (assumption \ref{ass_pl})}
	In Section~\ref{section_pl_main} we have formulated Theorem~\ref{thm_noncvx_pl} on convergence of \algname{BiasedSGD} under \hyperlink{Biased ABC}{Biased ABC} assumption and compared the rate obtained to the known convergence results subject to P\L-condition. Below we present recent results, derive several corollaries from Theorem~\ref{thm_noncvx_pl} and make a formal comparison of our results to the known results.
	\subsection{Corollaries from Theorem~\ref{thm_noncvx_pl}}
	As before in the general nonconvex case, Theorem~\ref{thm_noncvx_pl} guarantees the convergence towards some neghborhood of the $\varepsilon$-stationary point, that can not be made less than $\frac{  c}{\mu b}.$ Therefore, we have the following corollary.
	\begin{corollary}\label{cor_pl_1}
		Choose stepsize $\gamma>0$ as $\gamma = \min\left\lbrace \frac{\mu b}{L(A+\mu B)}, \frac{1}{2\mu b}, \frac{2  c}{LC} \right\rbrace.$ Then, if
		\begin{equation*}
			T\geq \max\left\lbrace 2, \frac{L(A+\mu B)}{\mu^2 b^2}, \frac{LC}{2  c\mu b} \right\rbrace \log \frac{\mu b\delta^0}{  c},
		\end{equation*}
		we have $$\mathbb{E}\left[f(x^T) - f^{*}\right]\leq\frac{3  c}{\mu b}.$$
	\end{corollary}
	Without bias terms, we recover the best known rates under Polyak-- \L ojasiewicz condition (\citet{KarNutSch}) subject to milder conditions.
	\begin{corollary}\label{cor_pl_2}
		Suppose $  c=0.$ Choose the stepsize $\gamma > 0$ as $\gamma = \min\left\lbrace \frac{\mu b}{L(A+\mu B)}, \frac{1}{2\mu b}, \frac{\varepsilon\mu b}{LC} \right\rbrace.$ Then, if
		\begin{equation*}
			T\geq\max\left\lbrace 2, \frac{L(A+\mu B)}{\mu^2 b^2}, \frac{LC}{\varepsilon\mu^2 b^2} \right\rbrace \log \frac{2\delta^0}{\varepsilon},
		\end{equation*}
		we have $$\mathbb{E}\left[f(x^T) - f^{*}\right]\leq\varepsilon.$$
	\end{corollary}
	Plugging in $A=0,$ we recover the result similar to the one obtained in Theorem~6 of \citep{AjallStich}. However, we impose weaker assumptions (compare \hyperlink{Biased ABC}{Biased ABC} and \hyperlink{BND}{BND} in Figure \ref{fig_diagram}; see also Claim~\ref{claim_sampling_counterex_no_stich_but_abc}).
	\begin{corollary}\label{cor_pl_3}
		Suppose $A=0,$ $ b\leq 1.$ Choose stepsize $\gamma>0$ as $\gamma=\min\left\lbrace \frac{b}{BL}, \frac{\varepsilon\mu b + 2  c}{LC} \right\rbrace.$ Then, for $\varepsilon > 0,$ we have that
		\begin{equation*}
			\mathcal{T} = \mathcal{O}\left(\max\left\lbrace \frac{B}{ b}, \frac{C}{\varepsilon\mu b+2  c}\right\rbrace\frac{\kappa}{ b} \log\frac{2\delta^0}{\varepsilon}\right)
		\end{equation*}
		iterations suffice for $$\Exp{f(x^T) - f^{*}} = \mathcal{O}\left(\varepsilon + \frac{2  c}{\mu b}\right).$$ 
	\end{corollary}
	If we substitute $B$ for $2(M+1)(m+1),$ $C$ for $2(M+1)\varphi^2 +  \sigma^2,$ $ b$ for $\frac{1-m}{2},$ $  c$ for $\frac{\varphi^2}{2}$ in accordance with Theorem~\ref{thm_assumptions_in_our_framework} (see also Table~\ref{tab_assns_in_our_frame}), Corollary~\ref{cor_pl_3} yields the rate of $\mathcal{O}\left(\max\left\lbrace \frac{2(M+1)(m+1)}{1-m},\frac{2(M+1)\varphi^2+\sigma^2}{\epsilon\mu(1-m) + 2\varphi^2} \right\rbrace \right)\frac{\kappa}{1-m}\log\frac{2\delta^0}{\varepsilon}$ which is worse by an additive term of $\mathcal{O}\left(\frac{(M+1)\varphi^2}{\varepsilon\mu(1-m) + 2\varphi^2}\frac{\kappa}{1-m}\log\frac{2\delta^0}{\varepsilon}\right)$ than the rate granted by Theorem~6 of \citet{AjallStich}.
	\subsection{Proof of Theorem \ref{thm_noncvx_pl}.} 
	Due to \eqref{eq_supplementary} and Assumption \ref{ass_pl}, we have
	\begin{equation*}
		\begin{split}
			\mathbb{E}\left[f(x^{t+1}) |x^t\right] & \leq f(x^t) - 2\gamma\mu \left( b - \frac{LB\gamma}{2}\right)\left(f(x^t) - f^{*}\right) \\
			& + 2\gamma^2\frac{LA}{2}\left(f(x^t) - f^{*}\right) + \frac{LC\gamma^2}{2}+  c\gamma\\
			& = f(x^t) - 2\gamma\left(f(x^t) - f^{*}\right)\left[\mu\left( b - \frac{LB\gamma}{2}\right)-\frac{LA\gamma}{2}\right]+\frac{LC\gamma^2}{2}+  c\gamma.
		\end{split}
	\end{equation*}
	Subtract $f^{*}$ from both sides. Take expectation of both sides and use the tower property. Applying inequality \eqref{eq_gamma_pl}, we obtain
	\begin{equation*}
		\begin{split}
			\mathbb{E}\left[f(x^{t+1}) - f^{*}\right] & \leq \left(1 - \gamma\mu b\right)\mathbb{E}\left[f(x^t) - f^{*}\right] + \frac{LC\gamma^2}{2}+  c\gamma.
		\end{split}
	\end{equation*}
	Unrolling the recursion, we arrive at
	\begin{equation*}
		\mathbb{E}\left[f(x^T) - f^{*}\right] \leq \left(1 - \gamma\mu b\right)^T\mathbb{E}\left[f(x^0) - f^{*}\right] + \frac{LC\gamma}{2\mu b} + \frac{  c}{\mu b}.
	\end{equation*}
	\begin{flushright}
		$\blacksquare$
	\end{flushright}
	\subsection{Proof of Corollary \ref{cor_pl_1}}
	We bound every term of \eqref{eq_convergence_pl} by $\frac{  c}{\mu b}.$
	
	If $\gamma = \frac{\mu b}{L(A+\mu B)},$ and if $T\geq \frac{L(A+\mu B)}{\mu^2 b^2}\log\frac{\mu b\delta^0}{  c},$ we have
	\begin{equation*}
		\begin{split}
			\left(1-\gamma\mu b\right)^T\delta^0 &= \left(1-\frac{1}{L(A+\mu B)}\right)^T\delta^0\leq e^{- \frac{T}{L(A+\mu B)}}\delta^0\leq \frac{  c}{\mu b}.
		\end{split}
	\end{equation*}
	
	If $\gamma = \frac{1}{2\mu  b},$ and if $T\geq 2\log\frac{\mu b\delta^0}{  c},$ we have
	\begin{equation*}
		\begin{split}
			\left(1-\gamma\mu b\right)^T\delta^0\leq e^{- \frac{T}{2}}\delta^0\leq \frac{  c}{\mu b}.
		\end{split}
	\end{equation*}
	
	If $\gamma = \frac{2  c}{LC},$ and if $T\geq \frac{LC}{2  c\mu b}\log\frac{\mu b\delta^0}{  c},$ we have
	\begin{equation*}
		\begin{split}
			\left(1-\gamma\mu b\right)^T\delta^0 &= \left(1-\frac{2  c\mu b}{LC}\right)^T\delta^0\leq e^{- \frac{2  c\mu b T}{LC}}\delta^0 \leq \frac{  c}{\mu b}.
		\end{split}
	\end{equation*}
	
	Due to the choice of $\gamma,$ we have $\frac{LC\gamma}{2\mu b}\leq\frac{  c}{\mu b}.$
	
	Therefore, we obtain that $\mathbb{E}\left[f(x^T) - f^{*}\right] \leq \frac{3  c}{\mu b}.$
	\begin{flushright}
		$\blacksquare$
	\end{flushright}
	\subsection{Proof of Corollary \ref{cor_pl_2}}
	If we substitute $c$ for $0$ in \eqref{eq_convergence_pl}, then, for every $T\geq 1,$ we obtain
	\begin{equation*}
		\mathbb{E}\left[f(x^T) - f^{*}\right] \leq \left(1 - \gamma\mu b\right)^T\delta^0+ \frac{LC\gamma}{2\mu b}.
	\end{equation*}
	We bound every term in the right-hand side of the latter inequality by $\frac{\varepsilon}{2}.$
	
	If $\gamma = \frac{\mu b}{L(A+\mu B)},$ and if $T\geq \frac{L(A+\mu B)}{\mu^2 b^2}\log\frac{2\delta^0}{\varepsilon},$ then we have
	\begin{equation*}
		\begin{split}
			\left(1-\gamma\mu b\right)^T\delta^0 &= \left(1 - \frac{\mu^2 b^2}{L(A+\mu B)}\right)^T\delta^0\leq e^{-\frac{\mu^2 b^2T}{L(A+\mu B)}}\delta^0\leq \frac{\varepsilon}{2}.
		\end{split}
	\end{equation*}
	If $\gamma = \frac{1}{2\mu b},$ and if $T\geq 2\log\frac{2\delta^0}{\varepsilon},$
	\begin{equation*}
		\begin{split}
			\left(1-\gamma\mu b\right)^T\delta^0 &\leq e^{-\frac{T}{2}}\delta^0\leq\frac{\varepsilon}{2}.
		\end{split}
	\end{equation*}
	If $\gamma = \frac{\varepsilon\mu b}{LC},$ and if $T\geq \frac{LC}{\varepsilon\mu^2 b^2}\log\frac{2\delta^0}{\varepsilon},$ then we have
	\begin{equation*}
		\begin{split}
			\left(1-\gamma\mu b\right)^T\delta^0 &= \left(1-\frac{\mu^2 b^2\varepsilon}{LC}\right)\delta^0\leq e^{-\frac{\mu^2 b^2\varepsilon T}{LC}}\delta^0 \leq \frac{\varepsilon}{2}.
		\end{split}
	\end{equation*}
	Due to the choice of $\gamma,$ we have $\frac{LC\gamma}{2\mu b}\leq\frac{\varepsilon}{2}.$
	
	Then, if
	\begin{equation*}
		T\geq\max\left\lbrace 2, \frac{L(A+\mu B)}{\mu^2 b^2}, \frac{LC}{\varepsilon\mu^2 b^2} \right\rbrace \log \frac{2\delta^0}{\varepsilon},
	\end{equation*}
	we obtain $\mathbb{E}\left[f(x^T) - f^{*}\right]\leq\varepsilon.$ 
	\begin{flushright}
		$\blacksquare$
	\end{flushright}
	\subsection{Proof of Corollary \ref{cor_pl_3}}
	From \eqref{eq_convergence_pl}, when $A=0,$ $ b\leq 1,$ $0<\gamma<\min\left\lbrace \frac{ b}{LB},\frac{1}{\mu  b}\right\rbrace,$ for every $T\geq 1,$ we have
	\begin{equation*}
		\Exp{f(x^T) - f^{*}}\leq \left(1-\gamma\mu b\right)^T\delta^0+\frac{LC\gamma}{2\mu b} + \frac{  c}{\mu b}.
	\end{equation*}
	Observe that $\frac{ b}{LB}\leq\frac{1}{\mu b}.$ Let $\gamma=\min\left\lbrace \frac{ b}{LB}, \frac{\varepsilon\mu b + 2  c}{LC} \right\rbrace.$
	
	If minimum is attained when $\gamma = \frac{ b}{BL},$ then we have that $\frac{C}{\mu B} - \frac{2  c}{\mu b}\leq\varepsilon.$ If $T\geq\frac{B}{ b}\frac{\kappa}{ b}\log\frac{2\delta^0}{\varepsilon},$ then
	\begin{equation*}
		\begin{split}
			\left(1-\gamma\mu b\right)^T\delta^0 + \frac{LC\gamma}{2\mu b} + \frac{  c}{\mu b}& \leq e^{-\frac{T\mu b^2}{BL}}\delta^0 + \frac{C}{2\mu B} + \frac{  c}{\mu b}\leq \varepsilon + \frac{2  c}{\mu b}.
		\end{split}
	\end{equation*}
	If $\gamma=\frac{\varepsilon\mu b + 2  c}{LC}$ and $T\geq \frac{C}{\varepsilon\mu b + 2  c}\frac{\kappa}{ b}\log\frac{2\delta^0}{\varepsilon},$ then
	\begin{equation*}
		\begin{split}
			\left(1-\gamma\mu b\right)^T\delta^0 + \frac{LC\gamma}{2\mu b} + \frac{  c}{\mu b}& \leq
			e^{-\frac{T\mu b\left(\varepsilon\mu b + 2  c\right)}{LC}}\delta^0 + \frac{\varepsilon}{2} + \frac{  c}{\mu b} + \frac{  c}{\mu b} = \varepsilon + \frac{2  c}{\mu b}.
		\end{split}
	\end{equation*}
	Then, if $T\geq \max\left\lbrace \frac{B}{ b}, \frac{C}{\varepsilon\mu b + 2  c} \right\rbrace \frac{\kappa}{\varepsilon}\log\frac{2\delta^0}{\varepsilon},$ then $\Exp{f(x^T) - f^{*}} = \mathcal{O}\left(\varepsilon + \frac{2  c}{\mu b}\right).$
	\begin{flushright}
		$\blacksquare$
	\end{flushright}
	\section{Strongly convex case}\label{section_strongcvx_proofs}
	In Section~\ref{section_strongly_cvx_main} we have stated that Theorem~\ref{thm_noncvx_pl} on convergence of \algname{BiasedSGD} under \hyperlink{Biased ABC}{Biased ABC} assumption can be applied in strongly convex settings. We compared the rate obtained to the known convergence results in strongly convex scenario. Below we present recent results, derive several corollaries from Theorem~\ref{thm_noncvx_pl} and make a formal comparison of our results to the known results.
	 
	\subsection{Known results for convergence in function values}
	The next theorem is Theorem 4.6 from \citep{BotCurNoce}.
	\begin{theorem}\label{thm_bottou}
		Let Assumptions \ref{ass_smooth}, \ref{ass_first_and_second_mmt_limits} and \ref{ass_mu_conv} hold. Then, as long as $0<\gamma\leq \frac{q}{L(U + u^2)},$ for all $T\geq 1,$ we have
		\begin{equation*}
			\Exp{f(x^T) - f(x^{*})}\leq \left(1-\gamma \mu q\right)^{T}\left(\delta^0 - \frac{\gamma L Q}{2\mu q}\right) + \frac{\gamma L Q}{2\mu q}.
		\end{equation*}
	\end{theorem}
	Let us derive the convergence rate in Theorem \ref{thm_bottou} to compare it to our result obtained in the next section.
	\begin{corollary}\label{cor_from_bottou}
		Choose stepsize $\gamma>0$ as $\gamma=\min\left\lbrace \frac{q}{L(U+u^2)}, \frac{\varepsilon\mu q}{LQ}, \frac{1}{2\mu q}\right\rbrace.$ Then, if 
		$$
		T\geq \max\left\lbrace 2, \frac{L\left(U + u^2\right)}{q^2\mu}, \frac{LQ}{\varepsilon\mu^2q^2} \right\rbrace \log\frac{2\delta^0}{\varepsilon},
		$$
		we have $$\Exp{f(x^T) - f(x^{*})}\leq\varepsilon.$$
	\end{corollary}
	
	Next three theorems are analogues of Theorems 12 -- 14 from \citep{BezHorRichSaf} respectively.
	\begin{theorem}\label{thm_first_set}
		Let Assumptions~\ref{ass_smooth}~and~\ref{ass_mu_conv} hold. Let $g\in \mathbb{B}^1\left(\alpha,\beta\right)$ (that is, let Assumption~\ref{ass_first_set} be satisfied). Then as long as $0\leq\gamma\leq\frac{2}{\beta L},$ for all $t\in\mathbb{N},$ we have
		\begin{equation*}
			\mathbb{E}\left[f(x^t) - f(x^{*})\right] \leq \left(1-\frac{\alpha}{\beta}\gamma\mu(2-\gamma\beta L)\right)^t\left(f(x^0) - f(x^{*})\right).
		\end{equation*}
		If we choose $\gamma=\frac{1}{\beta L},$ then
		\begin{equation*}
			\mathbb{E}\left[f(x^t) - f(x^{*})\right] \leq \left(1-\frac{\alpha}{\beta^2}\frac{\mu}{L}\right)^t\left(f(x^0) - f(x^{*})\right).
		\end{equation*}
	\end{theorem}
	
	\begin{theorem}\label{thm_second_set}
		Let Assumptions~\ref{ass_smooth}~and~\ref{ass_mu_conv} hold. Let $g\in \mathbb{B}^2\left(\tau,\beta\right)$ (that is, let Assumption~\ref{ass_second_set} be satisfied). Then as long as $0\leq\gamma\leq\frac{2}{\beta L},$  for all $t\in\mathbb{N},$ we have
		\begin{equation*}
			\mathbb{E}\left[f(x^t) - f(x^{*})\right] \leq \left(1-\tau\gamma\mu(2-\gamma\beta L)\right)^t\left(f(x^0) - f(x^{*})\right).
		\end{equation*}
		If we choose $\gamma=\frac{1}{\beta L},$ then
		\begin{equation*}
			\mathbb{E}\left[f(x^t) - f(x^{*})\right] \leq \left(1-\frac{\tau}{\beta}\frac{\mu}{L}\right)^t\left(f(x^0) - f(x^{*})\right).
		\end{equation*}
	\end{theorem}
	
	\begin{theorem}\label{thm_third_set}
		Let Assumptions~\ref{ass_smooth}~and~\ref{ass_mu_conv} hold. Let $g\in \mathbb{B}^3\left(\delta\right)$ (that is, let Assumption~\ref{ass_third_set} be satisfied). Then as long as $0\leq\gamma\leq\frac{1}{L},$ for all $t\in\mathbb{N},$ we have
		\begin{equation*}
			\mathbb{E}\left[f(x^t) - f(x^{*})\right] \leq \left(1-\frac{\gamma\mu}{\delta}\right)^t\left(f(x^0) - f(x^{*})\right).
		\end{equation*}
		If we choose $\gamma=\frac{1}{L},$ then
		\begin{equation*}
			\mathbb{E}\left[f(x^t) - f(x^{*})\right] \leq \left(1-\frac{\mu}{\delta L}\right)^t\left(f(x^0) - f(x^{*})\right).
		\end{equation*}
	\end{theorem}
	
	The authors of \citep{BezHorRichSaf} make the following observation. For every gradient estimator $g\in\mathbb{B}^1\left(\alpha,\beta\right),$ there exists a unique gradient estimator $\frac{1}{\beta}g\in\mathbb{B}^3\left(\frac{\beta^2}{\alpha}\right).$ By Theorem \ref{thm_third_set}, we get the bound of $\mathcal{O}\left(\frac{\beta^2}{\alpha}\frac{L}{\mu}\log\frac{1}{\varepsilon}\right)$ on $\mathcal{T}$ which coincides with the result of Theorem \ref{thm_first_set} applied to $g.$ If $g\in\mathbb{B}^3\left(\delta\right),$ then $g\in\mathbb{B}^1\left(\frac{1}{4\delta^2}, 2\right).$ Applying Theorem \ref{thm_first_set}, we get that $\mathcal{O}\left(16\delta^2\frac{L}{\mu}\log\frac{1}{\varepsilon}\right)$ which is worse than the result of Theorem \ref{thm_third_set} by a factor of $16\delta.$ For every $g\in\mathbb{B}^2\left(\tau,\beta\right),$ there exists a unique $g\in\mathbb{B}^1\left(\tau^2,\beta\right).$ Applying Theorem \ref{thm_second_set} we obtain $\mathcal{O}\left(\frac{\beta}{\tau}\frac{L}{\mu}\log\frac{1}{\varepsilon}\right),$ whence applying Theorem \ref{thm_first_set} we obtain $\mathcal{O}\left(\frac{\beta^2}{\tau^2}\frac{L}{\mu}\log\frac{1}{\varepsilon}\right).$ Since $\beta\geq\tau,$ the second result is worse by a factor of $\frac{\beta}{\tau}.$
	\subsection{Convergence in function values: our results}
	Observe that Assumption \ref{ass_pl} is more general than Assumption \ref{ass_mu_conv}. Therefore, Theorem \ref{thm_noncvx_pl} can be applied to functions that satisfy Assumption \ref{ass_mu_conv}.
	\begin{theorem}\label{thm_strong_cvx}
		Let Assumptions \ref{ass_smooth}, \ref{ass_scalar_ABC} and \ref{ass_mu_conv} hold. Choose a stepsize such that
		\begin{equation*}
			0<\gamma<\min\left\lbrace \frac{\mu b}{L(A+\mu B)}, \frac{1}{\mu  b}\right\rbrace.
		\end{equation*}
		Then, for every $T\geq 1,$ we have
		\begin{equation}\label{eq_mu_convex_functional_convergence}
			\mathbb{E}\left[f(x^T) - f(x^{*})\right] \leq \left(1 - \gamma\mu b\right)^T\delta^0 + \frac{LC\gamma}{2\mu b} + \frac{  c}{\mu b},
		\end{equation}
		where $\delta^0=f(x^0) - f(x^{*}).$
	\end{theorem}
	
	Clearly, all of the corollaries from Theorem~\ref{thm_noncvx_pl} hold in the strongly convex setup as well. Therefore, we do not write them here again.
	
	Observe that if $A =   c = 0,$ we recover the result of Theorem~\ref{thm_bottou} (see Theorem~4.6 from \citep{BotCurNoce}).
	\begin{corollary}\label{cor_strongly_cvx_recover_bottou}
		Suppose $A=  c=0.$ Choose stepsize $\gamma>0$ as $\gamma = \min\left\lbrace\frac{ b}{LB}, \frac{\varepsilon b\mu}{LC}, \frac{1}{2\mu b} \right\rbrace.$ Then, if 
		\begin{equation*}
			T \geq \max\left\lbrace 2, \frac{LB}{ b^2\mu}, \frac{LC}{\varepsilon b^2\mu^2}\right\rbrace\log\frac{2\delta^0}{\varepsilon},
		\end{equation*}
		we have $$\Exp{f(x^T) - f(x^{*})}\leq \varepsilon.$$
	\end{corollary}
	To recover the result from Corollary~\ref{cor_from_bottou}, one needs to substitute $B$ for $U+u^2,$ $C$ for $Q,$ $ b$ for $q$ in accordance with the representation of Assumption~\ref{ass_first_and_second_mmt_limits} in \hyperlink{Biased ABC}{Biased ABC} framework (see Theorem~\ref{thm_assumptions_in_our_framework} and Table~\ref{tab_assns_in_our_frame}).
	
	Observe that if $A = C =   c = 0,$ we retrieve the results similar to Theorems~\ref{thm_first_set}~--~\ref{thm_third_set}.
	
	\begin{corollary}\label{cor_strongly_cvx_recover_beznosikov}
		Suppose $A = C =   c = 0.$ Choose stepsize $\gamma>0$ as $\gamma=\frac{ b}{LB}.$ Then, for every $T\geq 1,$ we have
		\begin{equation*}
			\Exp{f(x^T) - f(x^{*})} \leq \left(1 - \frac{ b^2\mu}{BL}\right)^T\delta^0.
		\end{equation*}
		If $T\geq \frac{BL}{ b^2\mu}\log\frac{\delta^0}{\varepsilon},$ then we have $$\Exp{f(x^T) - f(x^{*})} \leq \varepsilon.$$
	\end{corollary}
	If we substitute $B$ for $\beta^2,$ $ b$ for $\frac{\alpha}{\beta}$ (see Theorem~\ref{thm_assumptions_in_our_framework} and Table~\ref{tab_assns_in_our_frame}), Corollary~\ref{cor_strongly_cvx_recover_beznosikov} yields the rate of $\mathcal{O}\left(\frac{\beta^4}{\alpha^2}\frac{L}{\mu}\log\frac{\delta^0}{\varepsilon}\right),$ which is worse by a factor of $\frac{\beta^2}{\alpha}$ than the rate granted by Theorem~\ref{thm_first_set} \citep[Theorem~12]{BezHorRichSaf}.
	
	If we substitute $B$ for $\beta^2,$ $ b$ for $\tau$ (see Theorem~\ref{thm_assumptions_in_our_framework} and Table~\ref{tab_assns_in_our_frame}), Corollary~\ref{cor_strongly_cvx_recover_beznosikov} yields the rate of $\mathcal{O}\left(\frac{\beta^2}{\tau^2}\frac{L}{\mu}\log\frac{\delta^0}{\varepsilon}\right),$ which is worse by a factor of $\frac{\beta}{\tau}$ than the rate granted by Theorem~\ref{thm_second_set} \citep[Theorem~13]{BezHorRichSaf}.
	
	If we substitute $B$ for $2\left(2-\frac{1}{\delta}\right),$ $ b$ for $\frac{1}{2\delta}$ (see Theorem~\ref{thm_assumptions_in_our_framework} and Table~\ref{tab_assns_in_our_frame}), Corollary~\ref{cor_strongly_cvx_recover_beznosikov} yields the rate of $\mathcal{O}\left(\delta^2\frac{L}{\mu}\log\frac{\delta^0}{\varepsilon}\right),$ which is worse by a factor of $\delta$ than the rate granted by Theorem~\ref{thm_third_set}  \citep[Theorem~14]{BezHorRichSaf}.
	\subsection{Proof of Corollary \ref{cor_from_bottou}}
	If $\gamma = \frac{q}{L(U+u^2)}$ and $T\geq \frac{L(U+u^2)}{q^2\mu}\log\frac{2\delta^0}{\varepsilon},$ then
	\begin{equation*}
		\begin{split}
			\left(1-\gamma\mu q\right)^T\left(\delta^0 - \frac{\gamma L Q}{2\mu q} \right)& \leq \left(1 - \frac{q^2\mu}{L(U+u^2)}\right)^T\delta^0 \leq e^{-\frac{q^2\mu T}{L(U+u^2)}}\delta^0\leq \frac{\varepsilon}{2}.
		\end{split}
	\end{equation*}
	If $\gamma = \frac{\varepsilon\mu q}{LQ}$ and $T\geq \frac{LQ}{\varepsilon\mu^2q^2}\log\frac{2\delta^0}{\varepsilon},$ then
	\begin{equation*}
		\begin{split}
			\left(1-\gamma\mu q\right)^T\left(\delta^0 - \frac{\gamma L Q}{2\mu q} \right)& \leq \left(1 - \frac{\varepsilon\mu^2q^2}{LQ}\right)^T\delta^0\leq e^{-\frac{\mu^2q^2 T}{LQ}}\delta^0\leq \frac{\varepsilon}{2}.
		\end{split}
	\end{equation*}
	If $\gamma=\frac{1}{2\mu q}$ and $T\geq 2\log\frac{2\delta^0}{\varepsilon},$ then
	\begin{equation*}
		\begin{split}
			\left(1-\gamma\mu q\right)^T\left(\delta^0 - \frac{\gamma L Q}{2\mu q} \right) \leq e^{-\frac{T}{2}}\delta^0 \leq \frac{\varepsilon}{2}.
		\end{split}
	\end{equation*}
	Due to the choice of $\gamma,$ we have $\frac{\gamma LQ}{2\mu q} \leq \frac{\varepsilon}{2}.$
	
	Then, if
	\begin{equation*}
		T \geq \max\left\lbrace 2, \frac{L(U+u^2)}{q^2\mu}, \frac{LQ}{\varepsilon\mu^2q^2}\right\rbrace \log\frac{2\delta^0}{\varepsilon},
	\end{equation*}
	we obtain $\Exp{f(x^T) - f(x^{*})}\leq \varepsilon.$
	\begin{flushright}
		$\blacksquare$
	\end{flushright}
	\subsection{Proof of Theorem \ref{thm_strong_cvx}}
	Follow exactly the same steps as in the proof of Theorem \ref{thm_noncvx_pl}.
	\begin{flushright}
		$\blacksquare$
	\end{flushright}	
	\subsection{Proof of Corollary \ref{cor_strongly_cvx_recover_bottou}}
	If $\gamma = \frac{ b}{LB}$ and $T\geq \frac{LB}{ b^2\mu}\log\frac{2\delta^0}{\varepsilon},$ then
	\begin{equation*}
		\begin{split}
			\left(1-\gamma\mu b\right)^T\delta^0 & = \left(1-\frac{ b^2\mu}{LB}\right)^T\delta^0 \leq e^{-\frac{T b^2\mu}{LB}}\delta^0 \leq \frac{\varepsilon}{2}.
		\end{split}
	\end{equation*}
	If $\gamma = \frac{\varepsilon b\mu}{LC}$ and $T\geq \frac{LC}{\varepsilon b^2\mu^2}\log\frac{2\delta^0}{\varepsilon},$ then
	\begin{equation*}
		\begin{split}
			\left(1-\gamma\mu b\right)^T\delta^0 & = \left(1-\frac{\varepsilon b^2\mu^2}{LC}\right)^T\delta^0\leq e^{-\frac{T\varepsilon b^2\mu^2}{LC}}\delta^0 \leq\frac{\varepsilon}{2}.
		\end{split}
	\end{equation*}
	If $\gamma = \frac{1}{2\mu b}$ and $T\geq 2\log\frac{2\delta^0}{\varepsilon},$ then
	\begin{equation*}
		\begin{split}
			\left(1-\gamma\mu b\right)^T\delta^0 & \leq e^{-\frac{T}{2}}\delta^0\leq \frac{\varepsilon}{2}.
		\end{split}
	\end{equation*}
	Due to the choice of $\gamma,$ we have $\frac{LC\gamma}{2\mu b}\leq\frac{\varepsilon}{2}.$
	Then, if
	\begin{equation*}
		T\geq \max\left\lbrace 2, \frac{LB}{ b^2\mu}, \frac{LC}{\varepsilon b^2\mu^2} \right\rbrace \log\frac{2\delta^0}{\varepsilon},
	\end{equation*}
	we obtain $\Exp{f(x^T) - f(x^{*})}\leq\varepsilon.$
	\begin{flushright}
		$\blacksquare$
	\end{flushright}
	\subsection{Proof of Corollary \ref{cor_strongly_cvx_recover_beznosikov}}
	Consider \eqref{eq_mu_convex_functional_convergence} and recall that $A=C=  c=0.$ Note that in this case $\frac{\mu b}{L(A+\mu B)} = \frac{ b}{LB}$ is no greater that $\frac{1}{\mu b}.$ Indeed, 
	\begin{equation*}
		b\left\|\nabla f(x)\right\|^2\leq\langle\Exp{g(x)}, \nabla f(x)\rangle\leq\left\|\Exp{g(x)}\right\|\cdot\left\|\nabla f(x)\right\|
	\end{equation*}
	(by Cauchy--Schwarz inequality), which (combined with \hyperlink{Biased ABC}{Biased ABC}) leads to
	\begin{equation*}
		b^2\left\|\nabla f(x)\right\|^2\leq\left\|\Exp{g(x)}\right\|^2\leq\Exp{\left\|g(x)\right\|^2}\leq B\left\|\nabla f(x)\right\|^2.
	\end{equation*}
	Therefore, we have that $ b^2\leq B.$ Then, $ b^2\leq \frac{L}{\mu}B \iff \frac{ b}{LB}\leq\frac{1}{\mu  b}.$
	
	Hence, we can choose $\gamma=\frac{ b}{LB},$ which yields that
	\begin{equation*}
		\Exp{f(x^T) - f(x^{*})} \leq \left(1 - \frac{ b^2\mu}{LB}\right)^T\delta^0.
	\end{equation*}
	If $T\geq \frac{LB}{ b^2\mu}\log\frac{\delta^0}{\varepsilon},$ then
	\begin{equation*}
		\begin{split}
			\left(1-\frac{ b^2\mu}{LB}\right)^T\delta^0\leq e^{-\frac{T b^2\mu}{LB}}\delta^0 \leq\varepsilon.
		\end{split}
	\end{equation*}
	\begin{flushright}
		$\blacksquare$
	\end{flushright}	
	\subsection{Iterate convergence: further discussion}
	In Section~\ref{section_strongly_cvx_main} we introduce strict Assumption~\ref{ass_ABC_consts} and formulate convergence Theorem~\ref{thm_weak_bias} subject to this condition. It is reasonable to ask whether Assumption~\ref{ass_ABC_consts} is realistic. In this part of the appendix we give a useful example of a setting that meets the requirements of the assumption imposed.
	
	It is easy to see that Assumption~\ref{ass_ABC_consts} holds only when $b$ is relatively large, and $A$ is small, which is not necessarily the case in practice. However, let us show that it can be satisfied. Consider the $k_2$-regularized logistic regression with $f_j = \log\left(1+e^{-b_j\langle e_j, x\rangle}\right) + \frac{1}{2}\left\|x\right\|^2,$ where $e_j$ is the $j$-th unit vector, $b_j\in\{0,1\},$ $j\in[n],$ $n\geq 2.$ It is straightforward to show that all $f_j$ and $f=\frac{1}{n}\sum_{j=1}^{n}f_j$ are $\frac{5}{4}$-smooth and $1$-strongly-convex. Consider the estimator from Definition~\ref{def_distributed_biased_rounding}, and let $a_k=k,$ $k\in\mathbb{N}\cup\{0\},$ $p_j=\frac{1}{5}.$ From \eqref{eq_A_distributed_rounding}--\eqref{eq_chi_distributed_rounding}, we obtain that $A_r = \frac{2}{n},$ $B_r=\frac{2}{5},$ $C_r=\frac{4\Delta^{*}}{n},$ $ b_r=\frac{4}{5},$ $  c_r=0.$ Then Assumption \ref{ass_ABC_consts} holds since $\frac{2}{n} - \frac{1}{4}<1.$
	\subsection{Proof of Theorem \ref{thm_weak_bias}}
	Let $r^t\eqdef x^t - x^{*}.$ We get
	\begin{equation*}
		\begin{split}
			\left\|r^{t+1}\right\|^2 = \left\|\left(x^t-\gamma g^t\right) - x^{*}\right\|^2  = \left\|x^t-x^{*} - \gamma g^t\right\|^2 = \left\|r^{t}\right\|^2 - 2\gamma\big{<}r^t, g^t\big{>} + \gamma^2\left\|g^t\right\|^2.
		\end{split}
	\end{equation*}
	Now we compute expectation of both sides of the inequality, conditional on $x^t:$
	\begin{equation*}
		\mathbb{E}\left[\left\|r^{t+1}\right\|^2|x^t\right] = \left\|r^{t}\right\|^2 - 2\gamma\big{<}r^t, \mathbb{E}[g^t|x^t]\big{>} + \gamma^2\mathbb{E}\left[\left\|g^t\right\|^2|x^t\right].
	\end{equation*}
	Notice that
	\begin{equation*}
		2\big{<}r^t, \mathbb{E}[g^t|x^t]\big{>} = 2\big{<}r^t, \mathbb{E}[g^t|x^t] - \nabla f(x^{t})\big{>} + 2\big{<}r^t, \nabla f(x^{t})\big{>}.
	\end{equation*}
	Due to $\mu$-convexity, we have
	\begin{equation}\label{eq_scalar_mu}
		\big{<}r^t, \nabla f(x^{t})\big{>}\geq D_f(x^t, x^{*}) + \frac{\mu}{2}\left\|r^t\right\|^2.
	\end{equation}
	Further, using Young's Inequality (Lemma \ref{lemma_young}, \eqref{eq_scalar_prod_young}), we get
	\begin{equation}\label{eq_scalar_final}
		-2\big{<}r^t, \mathbb{E}[g^t|x^t] - \nabla f(x^{t})\big{>}\leq s\left\|r^t\right\|^2 + \frac{1}{s}\left\|\mathbb{E}[g^t|x^t] - \nabla f(x^{t})\right\|^2.
	\end{equation}
	Notice that
	\begin{equation*}
		\begin{split}
			\left\|\mathbb{E}[g^t|x^t] - \nabla f(x^{t})\right\|^2 & = \left\|\mathbb{E}[g^t|x^t]\right\|^2 - 2\langle \mathbb{E}[g^t|x^t], \nabla f(x^{t})\rangle + \left\| \nabla f(x^{t})\right\|^2\\
			& \leq 2AD_f(x^t, x^{*}) + B\left\|\nabla f(x^t)\right\|^2 + C \\
			& - 2\left( b\left\|\nabla f(x^t)\right\|^2 -   c\right) + \left\|\nabla f(x^{t})\right\|^2.\\
		\end{split}
	\end{equation*}
	Below we use this fact from Lemma~\ref{lemma_smooth_bregman}:
	\begin{equation}\label{eq_fxk}
		\left\|\nabla f(x^t)\right\|^2 \leq2LD_f(x^t, x^{*}).
	\end{equation}
	This leads to
	\begin{equation*}
		\begin{split}
			\mathbb{E}\left[\left\|r^{t+1}\right\|^2|x^t\right]& \stackrel{\eqref{eq_scalar_mu}, \eqref{eq_scalar_final}}{\leq} \left(1-\gamma\left(\mu - s\right)\right)\left\|r^t\right\|^2 - 2\gamma D_f(x^t, x^{*})\\
			& + \gamma^2\mathbb{E}\left[\left\|g^t\right\|^2|x^t\right] + \frac{\gamma}{s}\left(\left\|\mathbb{E}[g^t|x^t] - \nabla f(x^t)\right\|^2\right)\\
			& \stackrel{\eqref{eq_fxk}}{\leq}\left(1-\gamma\left(\mu -s\right)\right)\left\|r^t\right\|^2 - 2\gamma D_f(x^t, x^{*})\\
			& + \gamma^2\left(2AD_f(x^t, x^{*}) + B\left\|\nabla f(x^t)\right\|^2 + C\right)\\
			& +\frac{\gamma}{s}\left[ 2AD_f(x^t, x^{*}) + B\left\|\nabla f(x^t)\right\|^2 + C\right.\\
			& \left.- 2\left( b\left\|\nabla f(x^t)\right\|^2 -   c\right) + \left\|\nabla f(x^t)\right\|^2\right]\\
			& = \left(1-\gamma\left(\mu - s\right)\right)\left\|r^{t}\right\|^2 \\
			& -2\gamma D_f(x^t, x^{*})\left[1 - A\gamma - \frac{A}{s}-L\left(\gamma B +\frac{B}{s} - \frac{2 b}{s} + \frac{1}{s}\right)\right]+\\
			& + \gamma^2C + \frac{\gamma \left(C + 2  c\right)}{s}.
		\end{split}
	\end{equation*}
	Due to \eqref{eq_gamma_weak_bias}, we have
	\begin{equation*}
		\begin{split}
			\mathbb{E}\left[\left\|r^{t+1}\right\|^2|x^t\right]&\leq \left(1-\gamma\left(\mu - s\right)\right)\left\|r^t\right\|^2 +\gamma^2C + \frac{\gamma \left(C+2  c\right)}{s}.\\
		\end{split}
	\end{equation*}
	Take expectation again on both sides and use the tower property
	\begin{equation*}
		\mathbb{E}\left[\left\|r^{t+1}\right\|^2\right]=\mathbb{E}\left[\mathbb{E}\left[\left\|r^{t+1}\right\|^2|x^t\right]\right].
	\end{equation*}
	We arrive at
	\begin{equation*}
		\begin{split}
			\mathbb{E}\left[\left\|r^{t+1}\right\|^2\right] & \leq \left(1-\gamma\left(\mu - s\right)\right)\mathbb{E}\left[\left\|r^t\right\|^2\right]+ \gamma^2C + \frac{\gamma\left(C+2  c\right)}{s}.\\
		\end{split}
	\end{equation*}
	Unrolling the recurrence and noting that $\mathbb{E}\left[\left\|r^{0}\right\|^2\right]=\left\|r^0\right\|^2$ gives us
	\begin{equation*}
		\begin{split}
			\mathbb{E}\left[\left\|r^t\right\|^2\right]&\leq\left(1-\gamma\left(\mu - s\right)\right)^t\left\|r^{0}\right\|^2\\
			& + \gamma\left(\gamma C + \frac{C+2  c}{s}\right)\sum_{i=0}^{t-1}\left(1-\gamma\left(\mu - s\right)\right)^i\\
			& \leq\left(1-\gamma\left(\mu - s\right)\right)^t\left\|r^{0}\right\|^2+ \frac{\gamma C +\frac{C+2  c}{s}}{\mu-s}.
		\end{split}
	\end{equation*}
	\begin{flushright}
		$\blacksquare$
	\end{flushright}

	\section{Assumptions \ref{ass_first_set}--\ref{ass_first_and_second_mmt_limits} in biased ABC framework}
	In Table~\ref{tab_assns_in_our_frame} we have presented the values of control variables $A,B,C,b$ and $c$ in our \hyperlink{Biased ABC}{Biased ABC} framework for a gradient estimator that satisfies any of assumptions listed in Section~\ref{sect_existing_models}. Here we give a formal proof of these results.
	\begin{theorem}\label{thm_assumptions_in_our_framework}
		The following relations hold.
		\begin{enumerate}[label=\roman*, wide, labelwidth=!, labelindent=5pt]
			\item\label{item_framework_first} Suppose $g(x)$ satisfies Assumption \ref{ass_first_set}. Then it satisfies Assumption \ref{ass_scalar_ABC} with $A=0,$ $B=\beta^2,$ $C = 0,$ $ b = \frac{\alpha}{\beta},$ $  c=0.$
			\item\label{item_framework_second} Suppose $g(x)$ satisfies Assumption \ref{ass_second_set}. Then it satisfies Assumption \ref{ass_scalar_ABC} with $A=0,$ $B=\beta^2,$ $C = 0,$ $ b = \tau,$ $  c=0.$
			\item\label{item_framework_third} Suppose $g(x)$ satisfies Assumption \ref{ass_third_set}. Then it satisfies Assumption \ref{ass_scalar_ABC} with $A = C =   c = 0,$ $B = 2\left(2-\frac{1}{\delta}\right),$ $ b = \frac{1}{2\delta}.$
			\item\label{item_framework_bv} Suppose $g(x)$ satisfies Assumption \ref{ass_BV}. Then it satisfies Assumption \ref{ass_scalar_ABC} with $A = C =   c =0,$ $B = 2(1 + \xi + \eta),$ $ b =\frac{1-\eta}{2} .$
			\item\label{item_framework_breq} Suppose $g(x)$ satisfies Assumption \ref{ass_breq}. Then it satisfies Assumption \ref{ass_scalar_ABC} with $A=0,$ $B = \zeta,$ $C=0,$ $ b = \rho,$ $  c = 0.$
			\item\label{item_framework_stich_decomposition} Suppose $g(x)$ satisfies Assumption \ref{ass_stich_decomposition}. Then it satisfies Assumption \ref{ass_scalar_ABC} with $A=0,$ $B~=~2(M+~1)~(m+1),$ $C = 2(M+1)\varphi^2 +  \sigma^2,$ $ b=\frac{1-m}{2},$ $  c=\frac{\varphi^2}{2}.$
			\item\label{item_framework_abs_compr} Suppose $g(x)$ satisfies Assumption \ref{ass_abs_compr}. Then it satisfies Assumption \ref{ass_scalar_ABC} with $A=0,$ $B = 2,$ $C = 2\Delta^2,$ $ b = \frac{1}{2},$ $  c= \frac{\Delta^2}{2}.$
			\item\label{item_framework_fsml} Suppose $g(x)$ satisfies Assumption \ref{ass_first_and_second_mmt_limits}. Then it satisfies Assumption \ref{ass_scalar_ABC} with $A=0,$ $B=U+u^2,$ $C=Q,$ $ b=q,$ $  c=0.$
		\end{enumerate}
	\end{theorem}
	\noindent\textbf{Proof of Theorem \ref{thm_assumptions_in_our_framework}.}
	Let us prove all of the assertions stated in Theorem \ref{thm_assumptions_in_our_framework} one by one.\\
	\ref{item_framework_first} From \eqref{eq_first_set}, we deirve that $\langle\nabla f(x, \Exp{g(x)})\rangle\geq \frac{\alpha}{\beta}\left\|\nabla f(x)\right\|^2.$ Therefore, we can choose $ b=\frac{\alpha}{\beta},$ $  c=0.$ From \eqref{eq_implication}, we obtain that $A=0,$ $B=\beta^2,$ $C = 0.$\\
	
	\ref{item_framework_second} From \eqref{eq_second_set}, we derive that $\langle\nabla f(x, \Exp{g(x)})\rangle\geq \tau\left\|\nabla f(x)\right\|^2.$ Therefore, we can choose $ b=\tau,$ $  c=0.$ From \eqref{eq_implication}, we obtain that $A=0,$ $B=\beta^2,$ $C = 0.$\\
	
	\ref{item_framework_third} From \eqref{eq_contractive_biased}, we derive that
	\begin{equation*}
		\begin{split}
			\langle\Exp{g(x)}, \nabla f(x)\rangle & \geq \frac{1}{2}\left(\Exp{\left\|g(x)\right\|^2} + \frac{1}{\delta}\left\|\nabla f(x)\right\|^2\right) \geq \frac{1}{2\delta}\left\|\nabla f(x)\right\|^2.
		\end{split}
	\end{equation*}
	Further,
	\begin{equation*}
		\begin{split}
			\Exp{\left\|g(x)\right\|^2} & = \Exp{\left\|g(x) - \nabla f(x) + \nabla f(x)\right\|^2}\\
			& \leq 2\Exp{\left\|g(x) - \nabla f(x)\right\|^2} + 2\left\|\nabla f(x)\right\|^2\\
			& \leq 2\left(2-\frac{1}{\delta}\right)\left\|\nabla f(x)\right\|^2.
		\end{split}
	\end{equation*}
	
	\ref{item_framework_bv} From \eqref{eq_BV_bias}, we derive that
	\begin{equation*}
		\begin{split}
			\langle \Exp{g(x)}, \nabla f(x) \rangle &\geq \frac{1}{2}\left(\left\|\Exp{g(x)}\right\|^2 + (1-\eta)\left\|\nabla f(x)\right\|^2\right)\geq \frac{1-\eta}{2}\left\|\nabla f(x)\right\|^2.
		\end{split}
	\end{equation*}
	Further, from \eqref{eq_bv_decomposition}, \eqref{eq_BV_bias} and \eqref{eq_BV_variance}, we obtain that
	\begin{equation*}
		\begin{split}
			\Exp{\left\|g(x)\right\|^2} & = \Exp{\left\|g(x) - \nabla f(x) + \nabla f(x)\right\|^2}\\
			& \leq 2\Exp{\left\|g(x) - \nabla f(x)\right\|^2} + 2\left\|\nabla f(x)\right\|^2\\
			& \leq 2\left(1 + \xi + \eta\right)\left\|\nabla f(x)\right\|^2.
		\end{split}
	\end{equation*}
	
	\ref{item_framework_breq} From \eqref{eq_breq_scalar}, we conclude that $ b = \rho,$ $  c = 0.$ From \eqref{eq_breq_second_mmt}, we derive that $A=0,$ $B = \zeta,$ $C=0.$
	
	\ref{item_framework_stich_decomposition} It follows from the proof of Theorem \ref{thm_informal_abc}--\ref{item_abc_from_bnd}.
	
	\ref{item_framework_abs_compr} From \eqref{eq_abs_compr}, we have
	\begin{equation*}
		\begin{split}
			\langle\Exp{g(x)}, \nabla f(x)\rangle & \geq \frac{1}{2}\left(\Exp{\left\|g(x)\right\|^2} + \left\|\nabla f(x)\right\|^2\right) - \frac{\Delta^2}{2}\geq \frac{1}{2}\left\|\nabla f(x)\right\|^2 - \frac{\Delta^2}{2},
		\end{split}
	\end{equation*}
	\begin{equation*}
		\begin{split}
			\Exp{\left\|g(x)\right\|^2} & = \Exp{\left\|g(x) - \nabla f(x) + \nabla f(x)\right\|^2}\\
			& \leq 2\Exp{\left\|g(x) - \nabla f(x)\right\|^2} + 2\left\|\nabla f(x)\right\|^2\\
			& \leq 2\left\|\nabla f(x)\right\|^2 + 2\Delta^2.
		\end{split}
	\end{equation*}
	\ref{item_framework_fsml} It follows from the proof of Theorem \ref{thm_informal_abc}--\ref{item_abc_from_fsml}.
	\begin{flushright}
		$\blacksquare$
	\end{flushright}
	\section{New estimators in biased ABC framework: proofs for Section \ref{section_sources_of_bias_appendix}}\label{section_sources_proofs}
	In this section we prove the results announced in Section~\ref{section_sources_of_bias_appendix}.
	\subsection{Proof of Claim \ref{claim_sampling_no_division}}
	First, let us find constants for \eqref{eq_scalar_prod}:
	\begin{equation*}
		\begin{split}
			\langle \nabla f(x), \mathbb{E}\left[g(x)\right] \rangle & = \bigg{<} \frac{1}{n}\sum_{i=1}^{n}\nabla f_i(x), \mathbb{E}\left[\frac{1}{|S|}\sum_{i=1}^nv_i\nabla f_i(x)\right]\bigg{>}\\
			& \geq \bigg{<} \frac{1}{n}\sum_{i=1}^{n}\nabla f_i(x), \frac{1}{n}\sum_{i=1}^n \min\{p_i\}\nabla f_i(x)\bigg{>} \\
			& \geq\min_i\left\lbrace p_i\right\rbrace \left\|\nabla f(x)\right\|^2.
		\end{split}
	\end{equation*}
	
	Second, let us find an upper bound on the variance of the gradient estimator $g(x).$ Notice that, since $\tilde{g}(x)$ is independent of $X,$ and $\mathbb{E}\left[X\right]=0,$ we can write that 
	$$
	\mathbb{E}\left[\left\|g(x)\right\|^2\right] = \mathbb{E}\left[\left\|\tilde{g}(x)\right\|^2\right] + \mathbb{E}\left[\left\|X\right\|^2\right] = \mathbb{E}\left[\left\|\tilde{g}(x)\right\|^2\right] + \sigma^2.
	$$
	
	Clearly, $\mathbb{E}[\mathbb{I}_i] = p_i.$ Note, that, for $i\neq j\in[n],$ random sets $S_i$ and $S_j$ are independent, random variables $\mathbb{I}_i$ and $\mathbb{I}_j$ are also independent. Therefore, 
	$$
	\mathbb{E}\left[\mathbb{I}_i\mathbb{I}_j\right] = \mathbb{E}[\mathbb{I}_i]\mathbb{E}[\mathbb{I}_i]=p_ip_j.
	$$
	Further, let us bound the second moment of $\tilde{g}(x)$ from above:
	\begin{equation*}
		\begin{split}
			\mathbb{E}\left[\left\|\tilde{g}(x)\right\|^2\right] & = \mathbb{E}\left[\left\|\frac{1}{|S|}\sum_{i=1}^{n}\mathbb{I}_i\nabla f_i(x)\right\|^2\right]\\
			&\leq \mathbb{E}\left[\frac{1}{|S|}\sum_{i=1}^{n}\mathbb{I}_i\left\|\nabla f_i(x)\right\|^2\right]\\
			&= \sum_{i=1}^{n}\mathbb{E}\left[\frac{\mathbb{I}_i}{|S|}\right]\left\|\nabla f_i(x)\right\|^2\\
			&\leq \sum_{i=1}^{n}\mathbb{E}\left[\frac{1}{|S|}\right]\left\|\nabla f_i(x)\right\|^2\\
			&\leq \frac{1}{n\min_i\{p_i\}}\sum_{i=1}^{n}\left\|\nabla f_i(x)\right\|^2.\\
		\end{split}
	\end{equation*}
	Due to Assumption \ref{ass_smooth_functionwise}, we obtain that
	\begin{equation*}
		\begin{split}
			\mathbb{E}\left[\left\|\tilde{g}(x)\right\|^2\right] & \leq \frac{2\max_i\{L_i\}}{n\min_i\{p_i\}}\sum_{i=1}^{n}D_{f_i}\left(x, x^{*}\right)\\
			&\leq \frac{2\max_i\{L_i\}}{\min_i\{p_i\}}D_f(x,x^{*}) + \frac{2\max_i\{L_i\}}{\min_i\{p_i\}}\Delta^{*}.\\
		\end{split}
	\end{equation*}
	Therefore, we can choose $A = \frac{\max_i\{L_i\}}{\min_{i}{p_i}},$ $B = 0,$ $C = 2A\Delta^{*}+\sigma^2,$ $b = \min_i\left\lbrace p_i\right\rbrace,$ $c=0.$ 
	\begin{flushright}
		$\blacksquare$
	\end{flushright}
	\subsection{Proof of Claim \ref{claim_distributions}}
	Let us establish \eqref{eq_scalar_prod} first:
	\begin{equation*}
		\begin{split}
			\langle \nabla f(x), \mathbb{E}\left[g(x)\right]\rangle & = \bigg{<} \nabla f(x), \frac{1}{n}\sum_{i=1}^{n}c_i\nabla f_i(x) \bigg{>} \geq \min_i\left\lbrace c_i \right\rbrace \left\|\nabla f(x)\right\|^2.\\
		\end{split}
	\end{equation*}
	Further, we establish \eqref{eq_ABC}. We use the convexity of the $k_2$-norm and Lemma \ref{lemma_smooth_bregman}.
	\begin{equation*}
		\begin{split}
			\mathbb{E}\left[\left\|g(x)\right\|^2\right] & \leq \frac{1}{n}\sum_{i=1}^{n}\mathbb{E}\left[\left\|v_i\nabla f_i(x)\right\|^2\right]\\
			& = \frac{1}{n}\sum_{i=1}^{n}\mathbb{E}\left[v_i^2\right]\left\|\nabla f_i(x)\right\|^2\\
			& \leq \frac{2\max_i\left\lbrace L_i\mathbb{E}\left[v_i^2\right] \right\rbrace}{n}\sum_{i=1}^{n}D_{f_i}(x, x^{*})\\
			& = 2\max_i\left\lbrace L_i\mathbb{E}\left[v_i^2\right] \right\rbrace D_f\left( x, x^{*}\right)+ 2\max_i\left\lbrace L_i\mathbb{E}\left[v_i^2\right] \right\rbrace\Delta^{*}.\\
		\end{split}
	\end{equation*}
	\begin{flushright}
		$\blacksquare$
	\end{flushright}
	\subsection{Proof of Claim \ref{claim_distributed_biased_rounding}}
	First, we establish that \eqref{eq_scalar_prod} holds:
	\begin{equation*}
		\begin{split}
			\langle \nabla f(x), \mathbb{E}\left[g(x)\right] \rangle &= \bigg{<} \nabla f(x), \frac{1}{n}\sum_{j=1}^{n}p_j\tilde{g}_j(x)\bigg{>} + \bigg{<} \nabla f(x), \frac{1}{n}\sum_{j=1}^{n}(1-p_j)\nabla f_j(x)\bigg{>}\\
			& \geq \max_j\{p_j\}\langle \nabla f(x), \tilde{g}(x) \rangle + \max_j\{1 - p_j\}\left\|\nabla f(x)\right\|^2\\
			& \geq \left(\max_j\{p_j\} \cdot \inf_{k\in\mathbb{Z}}\frac{2a_k}{a_k+a_{k+1}}  + \max_j\{1 - p_j\}\right)\left\|\nabla f(x)\right\|^2.
		\end{split}
	\end{equation*}
	Further, we need to show that \eqref{eq_ABC} is also valid.
	\begin{equation}\label{eq_distributed_rounding_snd_mmt}
		\begin{split}
			\mathbb{E}\left[\left\|g(x)\right\|^2\right] &= \mathbb{E}\left[\left\|\frac{1}{n}\sum_{j=1}^{n}\mathbb{I}_j\tilde{g}_j(x) + \frac{1}{n}\sum_{j=1}^{n}\left(1-\mathbb{I}_j\right)\nabla f_j(x)\right\|^2\right]\\
			& \leq 2\mathbb{E}\left[\left\|\frac{1}{n}\sum_{j=1}^{n}\mathbb{I}_j\tilde{g}_j(x)\right\|^2\right] + 2\mathbb{E}\left[\left\|\frac{1}{n}\sum_{j=1}^{n}\left(1-\mathbb{I}_j\right)\nabla f_j(x)\right\|^2\right]\\
			& = \frac{2}{n^2}\mathbb{E}\left[\left\|\sum_{j=1}^{n}\mathbb{I}_j\tilde{g}_j(x)\right\|^2\right] + \frac{2}{n^2}\mathbb{E}\left[\left\|\sum_{j=1}^n\left(1-\mathbb{I}_j\right)\nabla f_j(x)\right\|^2\right].\\
		\end{split}
	\end{equation}
	Let us deal with each term separately. For the first one we have
	\begin{equation*}
		\begin{split}
			\mathbb{E}\left[\left\|\sum_{j=1}^{n}\mathbb{I}_j\tilde{g}_j(x)\right\|^2\right] & = \sum_{j=1}^{n}\mathbb{E}\left[\mathbb{I}_j^2\right]\left\|\tilde{g}_j\right\|^2 + 2 \sum_{j\neq h}\mathbb{E}\left[\mathbb{I}_j\right]\mathbb{E}\left[\mathbb{I}_h\right]\langle \tilde{g}_j, \tilde{g}_h\rangle\\
			& = \sum_{j=1}^{n}p_j\left\|\tilde{g}_j\right\|^2 + 2 \sum_{j\neq h}p_jp_h\langle\tilde{g}_j, \tilde{g}_h\rangle\\
			& = \sum_{j=1}^{n}p_j(1-p_j)\left\|\tilde{g}_j\right\|^2 + \left\|\sum_{j=1}^{n}p_j\tilde{g}_j\right\|^2.\\
		\end{split}
	\end{equation*}
	From $L_j$-smoothness of $f_j(x),$ $j\in[n],$ and from Lemma~\ref{lemma_smooth_bregman}, we have that
	\begin{equation*}
		\begin{split}
			\mathbb{E}\left[\left\|\sum_{j=1}^{n}\mathbb{I}_j\tilde{g}_j(x)\right\|^2\right] & \leq \max_j\{p_j(1-p_j)\}\left(\sup_{k\in\mathbb{Z}}\frac{2a_{k+1}}{a_k+a_{k+1}}\right)^2\sum_{j=1}^n\left\|\nabla f_j(x)\right\|^2\\
			& + n^2\max_j\{p_j^2\}\left(\sup_{k\in\mathbb{Z}}\frac{2a_{k+1}}{a_k+a_{k+1}}\right)^2\left\|\nabla f(x)\right\|^2\\
			& \leq 2\max_j\{p_j(1-p_j)\}\left(\sup_{k\in\mathbb{Z}}\frac{2a_{k+1}}{a_k+a_{k+1}}\right)^2\sum_{j=1}^{n}L_jD_{f_j}\left(x, x^{*}\right)\\
			& + n^2\max_j\{p_j^2\}\left(\sup_{k\in\mathbb{Z}}\frac{2a_{k+1}}{a_k+a_{k+1}}\right)^2\left\|\nabla f(x)\right\|^2\\
			& \leq 2n\max_j\{L_j\}\max_j\{p_j(1-p_j)\}\left(\sup_{k\in\mathbb{Z}}\frac{2a_{k+1}}{a_k+a_{k+1}}\right)^2\cdot D_{f}\left(x, x^{*}\right)\\
			& + 2n\max_j\{L_j\}\max_j\{p_j(1-p_j)\}\left(\sup_{k\in\mathbb{Z}}\frac{2a_{k+1}}{a_k+a_{k+1}}\right)^2\Delta^{*}\\
			& + n^2\max_j\{p_j^2\}\left(\sup_{k\in\mathbb{Z}}\frac{2a_{k+1}}{a_k+a_{k+1}}\right)^2\left\|\nabla f(x)\right\|^2.\\
		\end{split}
	\end{equation*}
	For the second term in \eqref{eq_distributed_rounding_snd_mmt}, we have
	\begin{equation*}
		\begin{split}
			\mathbb{E}\left[\left\|\sum_{j=1}^n\left(1-\mathbb{I}_j\right)\nabla f_j(x)\right\|^2\right] & = \sum_{j=1}^{n}\Exp{\left(1-\mathbb{I}_j\right)^2}\left\|\nabla f_j(x)\right\|^2\\
			& + 2\sum_{j\neq h}\Exp{(1 - \mathbb{I}_j)}\Exp{(1 - \mathbb{I}_h)}\langle \nabla f_j(x), \nabla f_h(x) \rangle\\
			& = \sum_{j=1}^n(1 - p_j)\left\|\nabla f_j(x)\right\|^2\\
			& + 2\sum_{j\neq h}(1 - p_j)(1 - p_h)\langle \nabla f_j(x), \nabla f_h(x) \rangle\\
			& = \sum_{j=1}^n(1 - p_j)p_j\left\|\nabla f_j(x)\right\|^2 + \left\|\sum_{j=1}^{n}\left(1-p_j\right)\nabla f_j(x)\right\|^2\\
			& \leq\max_j\{p_j(1-p_j)\}\sum_{j=1}^n\left\|\nabla f_j(x)\right\|^2\\
			& + n^2\max_j\{(1-p_j)^2\}\left\|\nabla f(x)\right\|^2.\\
		\end{split}
	\end{equation*}
	Further, due to $L_j$-smoothness of $f_j,$ $j\in[n],$ and due to Lemma~\ref{lemma_smooth_bregman}, we obtain
	\begin{equation*}
		\begin{split}
			\mathbb{E}\left[\left\|\sum_{j=1}^n\left(1-\mathbb{I}_j\right)\nabla f_j(x)\right\|^2\right] & \leq 2\max_j\{p_j(1-p_j)\}\sum_{j=1}^nL_jD_{f_j}(x,x^{*})\\
			& + n^2\max_j\{(1-p_j)^2\}\left\|\nabla f(x)\right\|^2\\
			& \leq 2n\max_j\{p_j(1-p_j)\}\max_j\{L_j\}D_{f}(x,x^{*})\\
			& + 2n\max_j\{p_j(1-p_j)\}\max_j\{L_j\}\Delta^{*}\\
			& + n^2\max_j\{(1-p_j)^2\}\left\|\nabla f(x)\right\|^2\\
		\end{split}
	\end{equation*}
	Therefore, from \eqref{eq_distributed_rounding_snd_mmt}, we have
	\begin{equation*}
		\begin{split}
			\Exp{\left\|g(x)\right\|^2} & \leq\frac{4}{n}\max_j\{L_j\}\max_j\{p_j(1-p_j)\}\left(\left(\sup_{k\in\mathbb{Z}}\frac{2a_{k+1}}{a_k+a_{k+1}}\right)^2 + 1\right)D_{f}\left(x, x^{*}\right)\\
			& + 2\max_j\{p_j^2\}\left(\left(\sup_{k\in\mathbb{Z}}\frac{2a_{k+1}}{a_k+a_{k+1}}\right)^2 + 1\right)\left\|\nabla f(x)\right\|^2\\
			& + \frac{4}{n}\max_j\{L_j\}\max_j\{p_j(1-p_j)\}\left(\left(\sup_{k\in\mathbb{Z}}\frac{2a_{k+1}}{a_k+a_{k+1}}\right)^2 + 1\right)\Delta^{*}.
		\end{split}
	\end{equation*}
	\begin{flushright}
		$\blacksquare$
	\end{flushright}
	
	\section{Known estimators in biased ABC framework: proofs for Section~\ref{section_grad_estimators}}\label{section_grad_estimators_proofs}
	\subsection{Proof of Claim \ref{claim_top_l}}
	Observe that
	\begin{equation*}
		\frac{\left(\nabla f(x)\right)^2_{(d-k+1)} + \ldots + \left(\nabla f(x)\right)^2_{(d)}}{k} \geq \frac{\left(\nabla f(x)\right)_1^2 + \ldots + \left(\nabla f(x)\right)_d^2}{d},
	\end{equation*}
	and $$\langle g(x), \nabla f(x)\rangle = \left\|g(x)\right\|^2=\left(\nabla f(x)\right)^2_{(d-k+1)} + \ldots + \left(\nabla f(x)\right)^2_{(d)}.$$ Therefore, $$\langle g(x), \nabla f(x)\rangle\geq \frac{k}{d}\left\|\nabla f(x)\right\|^2,$$ and $ b$ can be set to $\frac{k}{d},$ $  c$ can be set to $0.$
	
	Clearly, $\left\|g(x)\right\|^2\leq \left\|\nabla f(x)\right\|^2$ which implies that $A=C=0,$ $B=1.$
	\begin{flushright}
		$\blacksquare$
	\end{flushright}
	
	\subsection{Proof of Claim \ref{claim_random_l}}
	Observe that
	\begin{equation*}
		\left\langle\Exp{g(x)}, \nabla f(x) \right\rangle = \left\|\nabla f(x)\right\|^2.
	\end{equation*}
	This implies that $ b=1,  c=0.$ Also, notice that
	\begin{equation*}
		\Exp{\left\|g(x)\right\|^2} = \left(\frac{d}{k}\right)^2\Exp{\sum_{i \in S}\left(\nabla f(x)\right)_i^2e_i} = \frac{d}{k}\left\|\nabla f(x)\right\|^2.
	\end{equation*}
	Therefore, $A=C=0,$ $B=\frac{d}{k}.$
	\begin{flushright}
		$\blacksquare$
	\end{flushright}	
	\subsection{Proof of Claim \ref{claim_biased_random_l}}
	Observe that
	\begin{equation*}
		\langle \mathbb{E}\left[g(x)\right], \nabla f(x) \rangle  = \frac{k}{d}\left\|\nabla f(x)\right\|^2.
	\end{equation*}
	This implies that $ b=\frac{k}{d},$ $  c = 0.$ Also, notice that
	\begin{equation*}
		\mathbb{E}\left[\left\|g(x)\right\|^2\right] = \mathbb{E}\left[\sum_{i\in S}\left(\nabla f(x)\right)_i^2e_i\right] = \frac{k}{d}\left\|\nabla f(x)\right\|^2.
	\end{equation*}
	Therefore, $A=C=0,$ $B=\frac{k}{d}.$
	\begin{flushright}
		$\blacksquare$
	\end{flushright}
	\subsection{Proof of Claim \ref{claim_adaptive_random_l_sparsifier}} 
	Lemma 6 of \citep{BezHorRichSaf} states that adaptive random sparsification operator belongs to $\mathbb{B}^1\left(\frac{1}{d}, 1\right), \mathbb{B}^2\left(\frac{1}{d}, 1\right), \mathbb{B}^3(d)$. It follows that $A=0,$ $B=1,$ $C=0$ (see \eqref{eq_implication}) and $ b=\frac{1}{d},$ $  c=0.$ 
	\begin{flushright}
		$\blacksquare$
	\end{flushright}
	\subsection{Proof of Claim \ref{claim_general_adaptive_rounding}}
	\begin{definition}\label{def_unbiased_compressive}
		Let $\omega\geq 1.$ An estimator $g(x)$ belongs to a set $\mathbb{U}\left(\omega\right),$ if $g(x)$ is unbiased $(\Exp{g(x)}=\nabla f(x),$ for all $x\in\mathbb{R}^d),$ and if its second moment is bounded as
		\begin{equation}\label{eq_unbiased_compressive}
			\Exp{\left\|g(x)\right\|^2}\leq\omega\left\|\nabla f(x)\right\|^2, \quad\forall x\in\mathbb{R}^d.
		\end{equation}
	\end{definition}
	Lemma 8 of \citep{BezHorRichSaf} states that general unbiased rounding operator belongs to $\mathbb{U}(\omega)$ with
	$$
	\omega=\frac{Z}{4} = \frac{1}{4} \sup _{k \in \mathbb{Z}}\left(\frac{a_k}{a_{k+1}}+\frac{a_{k+1}}{a_k}+2\right),
	$$
	where $Z$ is defined in \eqref{eq_rounding_constant}.
	
	Since $g(x)$ is unbiased, we have $ b=1,$ $  c=0.$ From \eqref{eq_unbiased_compressive} we have that $A=C=0,$ $B=\frac{Z}{4}.$
	\begin{flushright}
		$\blacksquare$
	\end{flushright}
	\subsection{Proof of Claim \ref{claim_biased_rounding}} 
	Lemma 9 of \citep{BezHorRichSaf} states that general biased rounding operator belongs to $\mathbb{B}^1(\alpha, \beta), \mathbb{B}^2(\gamma, \beta)$, and $\mathbb{B}^3(\delta)$, where
	$$
	\beta=F, \quad \gamma=G, \quad \alpha=\gamma^2, \quad \delta=\sup _{k \in \mathbb{Z}} \frac{\left(a_k+a_{k+1}\right)^2}{4 a_k a_{k+1}} .
	$$
	
	Therefore, 
	$$
	A=C=  c=0, \quad B=F^2,\quad  b = \frac{G^2}{F}.
	$$
	with $F$ and $G$ defined in \eqref{eq_biased_rounding_consts}.
	\begin{flushright}
		$\blacksquare$
	\end{flushright}
	\subsection{Proof of Claim \ref{claim_natural_compression}}
	Since natural compression estimator is a special case of general unbiased rounding estimator with $a_k=2^k,$ we obtain that $g(x)$ belongs to a set $\mathbb{U}\left(\frac{9}{8}\right),$ and, in a similar way as in the proof of Claim~\ref{claim_general_adaptive_rounding}, we obtain that $A=C=  c=0,$ $B=\frac{9}{8},$ $ b=1.$
	\begin{flushright}
		$\blacksquare$
	\end{flushright}
	\subsection{Proof of Claim \ref{claim_general_exponential_dithering}}
	Lemma 10 of \citep{BezHorRichSaf} states that exponential dithering operator belongs to $\mathbb{U}\left(H_a\right).$ Since $g(x)$ is unbiased, we have that $ b=1,$ $  c=0.$ From \eqref{eq_unbiased_compressive} we have that $A=C=0,$ $B=H_a.$
	\begin{flushright}
		$\blacksquare$
	\end{flushright}
	\subsection{Proof of Claim \ref{claim_natural_dithering}} Natural dithering estimator is a special case of exponential dithering operator in case when $a=2.$ Therefore, Claim~\ref{claim_natural_dithering} is a direct consequence of Claim~\ref{claim_general_exponential_dithering}, and we have $A=C=  c=0,$ $B=H_2,$ $ b=1.$
	\begin{flushright}
		$\blacksquare$
	\end{flushright}
	\subsection{Proof of Claim \ref{claim_composition_top_l_exp_dithering}} Lemma 11 of \citep{BezHorRichSaf} states that the composition operator of Top-$k$ sparsification and exponential dithering with base $a$ belongs to $\mathbb{B}^1\left(\frac{k}{d}, H_a\right), \mathbb{B}^2\left(\frac{k}{d}, H_a\right), \mathbb{B}^3\left(\frac{d}{k} H_a\right)$, where $H_a$ is a constant defined in \eqref{eq_dithering_constant}.
	
	Therefore, from \eqref{eq_implication}, we have
	\begin{equation*}
		A = 0,\; B =H_a^2 ,\; C = 0,\;  b = \frac{k}{d H_a},\;   c = 0.
	\end{equation*}	
	\begin{flushright}
		$\blacksquare$
	\end{flushright}
	\subsection{Proof of Claim \ref{claim_gaussian_smoothing}}
	When $f$ is convex and satisfies Assumption \ref{ass_smooth} with a constant $L,$ \citet{NestSpok} (Lemma~3 and Theorem~4) bound the bias in the following way:
	\begin{equation*}
		\left\|\Exp{g_{GS}(x)} - \nabla f(x)\right\|^2\leq\frac{\tau^2}{4}L^2(d+3)^3.
	\end{equation*}
	Therefore, due to \eqref{eq_equiv_stich_scalar} and \eqref{eq_beat_stich}, we obtain that
	\begin{equation*}
		\langle\nabla f(x), \mathbb{E}\left[g(x)\right]\rangle \geq \frac{1}{2}\left\|\nabla f(x)\right\|^2 - \frac{\tau^2}{8}L^2(d+3)^3.
	\end{equation*}
	Further, from Theorem 4 of \citep{NestSpok}, we have that
	\begin{equation*}
		\Exp{\left\|g_{GS}(x)\right\|^2} \leq 2(d+4)\left\|\nabla f(x)\right\|^2 + \frac{\tau^2}{2}L^2(d+6)^3.
	\end{equation*}
	We can choose 
	\begin{equation*}
		A=A_{GS}\eqdef0,\;B=B_{GS}\eqdef2(d+4),\;C=C_{GS}\eqdef\frac{\tau^2}{2}L^2(d+6)^3,
	\end{equation*}
	\begin{equation*}
		 b= b_{GS}=\frac{1}{2},\;   c=  c_{GS}\eqdef\frac{\tau^2}{8}L^2(d+3)^3.
	\end{equation*}
	\begin{flushright}
		$\blacksquare$
	\end{flushright}
	\subsection{Proof of Claim~\ref{claim_hard_threshold_sparsifier}}
	It is easy to see that it satisfies Assumption~\ref{ass_abs_compr} with $\Delta = w\sqrt{d}.$ Then, it follows that $\left\|\Exp{g(x)} - \nabla f(x)\right\|^2\leq w^2d.$ Therefore, $\langle \Exp{g(x)}, \nabla f(x)\rangle \geq \left\|\nabla f(x)\right\|^2 - w^2d + \left\|\Exp{g(x)}\right\|^2 \geq \left\|\nabla f(x)\right\|^2 - w^2d.$ We can choose $ b=1,$ $  c=w^2d.$
	
	Further, $\Exp{\left\|g(x)\right\|^2}=\left\|g(x)\right\|^2 \leq\left\|\nabla f(x)\right\|^2.$ It means that we can choose $A=C=0,$ $B=1.$
	\begin{flushright}
		$\blacksquare$
	\end{flushright}
	\subsection{Proof of Claim~\ref{claim_scaled_rounding}}
	Observe, that $g(x)$ satisfies Assumption~\ref{ass_abs_compr} with $\Delta=\frac{\sqrt{d}}{\chi}.$ Indeed, for every $j\in[d],$ we have
	\begin{equation*}
		\frac{1}{n}\sum_{i=1}^{n}(\nabla f_i(x))_j - \frac{1}{\chi} \leq \frac{1}{n}\sum_{i=1}^{n}\frac{1}{\chi}\left(R\left(\chi\nabla f_i(x)\right)\right)_j \leq \frac{1}{n}\sum_{i=1}^{n}(\nabla f_i(x))_j + \frac{1}{\chi}.
	\end{equation*}
	Therefore, $\left\|g(x) - \nabla f(x)\right\|^2\leq\frac{d}{\chi^2}.$ In accordance with Theorem~\ref{thm_assumptions_in_our_framework}~-~\ref{item_framework_abs_compr}, we obtain that we can choose $A=0,$ $B=2,$ $C=\frac{2d}{\chi^2},$ $b=\frac{1}{2},$ $c=\frac{d}{2\chi^2}.$
	\begin{flushright}
		$\blacksquare$
	\end{flushright}
	\subsection{Proof of Claim~\ref{claim_biased_dithering}}
	In accordance with \citet[Lemma~2]{khirirat2018gradient}, $g(x)$ satisfies Assumption~\ref{ass_breq} with $\rho=1,$ $\zeta = d.$ It follows from Theorem~\ref{thm_assumptions_in_our_framework} that $g(x)$ satisfies $\hyperlink{Biased ABC}{Biased ABC}$ with $A=0,$ $B=d,$ $C=0,$ $b=1$ and $c=0.$
	\begin{flushright}
		$\blacksquare$
	\end{flushright}
	\subsection{Proof of Claim~\ref{claim_sign_compression}}
	In accordance with \citet[Lemma~8]{karimireddy2019ef} $g(x)$ satisfies Assumption~\ref{ass_third_set} with $\delta(x) = \frac{d\|{x}\|_2^2}{\|{x}\|_1^2} \leq d.$ It follows from Theorem~\ref{thm_assumptions_in_our_framework} that $g(x)$ satisfies $\hyperlink{Biased ABC}{Biased ABC}$ with $A=0,$ $B=2\left(2-\frac{1}{d}\right),$ $C=0,$ $b=\frac{1}{2d}$ and $c=0.$
	\begin{flushright}
		$\blacksquare$
	\end{flushright}
	\section{Proofs of the results presented in Table~\ref{tab_estimators_in_assumptions_short}}\label{apx_inclusion_estimators_frameworks}
	We proved in Claim~\ref{claim_distributions} that Biased independent sampling estimator (see Def.~\ref{def_biased_sampling_no_replacement}) satisfies \hyperlink{Biased ABC}{Biased ABC} assumption. On the other hand, in Theorem~\ref{thm_informal_abc} (parts~\ref{item_abc_from_fsml}~and~\ref{item_abc_from_bnd}) we show that it Assumptions~\ref{ass_stich_decomposition}~and~\ref{ass_first_and_second_mmt_limits} do not hold for it. Therefore, it does not satisfy Assumptions~\ref{ass_first_set}~--~\ref{ass_first_and_second_mmt_limits} (see Figure~\ref{fig_diagram}).
	
	In \citep[Lemma~7]{BezHorRichSaf} it is proven that Top-$k$ (see Def.~\ref{def_top_ell}) estimator satisfies Assumption~\ref{ass_third_set}. Therefore, in accordance with Figure~\ref{fig_diagram}, we only need to verify that Assumption~\ref{ass_breq} holds, and Assumption~\ref{ass_abs_compr} does not hold for Top-$k.$ The argument in the proof of Claim~\ref{claim_top_l} shows that Assumption~\ref{ass_breq} is satisfied for $g(x).$ Consider $f(x) = \frac{x_1^2}{2}+\frac{x_2^2}{2},$ $x\in\mathbb{R}^2,$ and Top-$1$ estimator. For every $x_2$ in $\mathbb{R},$ consider $x=(x_1,x_2)\in\mathbb{R}^2,$ such that $x_1\geq x_2.$ Clearly, $g(x)=\left(x_1, 0\right),$ $\nabla f(x) = (x_1, x_2).$ Then, $\left\|g(x) - \nabla f(x)\right\|^2=x_2^2.$ For any $\Delta\geq 0,$ there exists $x_2$ such that $x_2^2\geq\Delta^2.$ Therefore, Assumption~\ref{ass_abs_compr} does not hold for $g(x).$
	
	 Rand-$k$ (see Def.~\ref{def_random_l}) is a stochastic estimator, it does not satisfy Assumption~\ref{ass_breq}. Since $\left\|\Exp{g(x)} - \nabla f(x)\right\|^2=0,$ $\Exp{\left\|g(x) - \Exp{g(x)}\right\|^2} = \left(\frac{d}{k} - 1\right)\left\|\nabla f(x)\right\|^2,$ it satisfies Assumption~\ref{ass_BV}. It remains to show that it does not satisfy Assumptions~\ref{ass_third_set}~and~\ref{ass_abs_compr}. Consider $f(x) = \frac{x_1^2}{2}+\frac{x_2^2}{2},$ $x\in\mathbb{R}^2,$ and Rand-$1$ estimator. For every $x_2$ in $\mathbb{R},$ consider $x=(x_1,x_2)\in\mathbb{R}^2,$ such that $x_1\geq x_2.$ Clearly, $\Exp{\left\|g(x) - \nabla f(x)\right\|^2} = \left\|\nabla f(x)\right\|^2 = x_1^2+x_2^2,$ and this expression can not be bounded by any constant $\Delta^2\geq 0,$ which implies that Assumption~\ref{ass_abs_compr} does not hold. Also, there is no $\delta>0,$ such that $\left\|\nabla f(x)\right\|\leq\left(1 - \frac{1}{\delta}\right)\left\|\nabla f(x)\right\|^2,$ for all $x\in\mathbb{R}^2,$ which implies that Assumption~\ref{ass_third_set} does not hold.
	
	In \citep[Lemma~5]{BezHorRichSaf} it is proven that Biased Rand-$k$ estimator (see Def.~\ref{def_biased_random_l}) satisfies Assumption~\ref{ass_third_set}. Therefore, in accordance with Figure~\ref{fig_diagram}, we only need to verify that Assumptions~\ref{ass_breq}~and~\ref{ass_abs_compr} do not hold for Biased Rand-$k.$ Since this estimator is stochastic, Assumption~\ref{ass_breq} does not hold. Consider $f(x) = \frac{x_1^2}{2}+\frac{x_2^2}{2},$ $x=(x_1,x_2)\in\mathbb{R}^2,$ and Biased Rand-$1$ estimator. We have that $\Exp{\left\|g(x) - \nabla f(x)\right\|^2} = \frac{1}{2}\left\|\nabla f(x)\right\|^2=\frac{x_1^2}{2}+\frac{x_2^2}{2},$ and this expressioin can not be bounded by any constant $\Delta^2\geq 0.$ Therefore, Assumption~\ref{ass_abs_compr} is not satisfied.
	
	In \citep[Lemma~6]{BezHorRichSaf} it is proven that Adaptive random sparsification (see Def.~\ref{def_adaptive_random_sparsification}) satisfies Assumption~\ref{ass_third_set}. Therefore, in accordance with Figure~\ref{fig_diagram}, we only need to verify that Assumptions~\ref{ass_breq}~and~\ref{ass_abs_compr} do not hold for Adaptive random sparsification estimator. Since it is stochastic, Assumption~\ref{ass_breq} does not hold. Consider $f(x) = \frac{x_1^2}{2}+\frac{x_2^2}{2},$ $x=(x_1, x_2)\in\mathbb{R}^2,$ and Adaptive random sparsification estimator. Observe that 
	\begin{equation*}
		\begin{split}
			\Exp{\left\|g(x) - \nabla f(x)\right\|^2} & = \left\|\nabla f(x)\right\|^2_2\left(1 - \frac{\left\|\nabla f(x)\right\|^3_3}{\left\|\nabla f(x)\right\|_1\left\|\nabla f(x)\right\|^2_2}\right)\\
			& =\left(x_1^2+x_2^2\right)\left(1 - \frac{x_1^3+x_2^3}{\left(|x_1|+|x_2|\right)(x_1^2+x_2^2)}\right).\\
		\end{split}
	\end{equation*}
	Let $\lambda>0$ be some constant. Consider $x\in \mathbb{R}^2$ such that $|x_1|=\lambda|x_2|.$ Then
	$$
	\Exp{\left\|g(x) - \nabla f(x)\right\|^2} = \lambda x_2^2,
	$$
	and, for any $\Delta^2\geq 0,$ there exists $x_2\in\mathbb{R},$ such that $\lambda x_2^2 \geq \Delta^2.$ Therefore, Assumption~\ref{ass_abs_compr} does not hold.
	
	General unbiased rounding (see Def.~\ref{def_general_unbiased_rounding}) belongs to $\mathbb{U}\left(\frac{Z}{4}\right)$
	(see Claim~\ref{claim_general_adaptive_rounding}) with $Z$ defined in \eqref{eq_rounding_constant}. Then, $\Exp{\left\|g(x) - \nabla f(x)\right\|^2}\leq\left(\frac{Z}{4} - 1\right)\left\|\nabla f(x)\right\|^2,$ and $g(x)$ satisfies Assumption~\ref{ass_BV}. Therefore, in accordance with Figure~\ref{fig_diagram}, we only need to verify that Assumptions~\ref{ass_third_set}, \ref{ass_breq}~and~\ref{ass_abs_compr} do not hold. Let $a_k=6^k,$ $k\in\mathbb{Z}.$ Consider $f(x) = \frac{x^2}{2},$ $x=\in\mathbb{R},$ and General unbiased rounding estimator. Then,
	\begin{equation*}
		\begin{split}
			\Exp{\left\|g(x) - \nabla f(x)\right\|^2} &= \Exp{\left\|g(x)\right\|^2} - \left\|\nabla f(x)\right\|^2\\
			& = -6^{2k+1} + 7\cdot 6^k \cdot x - x^2.\\
		\end{split}
	\end{equation*}
	Let $x=\frac{7}{4}\cdot 6^k.$ Then $\Exp{\left\|g(x) - \nabla f(x)\right\|^2}\geq x^2:$
	\begin{equation*}
		\begin{split}
			\frac{\Exp{\left\|g(x) - \nabla f(x)\right\|^2}}{x^2} = \frac{51}{49}>1.
		\end{split}
	\end{equation*}
	Then, Assumption~\ref{ass_third_set} does not hold. Note that, for every constant $\Delta^2\geq 0,$ there exists $k\in\mathbb{Z},$ such that $x^2>\Delta^2.$ Therefore, Assumption~\ref{ass_abs_compr} is not satisfied. Since this estimator is stochastic, Assumption~\ref{ass_breq} does not hold as well.
	
	Natural compression (see Def.~\ref{def_natural_compression}) belongs to $\mathbb{U}\left(\frac{9}{8}\right)$
	(see Claim~\ref{claim_general_adaptive_rounding}). Then, $\Exp{\left\|g(x) - \nabla f(x)\right\|^2}\leq \frac{1}{8}\left\|\nabla f(x)\right\|^2,$ and $g(x)$ satisfies Assumption~\ref{ass_third_set}. Therefore, in accordance with Figure~\ref{fig_diagram}, we only need to verify that Assumptions~\ref{ass_breq}~and~\ref{ass_abs_compr} do not hold. Since $g(x)$ is a stochastic estimator, Assumption~\ref{ass_breq} is not satisfied. Consider $f(x) = \frac{x^2}{2},$ $x\in\mathbb{R},$ and Natural compression estimator. Then
	\begin{equation*}
		\begin{split}
			\Exp{\left\|g(x) - \nabla f(x)\right\|^2} &= \Exp{\left\|g(x)\right\|^2} - \left\|\nabla f(x)\right\|^2\\
			& = -2^{2k+1} + 3\cdot 2^k\cdot x - x^2.\\
		\end{split}
	\end{equation*}
	Let $x=\frac{3}{2}\cdot 2^k.$ Then $\Exp{\left\|g(x) - \nabla f(x)\right\|^2} = 2^{2k-2}.$ For every constant $\Delta^2\geq 0,$ there exists $k\in\mathbb{Z},$ such that $2^{2k-2}>\Delta^2.$ Therefore, Assumption~\ref{ass_abs_compr} does not hold.
	
	In Claim~\ref{claim_scaled_rounding} we prove that Scaled integer rounding (see Def.~\ref{def_scaled_rounding}) satisfies Assumption~\ref{ass_abs_compr}. Also, it is easy to see that $\left\|g(x) - \nabla f(x)\right\|^2\leq \left\|\nabla f(x)\right\|^2,$ and equality holds for $f(x)= \frac{x^2}{2},$ $n=d=1,$ $x=0.25.$ Therefore, $g(x)$ does not satisfy Assumption~\ref{ass_third_set}, and satisfies Assumption~\ref{ass_BV}. Since rounding preserves the sign (or rounds a number to $0$), we have that $\langle \nabla f(x), \frac{1}{\chi}R\left(\chi \left(\nabla f(x)\right)_i\right)\rangle\geq  0.$ Also, $\left\|g(x)\right\|^2\leq 4\left\|\nabla f(x)\right\|^2.$ This means, $g(x)$ satisfies Assumption~\ref{ass_breq}. There is a misprint in Table~\ref{tab_estimators_in_assumptions_short}, refer to Table~\ref{tab_estimators_in_assumptions_full}.
	\section{Relation between assumption~\ref{ass_third_set} and contractive compression}\label{apx_contractive_relations}
	In Assumption \ref{ass_third_set}, one can observe a resemblance to the contractive compression property, as shown in the following equation:
	\begin{equation}\label{eq_contractive_compression}
		\mathbb{E}\left[\left\|\cC(x) - x\right\|^2\right]\leq\left(1 - \frac{1}{\delta}\right)\left\|x \right\|^2 \quad \forall x \in \R^d.
	\end{equation}
	The contractive compression property is commonly utilized in methods dealing with biased compression (e.g., TopK), as demonstrated in various studies \citep{Stich-EF-NIPS2018, karimireddy2019ef, stich2020error, BezHorRichSaf, gorbunov2020linearly, cordonnier2018convex, EF21, EF21BW, richtarik20223pc}. However, equations \eqref{eq_contractive_biased} and \eqref{eq_contractive_compression} are not generally equivalent since in practise one may not aim to compress exactly a gradient itself.
	
	\section{Relation between Assumption~\ref{ass_abs_compr} and absolute compression}\label{apx_absolute_compression_relations}
	Within Assumption~\ref{ass_abs_compr}, a similarity to the absolute compression property
	\begin{equation}\label{eq_abs_compr1}
		\mathbb{E}\left[\|\cC(x)-x\|^2\right] \leq \Delta^2 \quad \forall x \in \R^d
	\end{equation}
	can be discerned.
	Nonetheless, it should be noted that the expressions in equations \eqref{eq_abs_compr} and \eqref{eq_abs_compr1} do not typically exhibit equivalence.
	
	Various instances of absolute compression have been extensively employed by practitioners over the years \citep{DoubleSqueeze, sahu2021rethinking, danilova2022distributed}. A prominent example is the hard-threshold sparsifier $\mathcal{C}_{\mathrm{HT}}(x)$ \citep{sahu2021rethinking, dutta2020discrepancy, strom2015scalable}. It can be demonstrated that $\mathcal{C}_{\mathrm{HT}}(x)$ adheres to Eq. \eqref{eq_abs_compr} with $\Delta=\lambda \sqrt{d}$. Additional examples encompass (stochastic) rounding schemes with limited error \citep{gupta2015deep, khirirat2020compressed} and integer rounding \citep{sapio2019scaling, mishchenko2021intsgd}.
	
	The absolute compression assumption has also been featured in several studies \citep{sahu2021rethinking, danilova2022distributed, khirirat2020compressed, khirirat2022eco, chen2021quantized}, which examine the Error Feedback mechanism \citep{Stich-EF-NIPS2018, karimireddy2019ef, stich2020error}.
	
	Specifically, \citet{sahu2021rethinking} established that hard-threshold sparsifiers are optimal for minimizing total error (a unique quantity that emerges in the analysis of EC-SGD) with respect to any fixed sequence of errors.
	
	Furthermore, the authors of \citep{sahu2021rethinking} elucidate both the theoretical and practical advantages of absolute compressors in comparison to $\delta$-contractive ones expressed in Equation \eqref{eq_contractive_compression}.
	
	\section{Relations between the estimators from Assumptions \ref{ass_first_set}--\ref{ass_third_set}}
	Below we restate Theorem 2 from \citep{BezHorRichSaf} about the relations between these sets in terms of biased gradient estimators instead of biased compressors.
	\begin{theorem}[Relations between the estimators from Assumptions \ref{ass_first_set}--\ref{ass_third_set}]\label{thm_gradients_equivalence} Let $\lambda>0$ be a scaling parameter.
		\begin{enumerate}
			\item If $g\in \mathbb{B}^{1}(\alpha,\beta),$ then
			\begin{itemize}
				\item $\beta^2\geq\alpha$ and $\lambda g\in \mathbb{B}^1\left(\lambda^2\alpha,\lambda\beta\right),$
				\item $g\in\mathbb{B}^2\left(\alpha,\beta^2\right)$ and $\frac{1}{\beta}g\in\mathbb{B}^3\left(\frac{\beta^2}{\alpha}\right).$
			\end{itemize}
			\item If $g\in\mathbb{B}^2\left(\tau, \beta\right),$ then
			\begin{itemize}
				\item $\beta\geq\tau$ and $\lambda g\in\mathbb{B}^2\left(\lambda\tau, \lambda\beta\right),$
				\item $g\in\mathbb{B}^1\left(\tau^2, \beta\right)$ and $\frac{1}{\beta}g\in\mathbb{B}^3\left(\frac{\beta}{\tau}\right)$
			\end{itemize}
			\item If $g\in\mathbb{B}^3\left(\delta\right),$ then
			\begin{itemize}
				\item $\delta\geq 1,$
				\item $g\in\mathbb{B}^2\left(\frac{1}{2\delta}, 2\right)\subseteq \mathbb{B}^1\left(\frac{1}{4\delta^2}, 2\right).$
			\end{itemize}
		\end{enumerate}
	\end{theorem}
	We do not prove it here and refer the reader to the original paper.
	\section{Equivalence of Assumption~\ref{ass_stich_decomposition} and \citep[Def. 1]{AjallStich}}\label{apx_equiv_stich_reformulation}
	Definition~1 in \citep{AjallStich} is written in the following way.
	\begin{definition}\label{def_original_stich}
		Let $\left(\mathcal{D}, \mathcal{F}\right)$ be a measurable space and $Y$ be a random element of this space.  Let gradient estimator $g(x,Y)$ have a form
		\begin{equation*}
			g(x,Y) = \nabla f(x) + b(x) + \mathcal{Z}(x,Y),
		\end{equation*}
		where $b(x):\mathbb{R}^d\to\mathbb{R}^d$ is a bias and $\mathcal{N}:\mathbb{R}^d\times \mathcal{D} \to\mathbb{R}^d$ is a zero-mean noise, i.e. $\mathbb{E}\left[\mathcal{Z}(x,Y)|Y\right]=0,$ for all $x\in\mathbb{R}^d.$
		
		There exist constants $M,\sigma^2\geq 0$ such that
		\begin{equation}\label{eq_noise_stich}
			\mathbb{E}\left[\left\|\mathcal{Z}(x,Y)\right\|^2\right]\leq M\left\|\nabla f(x) + b(x)\right\|^2+\sigma^2,\quad\forall x\in\mathbb{R}^d.
		\end{equation}
		There exist constants $0\leq m < 1$ and $\varphi^2\geq 0,$ such that
		\begin{equation}\label{eq_bias_stich}
			\left\|b(x)\right\|^2\leq m\left\|\nabla f(x)\right\|^2 + \varphi^2,\quad \forall x\in\mathbb{R}^d.
		\end{equation}
	\end{definition}
	For the purpose of clarity, we rewrote the inequalities \eqref{eq_noise_stich} and \eqref{eq_bias_stich} in the notation adopted in our paper (see Section~\ref{sect_existing_models}). Below we establish their equivalence.
	\begin{claim}\label{claim_stich_interpretation_equivalence}
		Definition \ref{def_original_stich} is equivalent to Assumption \ref{ass_stich_decomposition}.
	\end{claim}
	\noindent\textbf{Proof of Claim \ref{claim_stich_interpretation_equivalence}.}
	Observe that $\mathcal{Z}\left(x, Y\right) = g(x, Y) - \mathbb{E}\left[g(x, Y)\right],$ $\nabla f(x) + b(x) = \mathbb{E}\left[g(x, Y)\right],$ $b(x) = \mathbb{E}\left[g(x, Y) \right] - \nabla f(x).$ It remains to perform these substitutions in \eqref{eq_noise_stich} and \eqref{eq_bias_stich}.
	
	\section{Proof of Lemma \ref{lemma_smooth_bregman}}\label{section_lemma_smooth_bregman_proof}
	Let $x_{+} = x - \frac{1}{L}\nabla f(x),$ then using the $L$-smoothness of $f$ we obtain
	\begin{equation*}
		f(x_{+})\leq f(x) + \langle\nabla f(x), x_{+} - x\rangle + \frac{L}{2}\left\|x_{+}-x\right\|^2.
	\end{equation*}
	Since $f^{*}\leq f(x_{+})$ and the definition of $x_{+}$ we have,
	\begin{equation*}
		f^{*}\leq f(x_{+})\leq f(x) - \frac{1}{L}\left\|\nabla f(x)\right\|^2 + \frac{1}{2L}\left\|\nabla f(x)\right\|^2=f(x) - \frac{1}{2L}\left\|\nabla f(x)\right\|^2.
	\end{equation*}
	It remains to rearrange the terms to get the claimed result.
	\begin{flushright}
		$\blacksquare$
	\end{flushright}
	
	\section{Young's inequality}
	Throughout the paper we use the following version of a well-known inequality:
	\begin{lemma}[Young's Inequality]\label{lemma_young}
		For every $s>0,$ for any vectors $u,h\in\mathbb{R}^d,$ we have
		\begin{equation}\label{eq_peter_paul}
			||u\pm h||^2\leq\left(1+s\right)||u||^2 + \left(1+\frac{1}{s}\right)||h||^2.
		\end{equation}
		Or, equivalent,
		\begin{equation}\label{eq_scalar_prod_young}
			\pm2\langle u, h \rangle \leq s||u||^2+\frac{1}{s}||h||^2.
		\end{equation}
	\end{lemma}
	\noindent\textbf{Proof of Lemma \ref{lemma_young}.} Let $u' = \sqrt{s}u,$ $h' = \frac{h}{\sqrt{s}}.$ Then \eqref{eq_scalar_prod_young} can be rewritten as 
	$$
	\pm2\langle u', h' \rangle \leq ||u'||^2+||h'||^2.
	$$
	Or, equivalent, $\left\|u'\pm h'\right\|^2 \geq 0.$
	\begin{flushright}
		$\blacksquare$
	\end{flushright}

\end{document}